\documentclass[11pt]{article}

\usepackage[preprint]{acl}

\usepackage{times}
\usepackage{latexsym}
\usepackage{amsmath}
\usepackage{booktabs}
\usepackage{multirow}
\usepackage{makecell}
\usepackage{tcolorbox}
\usepackage{subcaption}
\usepackage{xcolor}
\usepackage{placeins}

\usepackage[T1]{fontenc}

\usepackage[utf8]{inputenc}

\usepackage{microtype}

\usepackage{inconsolata}

\usepackage{graphicx}

\title{Can Confidence Estimates Decide When Chain-of-Thought is Necessary for LLMs?}

\author{
  Samuel Lewis-Lim, Xingwei Tan, Zhixue Zhao, Nikolaos Aletras \\
  School of Computer Science, University of Sheffield \\
  United Kingdom \\
  \texttt{\small \{slewis-lim1, xingwei.tan, zhixue.zhao, n.aletras\}@sheffield.ac.uk}
}

\begin{document}
\maketitle
\begin{abstract}
Chain-of-thought (CoT) prompting is a common technique for improving the reasoning abilities of large language models (LLMs). However, extended reasoning is often unnecessary and substantially increases token usage. As such, a key question becomes how to optimally allocate compute to when reasoning is actually needed. We study this through \textit{confidence-gated CoT}, where a model produces a direct answer and a confidence estimate to decide whether to invoke CoT. We present an evaluation framework together with the first systematic study of confidence signals for this decision. We evaluate four representative confidence measures and compare them with random gating and an oracle upper bound. Experiments across two model families and diverse reasoning tasks show that existing training-free confidence measures can reduce redundant reasoning. However, we also find that the utility of individual confidence measures is inconsistent across settings. Through our evaluation framework and analysis, our study provides practical guidance toward developing and evaluating models that selectively use CoT. \footnote{Code included at: \href{https://github.com/samlewislim/cgr}{https://github.com/samlewislim/cgr}}%

\end{abstract}

\begin{figure}
    \centering\includegraphics[width=0.91\linewidth]{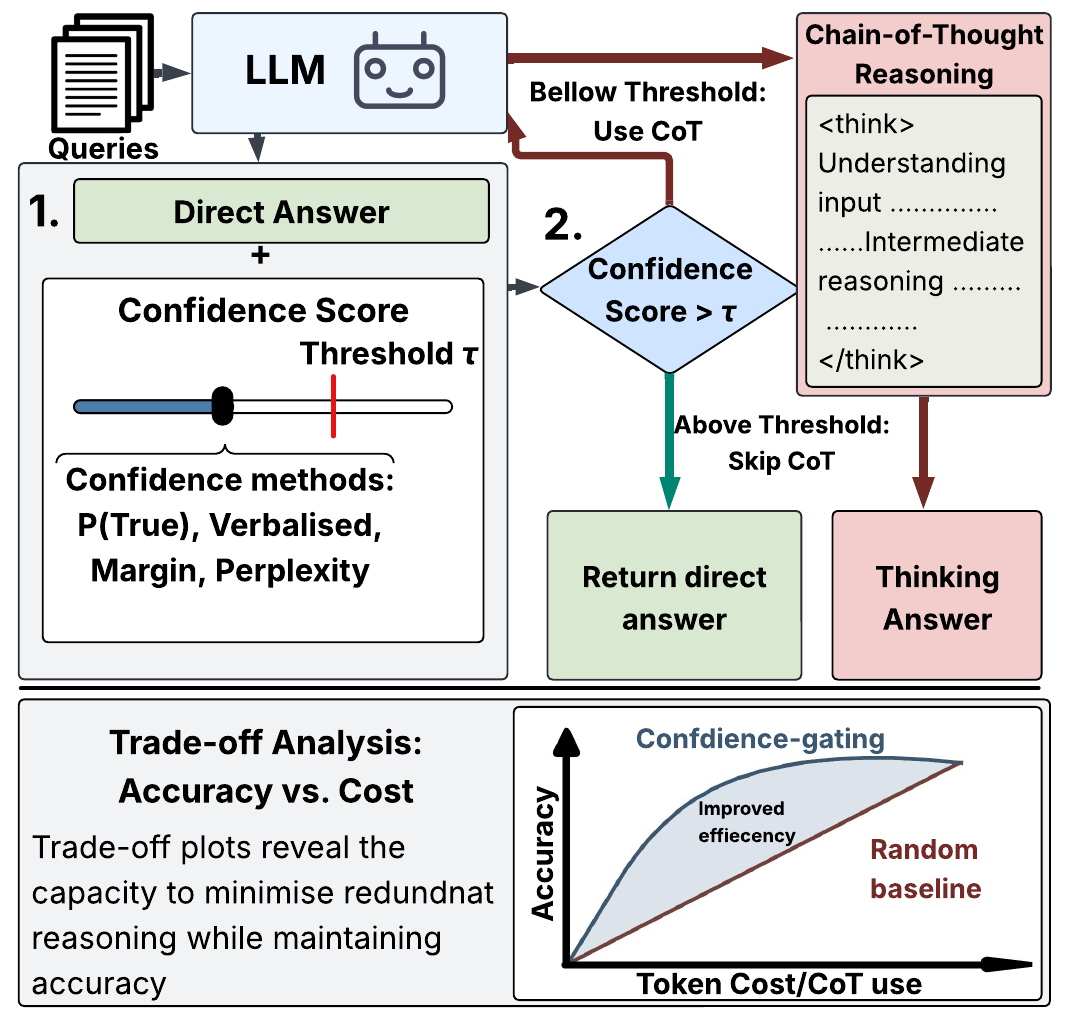}
    \caption{We evaluate if Confidence-Gated CoT (top) can effectively improve efficiency by deciding when extended reasoning is required. The trade-off analysis (bottom) illustrates how successful this trade-off is for each method compared to a random baseline.}
    \label{fig:overview_fig}
\end{figure}

\section{Introduction}

Chain-of-thought (CoT) prompting improves performance on multi-step reasoning tasks, including mathematics, symbolic reasoning, and scientific question answering \citep{wei_cot, deepseekr1report, qwen3technicalreport}. However, for tasks such as basic question answering and commonsense reasoning, CoT provides little benefit while substantially increasing token usage and latency \citep{liu2024mindstepbystep, sprague2025to, lewis-lim-etal-2025-analysing}.

Recent models offer hybrid thinking modes \citep{qwen3technicalreport}, while others provide separate instruct and thinking variants \citep{olmo2025olmo3}. In both cases, the user decides when using long CoT reasoning is appropriate. However, this still requires anticipating the necessity of CoT reasoning for each query. We refer to this as the \textit{when-to-think} decision, in contrast to \textit{how-long-to-think} methods that control the length of CoT once it has already been triggered. Most existing efficient reasoning methods address the \textit{how-long-to-think} problem \citep{yang2025dynamicearlyexitreasoning, huang2025mitigating, liu2025learnreasonefficientlyadaptive}; however, there is still comparatively little systematic analysis of the \textit{when-to-think} setting, especially using training-free signals. \citep{yue2025dontoverthinkitsurvey, jiang2025thinkneedlargehybridreasoning, chuang2025learningtoroute}.

Most past work relies on reinforcement learning or classifiers to predict when CoT helps~\citep{yue2025dontoverthinkitsurvey, jiang2025thinkneedlargehybridreasoning, chuang2025learningtoroute}. These approaches require additional training and are typically evaluated on verifiable tasks such as mathematical reasoning, making their generalisation to other task types and simpler queries unclear. While training-free methods exist, they have focused solely on perplexity~\citep{lu2025prolongedreasoningneedcertaintybased}. In contrast, \textit{we provide a unified and systematic framework for evaluating the \textit{when-to-think} decision and use it to evaluate how confidence signals can be used to make this decision.}

Confidence scores give a signal of how reliable a model's answer is~\citep{kadavath2022languagemodelsmostlyknow,kuhn2023semantic, Farquhar2024-gi}. They can be verbalised directly by the model \citep{tian-etal-2023-just-ask} or derived from its output probabilities \cite{kadavath2022languagemodelsmostlyknow}. They have already been used in model routing~\citep{titov_optimising_calls_margin, chuang2025confidentseekstrongerexploring}, where easy queries are sent to smaller models improve inference efficiency. This motivates our central question: \textit{can confidence estimates guide LLMs in deciding when to invoke CoT reasoning?}

We aim to evaluate \textit{confidence-gated CoT}, where confidence signals are used to decide if CoT reasoning is necessary. As illustrated in Figure \ref{fig:overview_fig}, we evaluate if these estimates can effectively activate CoT only when needed. To do this, we benchmark four representative confidence estimation methods across diverse reasoning tasks and models.  

\paragraph{Contributions:} (1) A unified evaluation framework for quantifying the accuracy-efficiency trade-offs of when LLMs should invoke long CoT; (2) Using this, we provide the first broad systematic evaluation of different confidence signals for compute-efficient reasoning across multiple models and diverse reasoning and non-reasoning tasks; (3) Detailed analysis of the cost-saving potential and failure modes of confidence-gated CoT.

\section{Related Work}

\subsection{Efficient and Adaptive Reasoning} 

Adaptive reasoning aims to enable LLMs to dynamically adjust the depth or length of their reasoning processes \citep{yue2025dontoverthinkitsurvey}. Prior methods either control \textit{how-long-to-think} once reasoning has started, or learn \textit{when-to-think} policies that decide whether to invoke deeper reasoning.  \textit{How-long-to-think} approaches include early-exit methods using confidence or entropy monitors, probes, or decoding-time controls. \citep{yang2025dynamicearlyexitreasoning, zhang2025reasoning, huang2025mitigating}. Complementary work shortens reasoning traces through training on shorter CoT, or uses length-aware rewards that discourage redundant steps \citep{liu2025learnreasonefficientlyadaptive, shen-etal-2025-dast, huang2025hapotraininglanguagemodels}. 

\textit{When-to-think} methods mostly rely on supervision or reinforcement learning to learn routing between reasoning modes, such as direct answering versus long CoT or fast versus slow thinking \citep{yue2025dontoverthinkitsurvey, jiang2025thinkneedlargehybridreasoning, chuang2025learningtoroute}. These approaches require additional training and are typically evaluated on verifiable domains such as mathematics. Closest to our setting, Certainty-based Adaptive Reasoning (CAR) uses answer perplexity as a trigger for longer reasoning \citep{lu2025prolongedreasoningneedcertaintybased}. In contrast, we focus on \textit{confidence-gated CoT} as a general, training-free when-to-think problem, systematically comparing multiple confidence signals across models and tasks.

\subsection{Uncertainty Estimation in LLMs}
Estimating the reliability of LLM predictions has been widely studied. Many methods derive confidence directly from model probabilities or logits. This includes perplexity, the difference between top token probabilities \citep{titov_optimising_calls_margin}, and $P($True$)$ \citep{kadavath2022languagemodelsmostlyknow}. Other approaches seek to simply prompt the LLM to output a confidence score in its response \citep{tian-etal-2023-just-ask, xiong2024can, yang2024verbalizedconfidencescoresllms}. \citet{zhou2025steerconf} build on this to develop prompting strategies specifically designed to steer the model to produce better calibrated verbalised confidence scores. Multi-sample methods generate multiple responses and measure agreement or semantic diversity, e.g., self-consistency or semantic entropy \citep{kuhn2023semantic, Farquhar2024-gi}. Finally, methods to either train the model itself or external predictors to produce more reliable confidence estimates have been proposed \citep{kossen2025semanticprobes, chuang2025learningtoroute, damani2025binaryrewardstraininglms, stangel2025rewardingdoubtreinforcementlearning}. In this work, we focus on single-pass, training-free confidence signals, prioritising computational efficiency and low inference cost.

\subsection{Model Cascades and Routing}
Model cascades and routing methods dynamically switch between multiple models.
\citet{ong2025routellm} propose to decide when to route based on a win prediction model that estimates the probability of a strong model win over a weak model for a given query.
\citet{feng2025graphrouter} predict the effect and cost of potential edges in a graph where the task, query, and LLM are modelled as heterogeneous nodes.
\citet{titov_optimising_calls_margin} find that simple confidence measures can effectively route harder queries to stronger models compared to trained routing models. 
\citet{chuang2025confidentseekstrongerexploring} investigates a set of confidence estimation methods for model routing. While prior work studies model cascades that route queries across different models, we study an analogous cascade within a single hybrid-reasoning model. 

\section{Confidence-Gated Chain-of-Thought}
We evaluate confidence-gated CoT, where a model selectively triggers reasoning based on its confidence estimate. Each query is first answered directly. If the confidence score is low, the model re-runs the query with CoT enabled. 

\subsection{Problem Definition}
More specifically, for each input $x_i$, a model, parametrised by $\theta$, generates a direct answer and a confidence score $s(x_i; \theta )$. If the score is above a threshold $\tau$, the direct answer is accepted; otherwise, the model answers the question with CoT enabled: 
\[
\resizebox{0.4\textwidth}{!}{%
$
gate(x_i; \tau, \theta) =
\begin{cases}
\textsc{CoT}(x_i; \theta), & s(x_i; \theta) < \tau \\
\textsc{Direct}(x_i; \theta), & s(x_i; \theta) \geq \tau
\end{cases}
$
}
\] 
This differs from early-exit methods, which require generating partial reasoning before deciding to stop \citep{yang2025dynamicearlyexitreasoning}. In this formulation, reasoning is skipped entirely when the confidence in the direct answer is sufficient. These two approaches are complementary since confidence gating selects when to trigger reasoning, and early exiting can still be applied once CoT has been selected. 

\paragraph{CoT:} The model generates an intermediate reasoning trace before emitting a final answer \cite{wei_cot,deepseekr1report, qwen3technicalreport}. 

\paragraph{Direct:} The model is instructed to output only the final answer without generating intermediate reasoning. To achieve this, we append a concise instruction such as ``\textit{Answer:}'' to the prompt, which elicits a short response with no CoT or explanation.

\subsection{Confidence Estimation Methods}
\label{sec:confidence_methods}
We focus on confidence estimates that are produced by the model itself or based on its outputs without using an external predictor. All methods can be implemented without sampling answers multiple times or additional training. 

\paragraph{Perplexity:} 
The perplexity of a generated answer is a measure of the LLM's confidence in it \citep{lu2025prolongedreasoningneedcertaintybased}. Given a direct answer sequence \(y = (y_1,\dots,y_T)\) with $T$ tokens, it is defined as: 
\[
\resizebox{0.45\textwidth}{!}{%
$\text{PPL}(y \mid x_i) = \exp\!\left(-\frac{1}{T} \sum_{t=1}^T \log p(y_t \mid y_{<t}, x_i)\right)$.
}
\]

\paragraph{$P($True$)$~\citep{kadavath2022languagemodelsmostlyknow}:} We first generate an answer via direct prompting. Then, we ask the LLM whether the generated answer is \textit{(A) True} or \textit{(B) False} in a second forward pass. We then extract the probability of generating the token ``A''. Full prompt details are found in Appendix~\ref{app:prompts_datasets}.

\paragraph{Margin Sampling:}
This measures the difference in the probabilities between the most likely and second most likely predictions produced by the model for a given input. Margin sampling has been used for model cascades \citep{titov_optimising_calls_margin}. 

\paragraph{Verbalised Confidence:} This approach prompts off-the-shelf LLMs to self-evaluate and express its confidence as part of its response~\citep{yang2024verbalizedconfidencescoresllms}. Following prior work \citep{yang2024verbalizedconfidencescoresllms, tian-etal-2023-just-ask}, we ask the model to output a confidence score between 0.0 and 1.0 after its answer, which has shown to provide good calibration. Full prompt details are found in Appendix~\ref{app:prompts_datasets}.

\section{Evaluation Framework}
We introduce a new evaluation framework for evaluating confidence-based CoT gating. This framework measures how effectively confidence signals balance accuracy and reasoning cost under different practical constraints. The framework consists of two components: (1) a budget-based evaluation that measures performance under constrained CoT usage, and (2) a Pareto-optimal analysis which identifies how well confidence can minimise token use while preserving full CoT accuracy.

\subsection{CoT Budget}
\label{sec:budgets_pareto_method}
First, we evaluate accuracy and inference cost while explicitly limiting how frequently the model is allowed to use CoT using a predefined budget. We define the \textit{CoT budget} as the fraction of input queries for which the model is permitted to generate a CoT response, with all remaining queries answered directly. To set these thresholds, we sweep percentiles of the confidence score distribution, which allocates a fixed fraction of queries to be routed to CoT. This allows us to trace accuracy-efficiency trade-offs across budgets, plotting accuracy against average token cost or CoT usage. As we use percentile thresholds, standard post-hoc calibration methods are not applicable, as they adjust confidence values but preserve their ranking.\footnote{These methods would not affect the outcome of experiments since the decision based on percentile thresholds only depends on the ranking of queries.}

\paragraph{Identifying Budget Thresholds}
\label{sec:pareto_framework}
We consider both offline and online settings for obtaining percentile thresholds. In the \textit{offline setting}, all direct answers and confidence scores are computed first, giving access to the full distribution of confidence scores before any decision about using CoT is made. This allows thresholds to be set exactly at chosen percentiles. 
In the \textit{online setting}, queries arrive sequentially, so thresholds must be decided on the fly without access to the overall confidence score distribution. To do this, the dynamic percentile method introduced by \citet{titov_optimising_calls_margin} is used. After each query $t$, the threshold $\tau_t$ is set to the $p$-th percentile of $\{s(x_1), \dots, s(x_{t-1})\}$. We randomise dataset order and use a short warm-up phase (the first 20 queries answered directly) to initialise the observations, and report the mean and standard deviation over 10 runs.

\subsection{Pareto-Optimal Thresholds}
CoT budget-based curves characterise the full range of accuracy–cost trade-offs. However, practically we often need to select a single threshold. Therefore, we propose an analysis that evaluates whether Pareto-optimal gating thresholds can be derived from confidence scores. A threshold is Pareto-optimal if no alternative threshold achieves equal or higher accuracy at a lower token cost. The set of such thresholds forms the Pareto front, which traces the best accuracy-cost trade-offs. From this front, we derive the threshold $\tau^*$ with the lowest token cost whose accuracy remains within a tolerance $\epsilon$ of always using CoT:  
\[ 
\resizebox{0.42\textwidth}{!}{%
  $\tau^* = \arg\min_{\tau} \text{Tok}(\tau) \quad 
\text{s.t. } \text{Acc}(\tau) \geq \text{Acc}_{\text{All-CoT}} - \epsilon$.
}
\]
\noindent To reflect realistic deployment, $\tau$ is estimated using a calibration set. Percentile thresholds are swept on this set to construct the Pareto front, after which the selected threshold is applied to the test set. To account for variability in calibration splits sampling, we repeat this process using Monte Carlo cross-validation \citep{XU20011mccv}, reporting the mean and standard deviation of accuracy and token usage across runs, \textit{testing if each confidence signal can preserve accuracy while reducing cost.}

\section{Experimental Setup}

\subsection{Models}
We use three open-weight models: Qwen3 (8B/32B) \citep{qwen3technicalreport} and GPT-OSS-20B \citep{openai2025gptoss120bgptoss20bmodel}, plus the closed-weight GPT-5.1 \citep{openai_gpt51_2025}. They all support both direct answering and explicit reasoning modes with varying levels of effort. Qwen3 provides non-thinking and thinking modes. GPT-OSS supports three reasoning effort levels (\emph{low}, \emph{medium}, \emph{high}), which control the length of the generated CoT via the prompt. Unless otherwise stated, GPT-OSS results use the \emph{medium} setting, with results for other effort levels reported in Appendix~\ref{app:gpt-oss-full-results}. GPT-5.1 similarly allows switching between direct generation and multiple reasoning effort levels.

\subsection{Datasets}
We include seven benchmarks (details in Table \ref{tab:dataset_stats}, Appendix~\ref{app:prompts_datasets}) from four reasoning types following \citet{sprague2025to}: (1) \textit{commonsense reasoning} including CommonsenseQA (CSQA) \citep{talmor-etal-2019-commonsenseqa} and StrategyQA \citep{geva-etal-2021-aristotle}; (2) \textit{knowledge-based reasoning} using MMLU-redux \citep{gema-etal-2025-done}; (3) \textit{mathematical and scientific reasoning} on GPQA \citep{rein2024gpqa} and GSM8k \citep{cobbe2021trainingverifierssolvemath}; and (4) \textit{soft reasoning} using LSAT-AGI \citep{zhong-etal-2024-agieval} and MUSR \citep{sprague2024musr}. This diverse range of reasoning types allows us to test tasks where CoT has shown different levels of effectiveness.

\subsection{Confidence Baselines}

\paragraph{Expected Random Baseline.} 
For a CoT budget $r \in [0,1]$, we compute the expected accuracy and token cost as a weighted average of direct and CoT performance, with weights $1-r$ and $r$, respectively. We compute these analytically rather than via random sampling, yielding a stable baseline.

\paragraph{Oracle.} 
We also include an oracle method that triggers CoT whenever the direct answer is incorrect, acting as a perfect predictor of correctness. It represents the maximum performance that any confidence signal could achieve, serving as an upper bound of confidence-guided CoT gating.
\begin{figure}[h]
    \centering
    \includegraphics[width=0.98\linewidth]{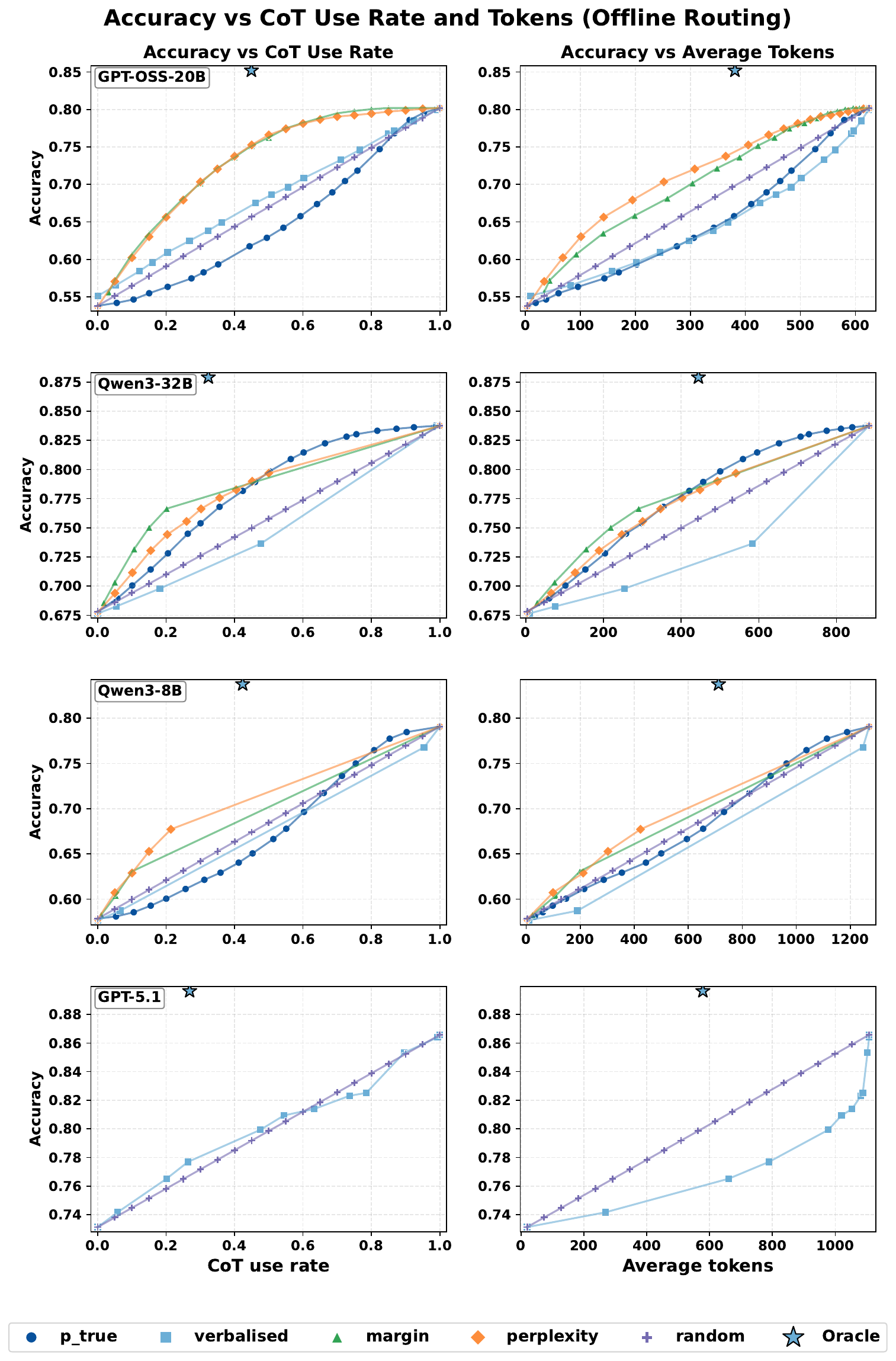}
\caption{\textbf{Offline accuracy–efficiency trade-offs under percentile budgets.}
Accuracy vs.\ CoT usage (left) and average tokens (right), aggregated over all datasets for all models. 
Curves show confidence signals vs.\ the random baseline; stars denote the oracle.}
    \label{fig:offline_accuracy_efficiency_tradeoffs}
\end{figure}
\section{Results}
\subsection{Accuracy–Efficiency Trade-offs}
We evaluate accuracy–efficiency curves using percentile budgets as defined in \S\ref{sec:budgets_pareto_method}. At each budget level, we report both accuracy and average token usage. Figure~\ref{fig:offline_accuracy_efficiency_tradeoffs} shows aggregate results for GPT-OSS-20B, Qwen3-32B, Qwen3-8B, and GPT-5.1 comparing confidence-based gating against random selection and the oracle. For GPT-OSS-20B and Qwen3-32B, several confidence methods clearly outperform random gating. In particular, \textit{margin} and \textit{perplexity} are most effective for GPT-OSS-20B, while \textit{$P(\text{True})$} is best for Qwen3-32B. Using these signals, both models match the accuracy of always using CoT while reducing CoT usage by roughly 30–40\%, saving an average of 70–100 tokens per query. In contrast, no confidence method consistently outperforms random gating for Qwen3-8B or GPT-5.1 across all budgets. Finally, the oracle performance indicates substantial headroom, e.g., GPT-OSS-20B could obtain higher accuracy while invoking CoT on fewer than half of the queries. \textit{In summary, while models like GPT-OSS-20B and Qwen3-32B achieve token savings that outperform random gating, a gap remains compared to oracle performance.}
\begin{figure}[t]
    \centering
    \includegraphics[width=0.90\linewidth]{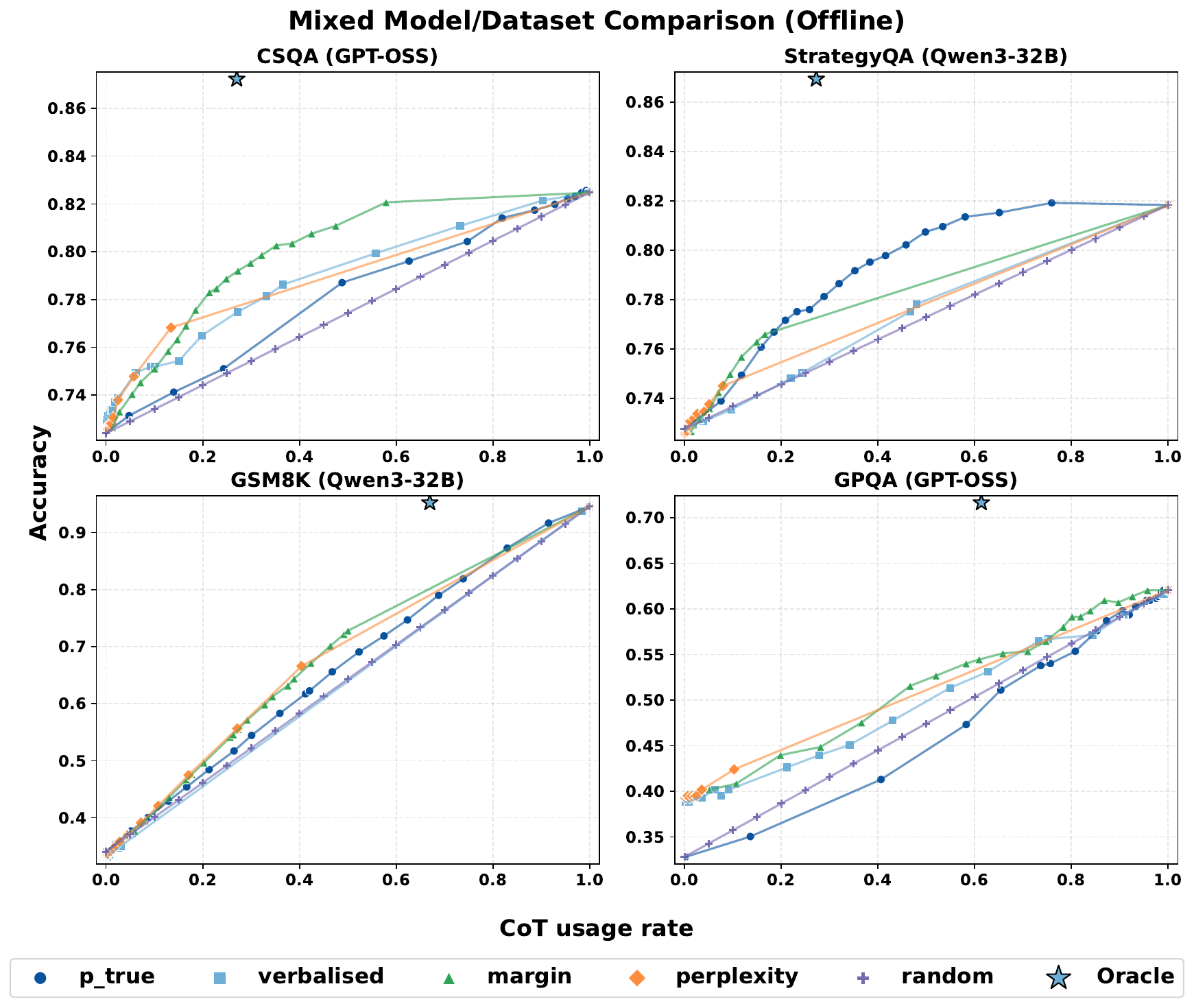}
\caption{\textbf{Task-level accuracy–efficiency trade-offs} in CSQA, StrategyQA, GSM8K and GPQA, comparing confidence-gating to random and oracle across models.}    \label{fig:task_comparison}
\end{figure}
\paragraph{Model-Specific Confidence Effectiveness.}
Figure~\ref{fig:offline_accuracy_efficiency_tradeoffs} shows the effectiveness of confidence signals across models. For GPT-OSS-20B, \textit{margin} and \textit{perplexity} outperform random gating in all budgets. In contrast, both \textit{$P(\text{True})$} and \textit{verbalised} confidence often perform worse than random. For Qwen3-32B, \textit{$P(\text{True})$} is the most effective method, outperforming other signals across a wide range of budgets. \textit{Margin} and \textit{perplexity} achieve above-random performance only at low budgets, but quickly saturate: the distributions collapse to narrow ranges, limiting their separating power. This can be seen at higher budgets where margin and perplexity scores can no longer be meaningfully separated because many examples collapse to the same value (1.0 for both). As the threshold increases with the higher CoT-use budget, these identical scores cannot be distinguished. Beyond a certain point, all thresholds yield the same gating behaviour. This results in fewer distinct points on the trade-off plots, as seen in Figure \ref{fig:offline_accuracy_efficiency_tradeoffs} for Qwen-8B. Finally, for Qwen3-8B and GPT-5.1, no method consistently outperforms random gating across all budgets. \textit{These results highlight that the utility of specific confidence signals is model-dependent, and that score saturation in methods like margin, perplexity and verbalisation can limit the granularity of confidence-gating.}

\paragraph{Commonsense, soft reasoning, and knowledge tasks benefit the most from confidence-based gating.} On tasks such as MMLU, StrategyQA, and MUSR, confidence-gated CoT enables both GPT-OSS and Qwen3-32B to match the accuracy of always using CoT while reducing token usage by 30–50\%. Figure~\ref{fig:task_comparison} shows representative examples. However, again, the oracle performance highlights room for improvement, using up to 75\% less CoT for CSQA and StrategyQA. In contrast, mathematical and scientific tasks show limited benefit. For GSM8K, direct answering without CoT has low accuracy, making it difficult to save tokens without hurting performance, which is also reflected by the oracle. Similarly, on GPQA, certain confidence methods (e.g., \textit{perplexity} for GPT-OSS-20B) outperform random gating, but efficiency gains remain limited. While the oracle highlights potential improvement, current models lack sufficient discriminative ability on these challenging questions. Full results are in Appendix~\ref{app:all_dataset-tradeoffs}. \textit{Overall, confidence-gating is most effective for tasks where high accuracy is possible with direct answer.}

\subsection{When Does Confidence Gating Work?}
To understand why CoT gating succeeds in some settings and fails in others, we examine the conditions that determine its utility. We find that performance is primarily driven by two factors: the discriminative power of the confidence signal and the impact of selection bias on the total token cost.

\begin{figure}
  \centering
  \includegraphics[width=0.91\linewidth]{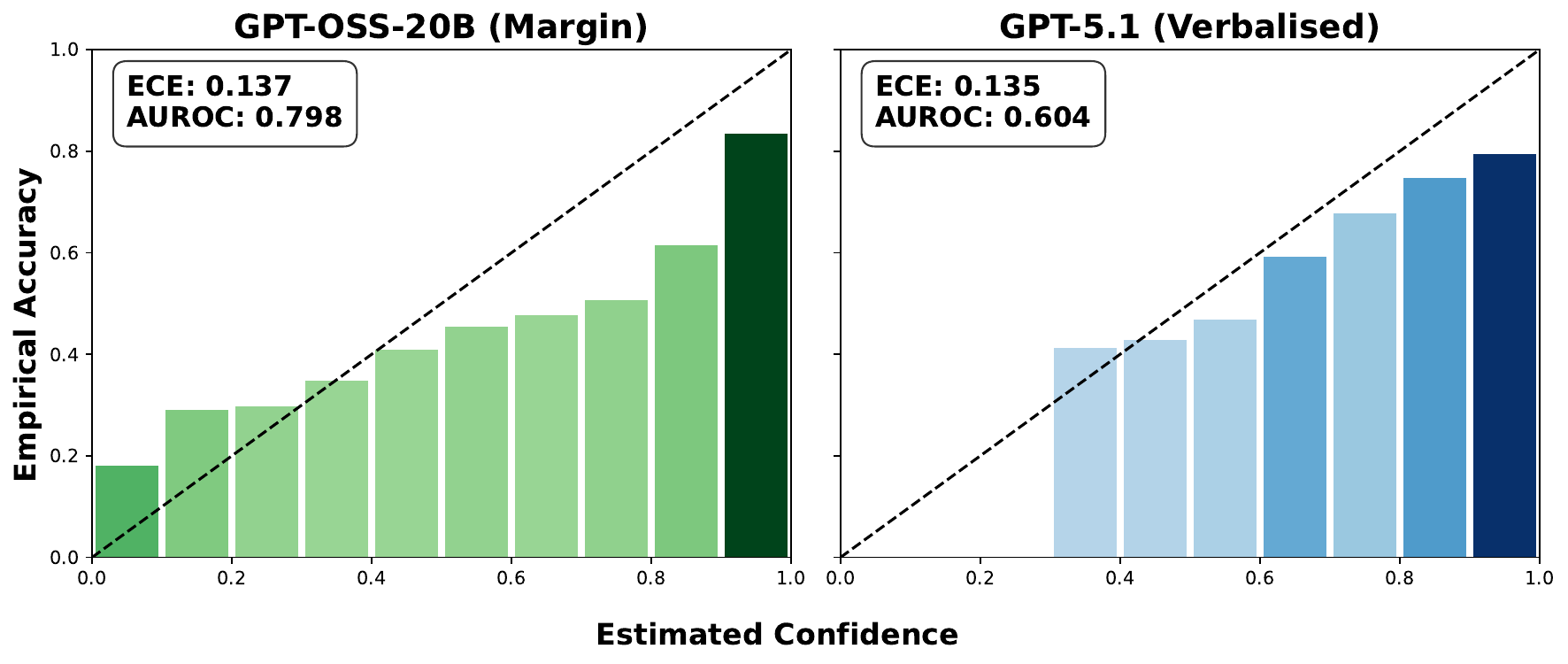}
    \caption{Reliability diagrams for GPT-OSS-20B (margin) and GPT-5.1 (verbalised). Both show similar ECE, whereas GPT-OSS has better AUROC.}
  \label{fig:rel-gptoss-margin-main}
\end{figure}

\paragraph{Scale, Calibration, and Discriminative Power.}
We study confidence calibration and discriminative power using expected calibration error (ECE) \citep{guo_calibration, pavlovic2025understandingmodelcalibration} and AUROC \citep{kadavath2022languagemodelsmostlyknow}. Although ECE denotes the quality of confidence estimates, gating depends on discriminative power, since decisions rely on ranking rather than absolute confidence. AUROC directly measures this ability. For example, GPT-5.1 (verbalised confidence) and GPT-OSS (margin) share nearly identical ECE scores, shown in Figure \ref{fig:rel-gptoss-margin-main}. However, GPT-OSS-20B achieves a higher AUROC and therefore substantially better gating performance. More broadly, GPT-OSS-20B and Qwen3-32B attain higher AUROC across confidence methods than Qwen3-8B, indicating that larger models more reliably separate correct from incorrect predictions, consistent with prior findings \citep{kadavath2022languagemodelsmostlyknow}. Full results can be found in Appendix~\ref{sec:reliability-diagrams}. Although Qwen3-8B benefits most from effective gating due to its longer CoT, its weaker discriminative power limits these gains. At the same time, larger models are not uniformly reliable, with some confidence methods performing close, or worse, than random, such as $P(\text{True})$ for GPT-OSS (AUROC = 0.439) and verbalised confidence for Qwen3-32B (AUROC = 0.55). \textit{In general, AUROC is a more reliable predictor of gating success than ECE, with larger models generally yielding higher discriminative power across logit-based signals like $P(\text{True})$, margin, and perplexity.}

\paragraph{Selection Bias of Hard Queries.} We observe differences between CoT usage rates and actual token savings. For example, while confidence gating for GPT-5.1 shows a similar trade-off to random selection in terms of CoT use rate, it leads to worse tradeoffs when considering average token usage. This is due to a selection bias introduced by confidence gating. We find that confidence scores are negatively correlated with CoT length (-0.51), meaning the low-confidence questions, which trigger CoT, are the most token-intensive to answer. Consequently, the average cost of a CoT call under confidence gating is higher than under random gating, which averages across all responses. For confidence gating to generate token savings, the reduction in CoT frequency must be sufficient to offset this increased cost per call, a threshold achieved by Qwen3-32B ($P(\text{True})$) and GPT-OSS-20B (Margin/Perplexity) (Figure \ref{fig:offline_accuracy_efficiency_tradeoffs}), but not by GPT-5.1. \textit{These results highlight that lower CoT usage does not guarantee efficiency; if confidence and length are negatively correlated, longer CoT responses can offset the gains from skipping shorter ones.}
\begin{table}[t]
\centering
\small
\setlength{\tabcolsep}{2pt}
\renewcommand{\arraystretch}{1.0}
\resizebox{0.37\textwidth}{!}{%
\begin{tabular}{@{}p{6mm}lcccc@{}}
\toprule
 & \textbf{Method} & Acc. $\uparrow$ & $\Delta$Acc $\uparrow$ & CoT (\%) $\downarrow$ & Tok. saved $\uparrow$ \\
\midrule
\multirow{7}{6mm}{\rotatebox{90}{\textbf{GPT-OSS-20B}}}
 & All CoT & 79.9 & 0.0 & 100.0 & 0.0 \\
 & All Direct & 54.1 & -25.9 & 0.0 & 483.3 \\
 \cmidrule(lr){2-6}
 & $P($True$)$ & 79.2 $_{\scriptsize 0.5}$ & -0.7 & 95.5 $_{\scriptsize 2.3}$ & 15.3 $_{\scriptsize 8.9}$ \\
 & Verbalised & 79.7 $_{\scriptsize 0.1}$ & -0.2 & 99.2 $_{\scriptsize 0.0}$ & 1.1 $_{\scriptsize 3.1}$ \\
 & Margin & 79.1 $_{\scriptsize 0.4}$ & -0.8 & \textbf{68.1}  $_{\scriptsize 3.8}$ & \textbf{65.0}  $_{\scriptsize 12.1}$ \\
 & Perplexity & 78.9 $_{\scriptsize 0.5}$ & -1.0 & \underline{70.6}  $_{\scriptsize 7.9}$ & \underline{65.7}  $_{\scriptsize 22.0}$ \\
 \cmidrule(lr){2-6}
 & Oracle & 85.0 & +5.1 & 45.9 & 187.2 \\
\midrule
\multirow{7}{6mm}{\rotatebox{90}{\textbf{Qwen3-32B}}}
 & All CoT & 83.8 & 0.0 & 100.0 & 0.0 \\
 & All Direct & 67.8 & -16.0 & 0.0 & 878.6 \\
 \cmidrule(lr){2-6}
 & $P($True$)$ & 82.8 $_{\scriptsize 0.5}$ & -1.0 & \textbf{73.8}  $_{\scriptsize 5.6}$ & \textbf{170.9}  $_{\scriptsize 45.1}$ \\
 & Verbalised & 83.7 $_{\scriptsize 0.1}$ & -0.1 & 98.9 $_{\scriptsize 0.0}$ & 3.8 $_{\scriptsize 4.1}$ \\
 & Margin & 83.8 $_{\scriptsize 0.1}$ & 0.0 & 100.0 $_{\scriptsize 0.0}$ & 0.0 $_{\scriptsize 4.1}$ \\
 & Perplexity & 83.8 $_{\scriptsize 0.1}$ & 0.0 & 100.0 $_{\scriptsize 0.0}$ & 0.0 $_{\scriptsize 4.1}$ \\
 \cmidrule(lr){2-6}
 & Oracle & 87.9 & +4.1 & 32.2 & 446.7 \\
\midrule
\multirow{7}{6mm}{\rotatebox{90}{\textbf{Qwen3-8B}}}
 & All CoT & 79.1 & 0.0 & 100.0 & 0.0 \\
 & All Direct & 57.8 & -21.3 & 0.0 & 1265.1 \\
 \cmidrule(lr){2-6}
 & $P($True$)$ & 78.4 $_{\scriptsize 0.5}$ & -0.7 & \textbf{90.8}  $_{\scriptsize 4.9}$ & \textbf{86.6}  $_{\scriptsize 54.9}$ \\
 & Verbalised & 79.0 $_{\scriptsize 0.3}$ & -0.1 & 100.0 $_{\scriptsize 0.5}$ & 0.2 $_{\scriptsize 6.2}$ \\
 & Margin & 79.1 $_{\scriptsize 0.2}$ & 0.0 & 100.0 $_{\scriptsize 0.0}$ & 0.0 $_{\scriptsize 5.6}$ \\
 & Perplexity & 79.1 $_{\scriptsize 0.2}$ & 0.0 & 100.0 $_{\scriptsize 0.0}$ & 0.0 $_{\scriptsize 5.6}$ \\
 \cmidrule(lr){2-6}
 & Oracle & 83.8 & +4.0 & 42.2 & 563.8 \\
\midrule
\multirow{4}{6mm}{\rotatebox{90}{\textbf{GPT-5.1}}}
 & All CoT & 87.3 & 0.0 & 100.0 & 0.0 \\
 & All Direct & 70.8 & -16.5 & 0.0 & 1326.8 \\
 \cmidrule(lr){2-6}
 & Verbalised & 86.8 $_{\scriptsize 0.5}$ & -0.4 & 97.8 $_{\scriptsize 3.4}$ & 1.1 $_{\scriptsize 14.2}$ \\
  \cmidrule(lr){2-6}
 & Oracle & 89.9 & +2.6 & 29.2 & 629.2 \\
\bottomrule
\end{tabular}
}
\caption{Performance with Pareto-optimal thresholds ($\epsilon=1\%$) across datasets, 
together with CoT usage and tokens saved per query.
}
\label{tab:calibration_compact_stacked}
\end{table}
\subsection{Performance in Practical Settings}
In practice, models must gate CoT without access to the full test distribution. We evaluate online budgeting to see if thresholds can be estimated dynamically on-the-fly, and Pareto-optimal selection to find a single static threshold that maximises savings without sacrificing accuracy.

\begin{figure}
    \centering
    \includegraphics[width=0.90\linewidth]{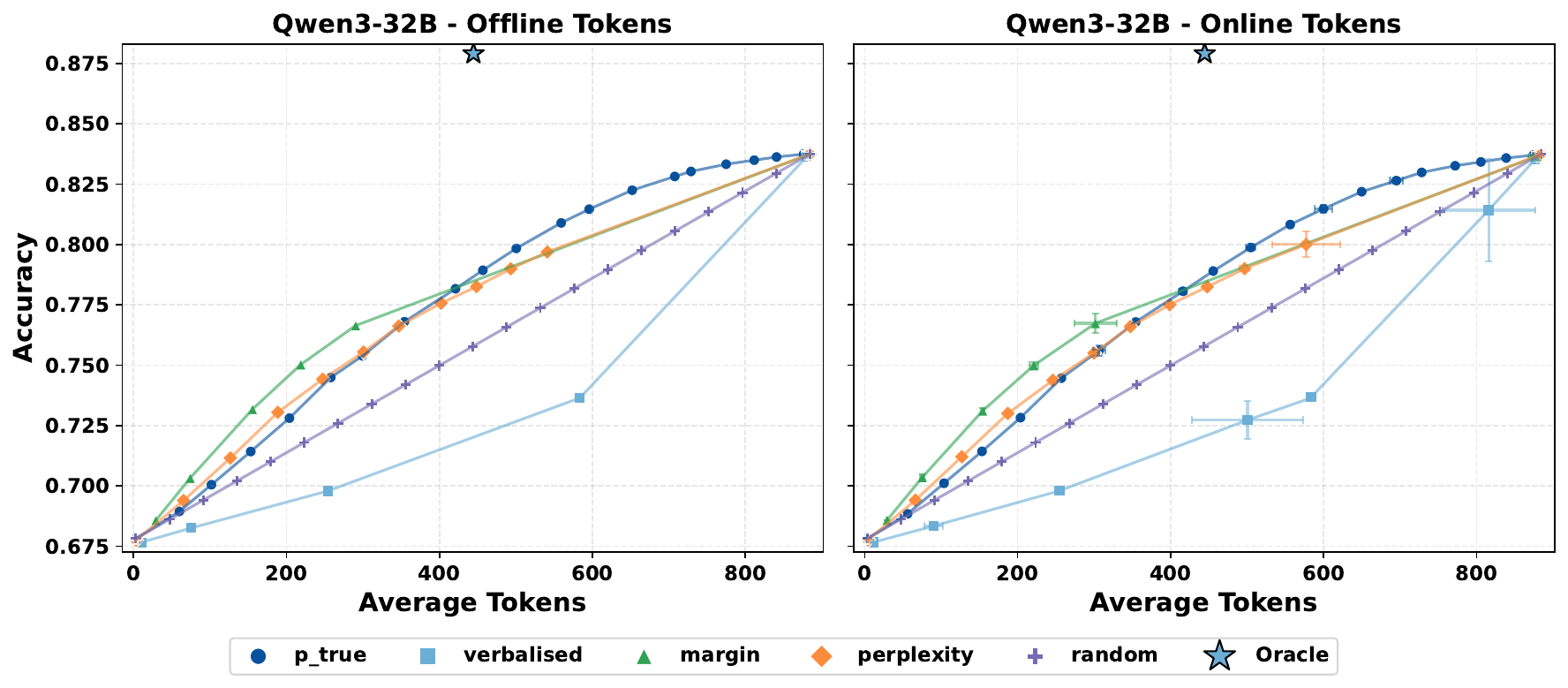}
    \caption{Online and offline budgeting for Qwen3-32B. Online performance closely tracks the offline behaviour.}
    \label{fig:online_vs_offline_qwen32b}
\end{figure}

\paragraph{Online CoT Budget–Accuracy Trade-offs}
\label{sec:deployment}
The online setting simulates queries arriving sequentially, so gating decisions must be made without access to a full predefined confidence score distribution (see \S\ref{sec:budgets_pareto_method}). Figure~\ref{fig:online_vs_offline_qwen32b} shows that for Qwen3-32B, the online approach remains stable and closely matches offline behaviour. Full results across all models are provided in Appendix~\ref{sec:online_budget_results}. GPT-OSS-20B also remains stable across budgets, closely matching offline behaviour.  In contrast, \textit{margin} and \textit{perplexity} for Qwen3-8B exhibits higher variability at mid-to-high budgets due to reduced score separability, while \textit{$P(\text{True})$} on Qwen3-32B remains stable and close to offline performance.
\textit{These results indicate that budget-based percentile thresholds can be estimated in an online setting, allowing CoT usage to be controlled without access to the full confidence distribution.} 
\paragraph{Pareto-Optimal Thresholds}
\label{sec:pareto-optimal-thresholds}
We apply the Pareto selection procedure from \S\ref{sec:budgets_pareto_method} using a 10\% calibration split, $\epsilon=1\%$, and 100 Monte Carlo repeats \citep{XU20011mccv}, reporting the mean and standard deviation of accuracy and token cost. Table~\ref{tab:calibration_compact_stacked} summarises the results.
We see that for Qwen3-8B, \textit{$P($True$)$} retains accuracy within 1\% of the full CoT baseline while reducing CoT usage by \~10\%, saving around 87 tokens per query on average. This indicates that although no method on Qwen3-8B  consistently outperforms the random baseline across the full budget sweep, confidence signals can still identify thresholds that deliver useful savings without hurting accuracy. However, this does not hold with GPT-5.1 where we only observe average savings of 1.1 tokens. The larger open-weight models show better results. Using \textit{margin} or \textit{perplexity}, GPT-OSS-20B achieves reductions of 30-35\% in CoT usage, and Qwen3-32B shows an average of 171 tokens per query using \textit{$P(\text{True})$}. \textit{In summary, we can reliably identify thresholds that reduce token usage while preserving accuracy, even in cases where the full budget trade-off curves do not consistently outperform the random baseline.}
\paragraph{OOD Calibration Set.}
To examine how well thresholds generalise, we extend our analysis beyond the mixed calibration setup. In our original experiments, both the calibration and test splits were drawn from all datasets. Here, we conduct an OOD experiment that uses each dataset as a calibration set for the others to assess how well thresholds transfer across tasks. Full results can be found in Appendix \ref{sec:OOD_pareto}. We observe that more challenging and diverse datasets, such as MMLU-Redux and GPQA, provide similar efficiency gains to the calibration split drawn from all datasets. For example, calibrating $P($True$)$ on MMLU-Redux gives a drop of only 0.8\% across all other datasets while still reducing CoT usage to 85\%. In contrast, datasets like CSQA consistently overestimate how much CoT can be removed and produce larger accuracy drops. This is because CSQA questions are significantly easier than many of the other datasets, which leads to thresholds that are too low for the more difficult datasets. \textit{Overall, this shows that calibration sets need to reflect the difficulty and diversity of the test tasks. When they do, the Pareto method transfers well, otherwise performance drops.} 

\section{Qualitative Analysis}
To better understand confidence-gated CoT behaviour, we examine examples from a maximum accuracy Pareto-optimal threshold. We categorise outcomes into: \textbf{CoT Fixed} (CoT corrects an initially wrong direct answer); \textbf{Direct} (policy saves tokens by correctly skipping CoT); \textbf{Excess CoT} (redundant reasoning for an already correct direct answer); \textbf{Missed Fix} (wrong direct answer, CoT not triggered); and \textbf{Both Fail} (neither strategy is correct). 
\begin{table}[t]
\centering
\setlength{\tabcolsep}{4pt}
\renewcommand{\arraystretch}{1.1}
\scalebox{0.75}{%
\begin{tabular}{lrrrr}
\toprule
\textbf{Category} & \textbf{Qwen8B} & \textbf{Qwen32B} & \textbf{GPT-20B} & \textbf{GPT-5.1 } \\
\midrule
CoT Fixed   & 24.7 $_{\scriptsize 0.8}$ & 18.8 $_{\scriptsize 0.6}$ & 29.9 $_{\scriptsize 0.6}$ & 15.9 $_{\scriptsize 0.6}$ \\
Direct      &  7.0 $_{\scriptsize 3.6}$ & 23.5 $_{\scriptsize 4.6}$ & 27.2 $_{\scriptsize 2.5}$ &  3.7 $_{\scriptsize 4.0}$ \\
Excess CoT  & 50.8 $_{\scriptsize 3.6}$ & 44.4 $_{\scriptsize 4.6}$ & 26.5 $_{\scriptsize 2.5}$ & 69.5 $_{\scriptsize 4.0}$ \\
Missed Fix  &  1.2 $_{\scriptsize 0.8}$ &  1.3 $_{\scriptsize 0.6}$ &  1.3 $_{\scriptsize 0.6}$ &  0.6 $_{\scriptsize 0.6}$ \\
Both fail   & 16.2 $_{\scriptsize 0.1}$ & 12.1 $_{\scriptsize 0.1}$ & 15.0 $_{\scriptsize 0.1}$ & 10.4 $_{\scriptsize 0.1}$ \\
\bottomrule
\end{tabular}%
}
\caption{Distribution of outcome types. Values denote \%, averaged over calibration runs, with standard deviation shown as subscript.}
\label{tab:distribution_outcome_types}
\end{table}

\paragraph{Outcome Breakdown.} Table~\ref{tab:distribution_outcome_types} shows the distribution of these types.
The largest share of cases for Qwen3-8B (50.8\%), Qwen3-32B (44.4\%) and GPT-5.1 (69.5\%) falls into \textit{Excess CoT}, where the direct answer is correct but the policy still uses reasoning. GPT-OSS-20B is lower at 26.5\%. The \textit{Direct} category, where the policy chooses to answer directly and its answer is correct, improves with scale, from 7.0\% (Qwen3-8B) to 23.5\% (Qwen3-32B) and 27.2\% (GPT-OSS-20B). This is expected since knowing when to answer directly requires the model to have a higher AUROC, and Qwen3-8B has a lower AUROC. \textit{CoT Fixed} accounts for 24.7\% of all queries on Qwen3-8B, 18.8\% on Qwen3-32B, and 29.9\% on GPT-OSS-20B, capturing cases where the direct answer would have been wrong but enabling CoT corrects it. \textit{Both Fail} cases where gating could not help, remain around 12--16\% across models. We present examples of some of these types below.

\paragraph{Example 1: \textit{CoT Fixed}.}
Across all models, 18–30\% of queries fall into this category. Good examples of this behaviour come from GSM8K, where direct answering often fails but CoT achieves high accuracy \citep{sprague2025to}. In Example \ref{ex:example1}, the direct answer is wrong, but the CoT reasoning solves the problem correctly. 

\newcounter{example}

\begin{tcolorbox}[float=htb, colback=green!5, colframe=green!60!black, title=Example 1]
\scriptsize
\textbf{Question:} Martha is planning her Christmas party. She invited 2 families with 6 people and 3 families with 4 people. 8 people couldn't come due to illness, and 1/4 that number had previous commitments. How many people show up for Martha's party?\\
\textbf{Direct answer:} 25 (incorrect).\\
\textbf{CoT reasoning (excerpt):} ``Okay, let me try to figure out how many people are coming to Martha's Christmas party...
Then, 1/4 of that number (2) had previous commitments...  
So total not coming = 8 + 2 = 10. Therefore, 24 - 10 = 14.  
...Yes, that seems correct. \textbf{The answer is 14.}''\\
\textbf{Ground Truth:} 14.
\refstepcounter{example}\label{ex:example1}
\end{tcolorbox}

\paragraph{Example 2: \textit{Direct}.} This represents a positive case where accuracy is preserved and tokens are saved by directly answering. Confidence-gated CoT saves 284 tokens by choosing to skip CoT. 

\begin{tcolorbox}[float=htb, colback=green!5, colframe=green!60!black, title=Example 2]
\scriptsize
\textbf{Question:} Would a Nike shoebox be too small to fit a swan in?\\
\textbf{Direct answer:} \textit{Yes} ($P(\text{True}) = 0.99$).\\
\textbf{Ground Truth:} \textit{Yes}.

\end{tcolorbox}

\paragraph{Example 3: \textit{Excess CoT}.} In this example, the direct answer was already correct, but the policy still used CoT, leading to redundant tokens.

\begin{tcolorbox}[float=htb, colback=red!5, colframe=red!60!black, title=Example 3]
\scriptsize
\textbf{Question:} Where would you put a glass after drinking from it?\\
\textbf{Answer choices:} (A) ocean, (B) water cooler, (C) cabinet, (D) dishwasher, (E) dining room.\\
\textbf{Direct answer:} \textit{(D)} ($P(\text{True}) = 0.59$).\\
\textbf{CoT reasoning (excerpt):} ``Option A doesn’t make sense… Option D, dishwasher, is correct. Therefore, the answer is D.''
\end{tcolorbox}
\newcommand{\std}[1]{\raisebox{-0.9ex}{\tiny\textpm#1}}
\noindent These examples highlight both the effectiveness and the limitations of confidence-gated CoT. The substantial portion of \textit{CoT Fixed} cases, combined with the minimal rate of \textit{Missed Fix} (approx. 1\%), confirms that confidence signals effectively identify necessary reasoning. However, the frequency of \textit{Excess CoT} shows that models often lack the confidence to skip reasoning even when correct. This analysis highlights that while current signals can be effective, further gains can be unlocked by better distinguishing correct from incorrect answers.

\section{Conclusion}
We conducted the first systematic study of confidence-gated CoT for efficient LLM reasoning. Our results show that training-free confidence signals can preserve accuracy while lowering token usage, confirming that LLMs themselves possess the ability to produce confidence signals that can make reasoning more efficient. However, these efficiency gains are not observed uniformly across all models, and a significant gap remains between current methods and Oracle performance. This highlights that while confidence gating holds promise, current confidence estimation methods lack consistency. By providing a systematic evaluation and framework to quantify these accuracy-efficiency trade-offs, our study establishes a foundation for evaluating how well different confidence signals can be leveraged to build efficient reasoning systems.

\clearpage

\section*{Limitations}
We focus on confidence estimation methods that do not require multiple samples or extra training, motivated by efficiency, and due to compute constraints. Similarly, we employ a standard prompting strategy for verbalised confidence consistent with prior work \citep{yang2024verbalizedconfidencescoresllms}, future work could extend this analysis to various prompting strategies. Additionally, future work could explore confidence estimation methods that require minimal sampling \citep{zhou2025steerconf, kuhn2023semantic} or models that are trained to produce better confidence estimates \citep{stangel2025rewardingdoubtreinforcementlearning, zhang2025reinforcementlearningbetterverbalized}, both of which can be studied within our framework.

\section*{Ethical Consideration}
Language models can generate content that is harmful \citep{risks_taxonomy}. Our contribution focuses on efficiency without training and as such will not affect the existing risks present in each model. All datasets are MIT-licensed, apart from GPQA, which is released under a CC-BY 4.0 license. We use these datasets to evaluate NLP models, which is in line with their intended purpose. To the best of our knowledge there is no PII or offensive content in these datasets. 

\bibliography{custom}

\appendix

\clearpage
\section{LLM Assistant Use}
The writing of this paper received proofreading and language polishing suggestions using LLMs. In addition, parts of our experimental code were drafted or refactored with the assistance of GitHub Copilot; all final text and code was manually reviewed and verified by the authors.

\section{Model Inference Settings}
\label{app:inference_hyperparams} For GPT-5.1 we use greedy decoding for direct answers and the default temperature of 1.0 when using the high reasoning effort. We use Hugging Face Transformers for inference on Qwen and GPT-OSS models. For Qwen models (8B and 32B), we follow the recommended decoding settings from the model cards, using a temperature of 0.6 and top-p of 0.95 to avoid degenerate repetition. For GPT-OSS-20B, we use the default sampling configuration with temperature 1.0 and top-p 1.0. In all settings, we set a maximum limit of 7000 thinking tokens and insert text that prompts the model to answer after this limit has been reached.

\section{Prompts}
\label{app:prompts_datasets}

\begin{tcolorbox}[colback=gray!5, colframe=gray!60, title=Verbalised Prompt, width=\linewidth]
\small
Please directly provide your best guess of the answer to the question and give the probability that you think it is correct (0.0 to 1.0). Take your uncertainty in the prompt, the task difficulty, your knowledge availability, and other sources of uncertainty into account. 

Give only the guess and probability, with no other words or explanation. 

\medskip
Format your final response as:\\
Answer: \textless your\_best\_guess\textgreater.\\
Probability: \textless score between 0.0 and 1.0\textgreater
\end{tcolorbox}
\begin{tcolorbox}[colback=gray!5, colframe=gray!60, title=$P($True$)$ Prompt, width=\linewidth]
\small
User:\\ Is this answer:\\(A) True \\(B) False  

\medskip
Assistant:\\ The answer is:
\end{tcolorbox}
\newpage
\section{Dataset Statistics}
\begin{table}[h]
\centering
\small
\begin{tabular}{lr}
\toprule
\textbf{Dataset} & \textbf{\# Samples} \\
\midrule
CommonsenseQA (CSQA) & 1221 \\
StrategyQA           & 2290 \\
MMLU-redux           & 3000 \\
GSM8K                & 1319 \\
GPQA                 & 448 \\
LSAT-AGI             & 1009 \\
MUSR                 & 756 \\
\bottomrule
\end{tabular}

\caption{Dataset statistics.}
\label{tab:dataset_stats}
\end{table}

\section{Online Budget Results}
\label{sec:online_budget_results}

\begin{figure}[h]
    \centering
    \includegraphics[width=\linewidth]{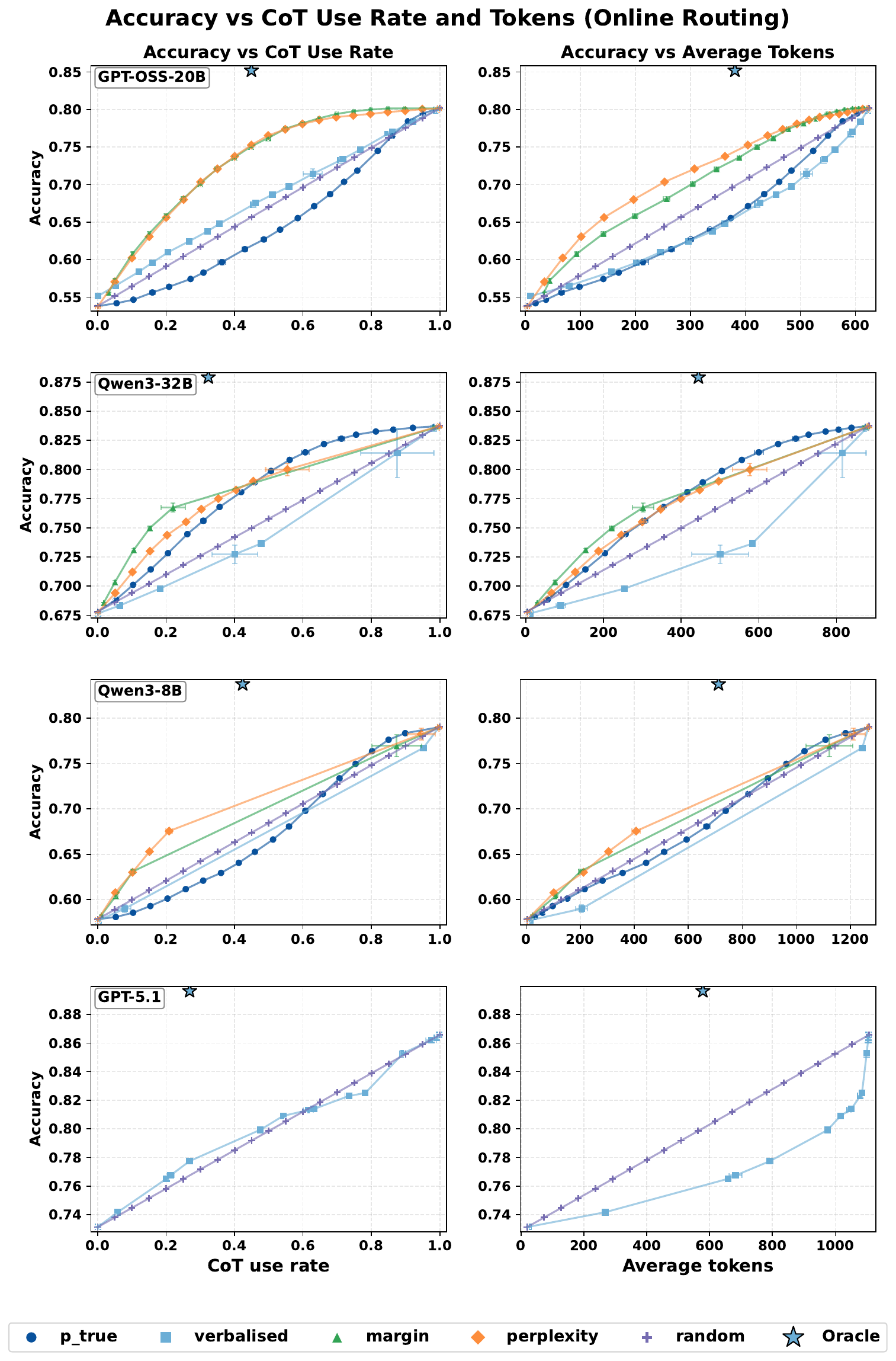}
    \caption{Online Accuracy vs. CoT use rate (left) and average tokens (right) across all datasets in the \textbf{online} setting. Stars show Oracle performance.}
    \label{fig:accuracy_efficiency_tradeoffs_online}
\end{figure}

\newpage
\section{OOD Pareto Results}
\label{sec:OOD_pareto}
\vspace{5.6em}
\begin{table}[h]
\centering
\resizebox{0.95\columnwidth}{!}{%
\label{tab:qwen32b_by_dataset}
\small
\setlength{\tabcolsep}{2.5pt}
\renewcommand{\arraystretch}{1.0}
\begin{tabular}{@{}p{8mm}lcccc@{}}
\toprule
& \textbf{Method} & Acc. $\uparrow$ & $\Delta$Acc $\uparrow$ & CoT (\%) $\downarrow$ & \makecell{Avg. Tok. \\ saved $\uparrow$}\\
\midrule
\multirow{7}{11mm}{\raisebox{-5pt}{\hspace{2pt}\rotatebox{90}{\textbf{CSQA}}}} & All CoT & 83.4 & 0.0 & 100.0 & 0.0 \\
& All Direct & 65.5 & -17.8 & 0.0 & 943.1 \\
\cmidrule(lr){2-6}
& $P($True$)$ & 70.4 & -13.0 & 19.5 & 746.7 \\
& Verbalised & 65.2 & -18.2 & 0.0 & 935.3 \\
& Margin & 83.4 & 0.0 & 100.0 & 0.0 \\
& Perplexity & 73.3 & -10.1 & 22.2 & 658.0 \\
\midrule
\multirow{7}{11mm}{\raisebox{-5pt}{\hspace{2pt}\rotatebox{90}{\textbf{GPQA}}}} & All CoT & 84.9 & 0.0 & 100.0 & 0.0 \\
& All Direct & 69.1 & -15.8 & 0.0 & 749.5 \\
\cmidrule(lr){2-6}
& $P($True$)$ & 84.4 & -0.4 & 81.2 & 105.8 \\
& Verbalised & 84.8 & -0.1 & 98.8 & 4.0 \\
& Margin & 84.9 & 0.0 & 100.0 & 0.0 \\
& Perplexity & 84.9 & 0.0 & 100.0 & 0.0 \\
\midrule
\multirow{7}{11mm}{\raisebox{-5pt}{\hspace{2pt}\rotatebox{90}{\textbf{GSM8K}}}} & All CoT & 82.1 & 0.0 & 100.0 & 0.0 \\
& All Direct & 72.9 & -9.2 & 0.0 & 919.4 \\
\cmidrule(lr){2-6}
& $P($True$)$ & 82.1 & 0.0 & 98.2 & 9.5 \\
& Verbalised & 82.1 & 0.0 & 98.9 & 3.1 \\
& Margin & 82.1 & 0.0 & 100.0 & 0.0 \\
& Perplexity & 82.1 & 0.0 & 100.0 & 0.0 \\
\midrule
\multirow{7}{11mm}{\raisebox{-5pt}{\hspace{2pt}\rotatebox{90}{\textbf{LSAT}}}} & All CoT & 83.1 & 0.0 & 100.0 & 0.0 \\
& All Direct & 67.1 & -16.0 & 0.0 & 811.1 \\
\cmidrule(lr){2-6}
& $P($True$)$ & 83.1 & 0.0 & 98.5 & 6.5 \\
& Verbalised & 83.0 & -0.1 & 98.7 & 4.3 \\
& Margin & 83.1 & 0.0 & 100.0 & 0.0 \\
& Perplexity & 83.1 & 0.0 & 100.0 & 0.0 \\
\midrule
\multirow{7}{11mm}{\raisebox{-5pt}{\hspace{2pt}\rotatebox{90}{\textbf{MMLU-REDUX}}}} & All CoT & 83.0 & 0.0 & 100.0 & 0.0 \\
& All Direct & 63.8 & -19.6 & 0.0 & 907.4 \\
\cmidrule(lr){2-6}
& $P($True$)$ & 82.2 & -0.8 & 84.8 & 116.9 \\
& Verbalised & 82.8 & -0.1 & 98.5 & 5.0 \\
& Margin & 83.0 & 0.0 & 100.0 & 0.0 \\
& Perplexity & 83.0 & 0.0 & 100.0 & 0.0 \\
\midrule
\multirow{7}{11mm}{\raisebox{-5pt}{\hspace{2pt}\rotatebox{90}{\textbf{MUSR}}}} & All CoT & 85.2 & 0.0 & 100.0 & 0.0 \\
& All Direct & 68.9 & -16.4 & 0.0 & 859.2 \\
\cmidrule(lr){2-6}
& $P($True$)$ & 80.9 & -4.3 & 47.0 & 407.4 \\
& Verbalised & 74.4 & -10.9 & 45.7 & 306.5 \\
& Margin & 85.2 & 0.0 & 100.0 & 0.0 \\
& Perplexity & 85.2 & 0.0 & 100.0 & 0.0 \\
\midrule
\multirow{7}{11mm}{\raisebox{-5pt}{\hspace{2pt}\rotatebox{90}{\textbf{STRATEGYQA}}}} & All CoT & 84.3 & 0.0 & 100.0 & 0.0 \\
& All Direct & 66.4 & -17.9 & 0.0 & 1014.3 \\
\cmidrule(lr){2-6}
& $P($True$)$ & 78.3 & -6.0 & 43.8 & 499.0 \\
& Verbalised & 84.2 & -0.1 & 99.6 & 1.9 \\
& Margin & 84.3 & 0.0 & 100.0 & 0.0 \\
& Perplexity & 84.3 & 0.0 & 100.0 & 0.0 \\
\bottomrule
\end{tabular}
}
\caption{Results for Qwen3-32B by dataset with Pareto-optimal thresholds ($\epsilon=1\%$).}
\end{table}

\begin{table}[t]
\centering
\resizebox{0.95\columnwidth}{!}{%
\label{tab:gptoss20b_by_dataset}
\small
\setlength{\tabcolsep}{2.5pt}
\renewcommand{\arraystretch}{1.0}
\begin{tabular}{@{}p{8mm}lcccc@{}}
\toprule
& \textbf{Method} & Acc. $\uparrow$ & $\Delta$Acc $\uparrow$ & CoT (\%) $\downarrow$ & \makecell{Avg. Tok. \\ saved $\uparrow$}\\
\midrule
\multirow{6}{11mm}{\raisebox{-5pt}{\hspace{2pt}\rotatebox{90}{\textbf{CSQA}}}} & All CoT & 79.7 & 0.0 & 100.0 & 0.0 \\
& All Direct & 51.2 & -28.5 & 0.0 & 514.0 \\
\cmidrule(lr){2-6}
& $P($True$)$ & 68.7 & -11.0 & 70.9 & 135.3 \\
& Verbalised & 71.9 & -7.7 & 70.8 & 59.8 \\
& Margin & 78.8 & -0.9 & 68.7 & 68.6 \\
& Perplexity & 76.9 & -2.8 & 54.7 & 126.0 \\
\midrule
\multirow{6}{11mm}{\raisebox{-5pt}{\hspace{2pt}\rotatebox{90}{\textbf{GPQA}}}} & All CoT & 80.6 & 0.0 & 100.0 & 0.0 \\
& All Direct & 54.8 & -25.9 & 0.0 & 408.3 \\
\cmidrule(lr){2-6}
& $P($True$)$ & 80.6 & -0.1 & 99.1 & 2.2 \\
& Verbalised & 80.4 & -0.3 & 99.2 & 1.2 \\
& Margin & 80.1 & -0.6 & 68.9 & 60.1 \\
& Perplexity & 77.5 & -3.2 & 48.4 & 140.6 \\
\midrule
\multirow{6}{11mm}{\raisebox{-5pt}{\hspace{2pt}\rotatebox{90}{\textbf{GSM8K}}}} & All CoT & 77.8 & 0.0 & 100.0 & 0.0 \\
& All Direct & 60.0 & -17.8 & 0.0 & 515.9 \\
\cmidrule(lr){2-6}
& $P($True$)$ & 77.6 & -0.1 & 98.7 & 3.0 \\
& Verbalised & 77.8 & -0.0 & 100.0 & 0.0 \\
& Margin & 75.8 & -2.0 & 53.6 & 108.1 \\
& Perplexity & 74.7 & -3.0 & 45.6 & 149.5 \\
\midrule
\multirow{6}{11mm}{\raisebox{-5pt}{\hspace{2pt}\rotatebox{90}{\textbf{LSAT}}}} & All CoT & 79.8 & 0.0 & 100.0 & 0.0 \\
& All Direct & 54.0 & -25.8 & 0.0 & 417.4 \\
\cmidrule(lr){2-6}
& $P($True$)$ & 77.4 & -2.4 & 89.2 & 40.2 \\
& Verbalised & 72.2 & -7.6 & 68.6 & 61.3 \\
& Margin & 79.3 & -0.5 & 69.8 & 54.6 \\
& Perplexity & 78.7 & -1.1 & 66.6 & 69.3 \\
\midrule
\multirow{6}{11mm}{\raisebox{-5pt}{\hspace{2pt}\rotatebox{90}{\textbf{MMLU-REDUX}}}} & All CoT & 78.2 & 0.0 & 100.0 & 0.0 \\
& All Direct & 49.8 & -28.4 & 0.0 & 519.2 \\
\cmidrule(lr){2-6}
& $P($True$)$ & 78.2 & 0.0 & 100.0 & 0.0 \\
& Verbalised & 73.7 & -4.5 & 84.3 & 25.4 \\
& Margin & 77.9 & -0.3 & 77.2 & 45.1 \\
& Perplexity & 77.0 & -1.2 & 67.8 & 72.5 \\
\midrule
\multirow{6}{11mm}{\raisebox{-5pt}{\hspace{2pt}\rotatebox{90}{\textbf{MUSR}}}} & All CoT & 81.4 & 0.0 & 100.0 & 0.0 \\
& All Direct & 54.1 & -27.3 & 0.0 & 480.0 \\
\cmidrule(lr){2-6}
& $P($True$)$ & 75.5 & -5.9 & 82.0 & 79.8 \\
& Verbalised & 67.8 & -13.6 & 44.1 & 150.3 \\
& Margin & 78.6 & -2.8 & 53.4 & 111.5 \\
& Perplexity & 78.8 & -2.6 & 53.3 & 116.5 \\
\midrule
\multirow{6}{11mm}{\raisebox{-5pt}{\hspace{2pt}\rotatebox{90}{\textbf{STRATEGYQA}}}} & All CoT & 81.7 & 0.0 & 100.0 & 0.0 \\
& All Direct & 51.6 & -30.1 & 0.0 & 567.9 \\
\cmidrule(lr){2-6}
& $P($True$)$ & 81.2 & -0.5 & 98.2 & 8.0 \\
& Verbalised & 79.7 & -2.0 & 94.3 & 7.6 \\
& Margin & 81.2 & -0.5 & 71.1 & 61.1 \\
& Perplexity & 81.7 & -0.1 & 86.4 & 31.0 \\
\bottomrule
\end{tabular}
}
\caption{Results for GPT-OSS-20B by dataset with Pareto-optimal thresholds ($\epsilon=1\%$).}
\end{table}

\clearpage
\section{Per Dataset Trade-off Plots}
\label{app:all_dataset-tradeoffs}
\vspace{6em}


\begin{figure}[h]
    \centering
    \begin{subfigure}{0.32\textwidth}
        \centering
        \includegraphics[width=\linewidth]{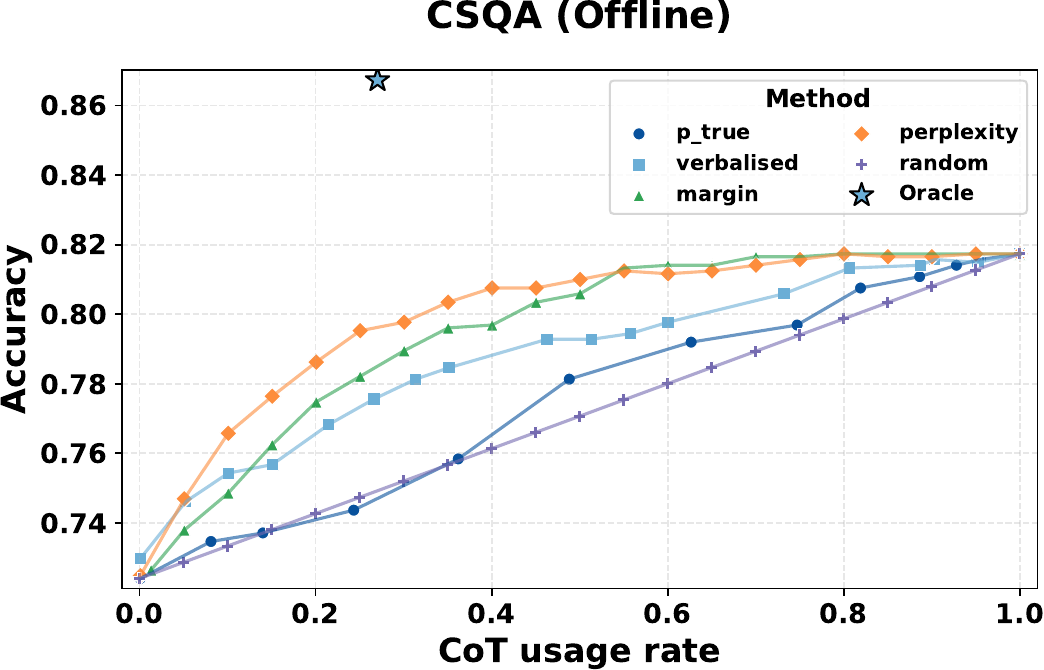}
        \caption{GPT-OSS Medium - CoT Use}
    \end{subfigure}
    \hfill
    \begin{subfigure}{0.32\textwidth}
        \centering
        \includegraphics[width=\linewidth]{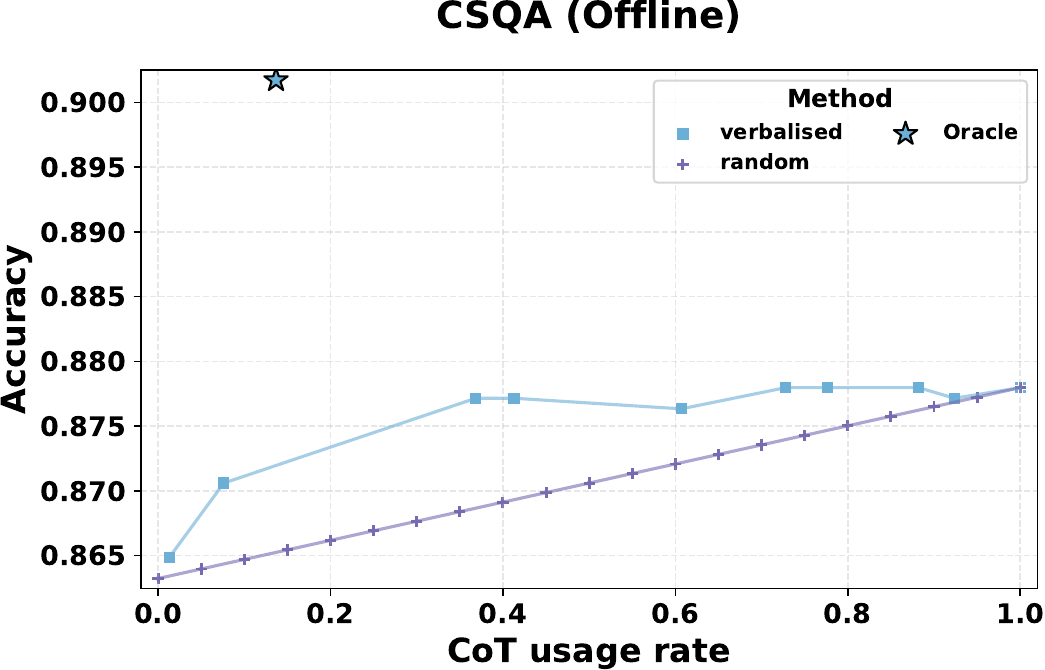}
        \caption{GPT-5 - CoT Use}
    \end{subfigure}
    \hfill
    \begin{subfigure}{0.32\textwidth}
        \mbox{}
    \end{subfigure}

    \vskip 0.5em

    \begin{subfigure}{0.32\textwidth}
        \centering
        \includegraphics[width=\linewidth]{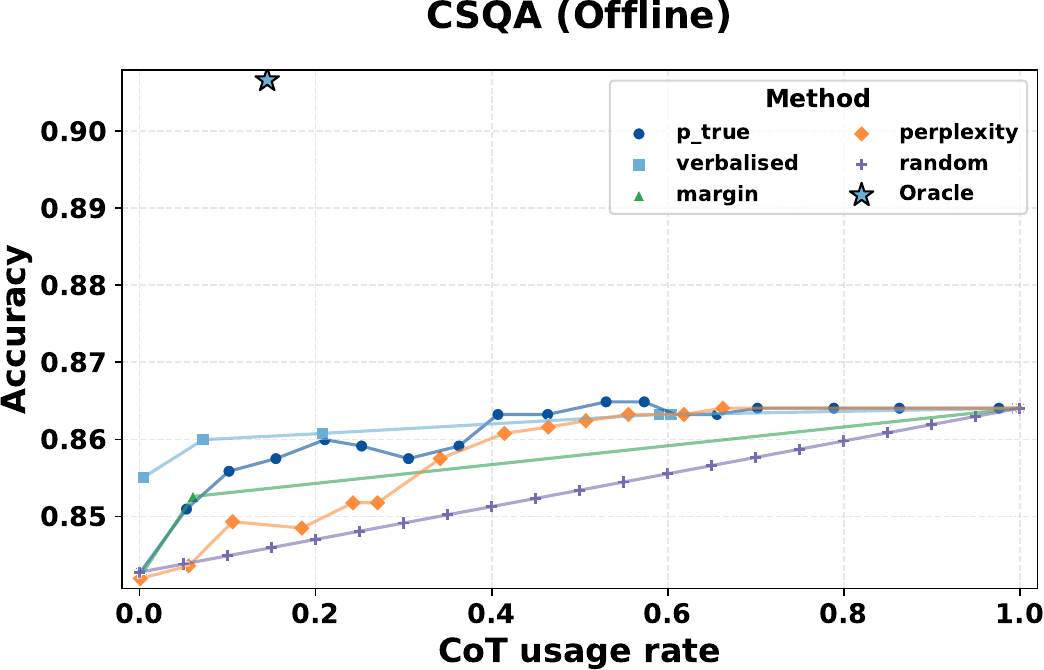}
        \caption{Qwen3-32B - CoT Use}
    \end{subfigure}
    \hfill
    \begin{subfigure}{0.32\textwidth}
        \centering
        \includegraphics[width=\linewidth]{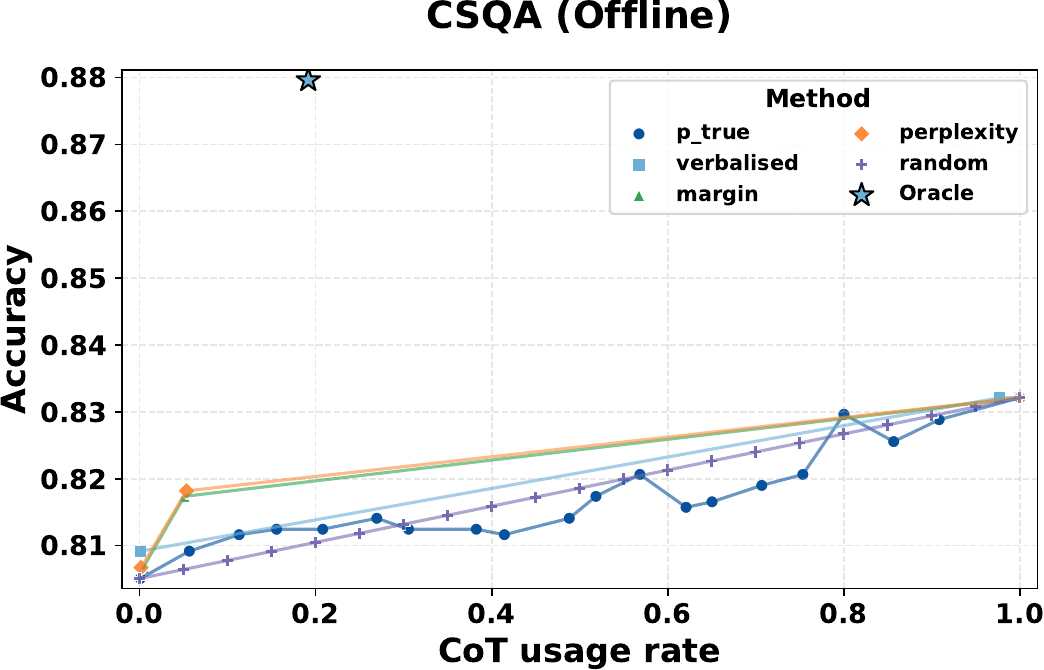}
        \caption{Qwen3-8B - CoT Use}
    \end{subfigure}
    \hfill
    \begin{subfigure}{0.32\textwidth}
        \mbox{}
    \end{subfigure}

    \caption{\textbf{CSQA (Part 1): Accuracy vs. CoT Use.}}
    
    \label{fig:csqa_cot}
\end{figure}

\begin{figure}[h]
    \centering
    \begin{subfigure}{0.32\textwidth}
        \centering
        \includegraphics[width=\linewidth]{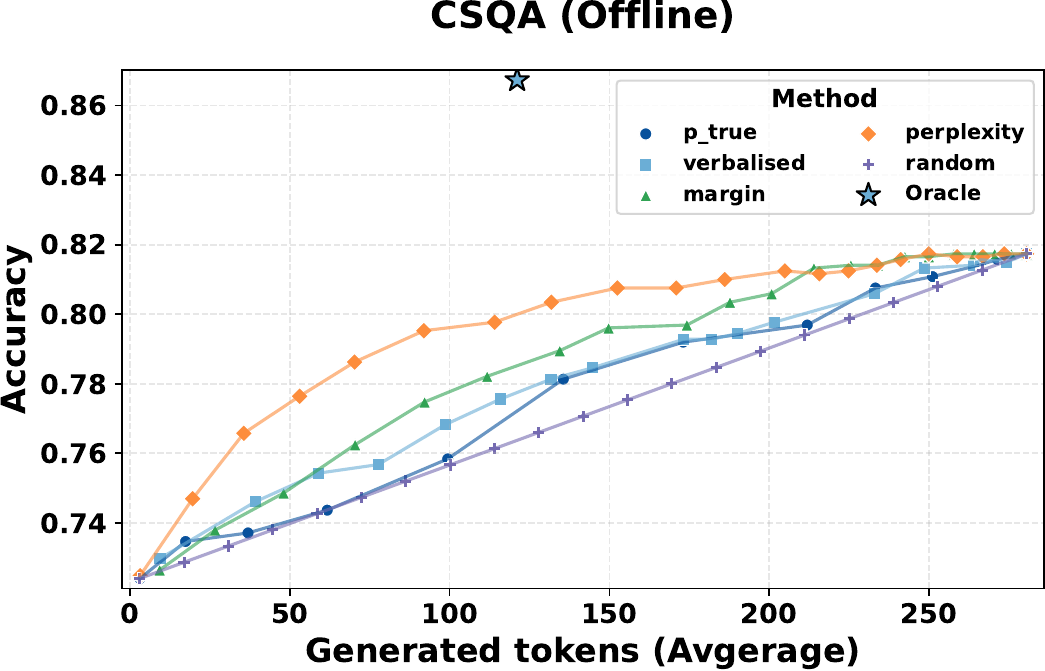}
        \caption{GPT-OSS Medium - Tokens}
    \end{subfigure}
    \hfill
    \begin{subfigure}{0.32\textwidth}
        \centering
        \includegraphics[width=\linewidth]{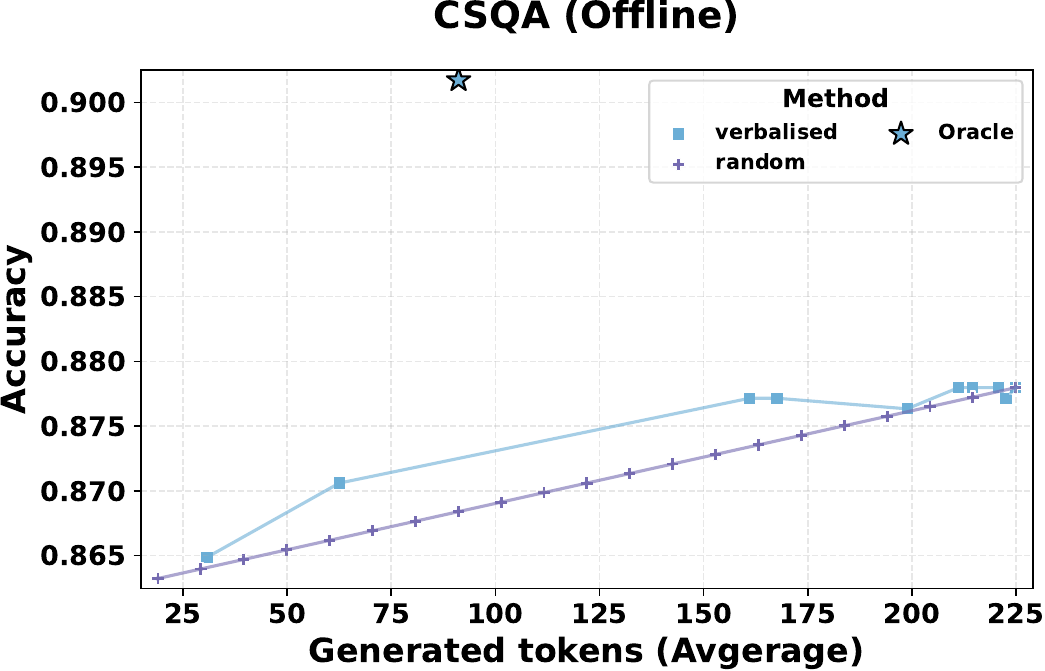}
        \caption{GPT-5 - Tokens}
    \end{subfigure}
    \hfill
    \begin{subfigure}{0.32\textwidth}
        \mbox{}
    \end{subfigure}

    \vskip 0.5em

    \begin{subfigure}{0.32\textwidth}
        \centering
        \includegraphics[width=\linewidth]{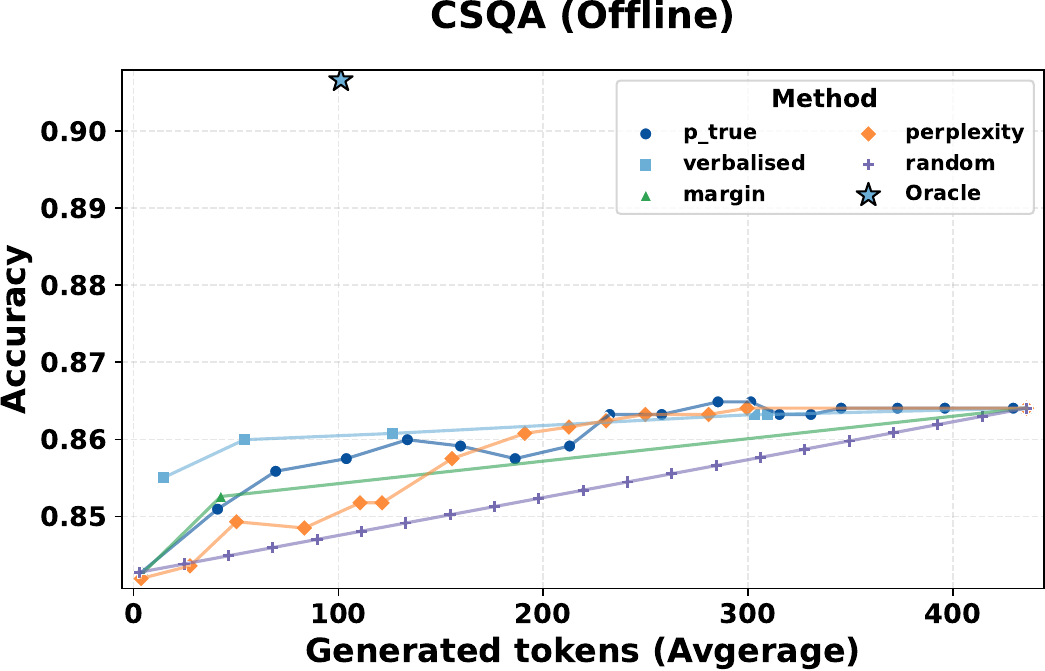}
        \caption{Qwen3-32B - Tokens}
    \end{subfigure}
    \hfill
    \begin{subfigure}{0.32\textwidth}
        \centering
        \includegraphics[width=\linewidth]{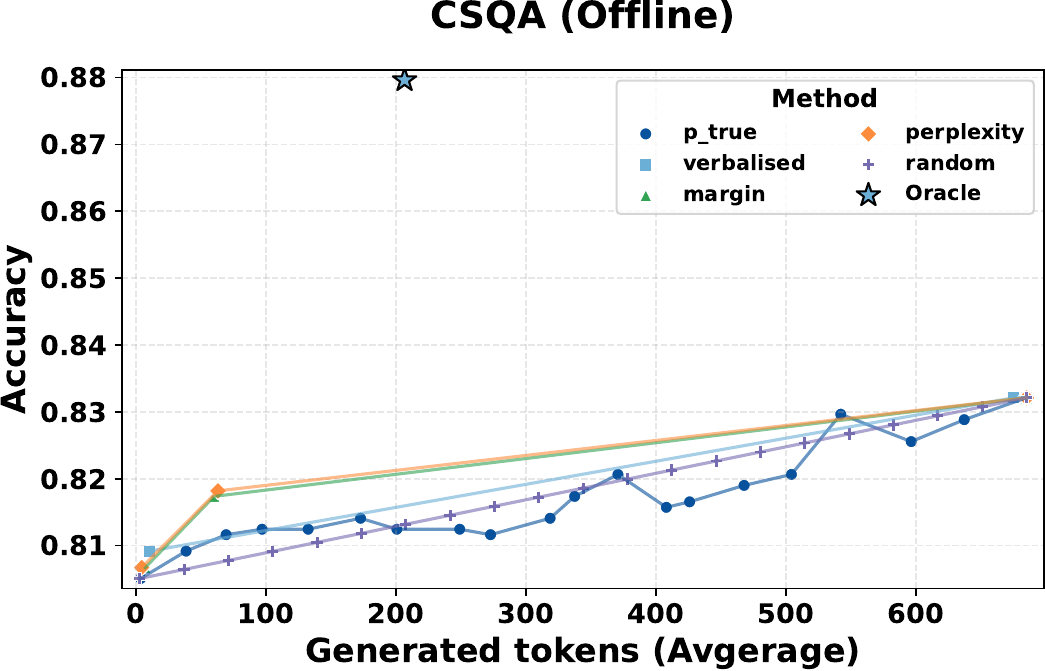}
        \caption{Qwen3-8B - Tokens}
    \end{subfigure}
    \hfill
    \begin{subfigure}{0.32\textwidth}
        \mbox{}
    \end{subfigure}

    \caption{\textbf{CSQA (Part 2): Average Tokens vs. Accuracy.}}
    \label{fig:csqa_tokens}
\end{figure}


\begin{figure}[h]
    \centering
    \begin{subfigure}{0.32\textwidth}
        \centering
        \includegraphics[width=\linewidth]{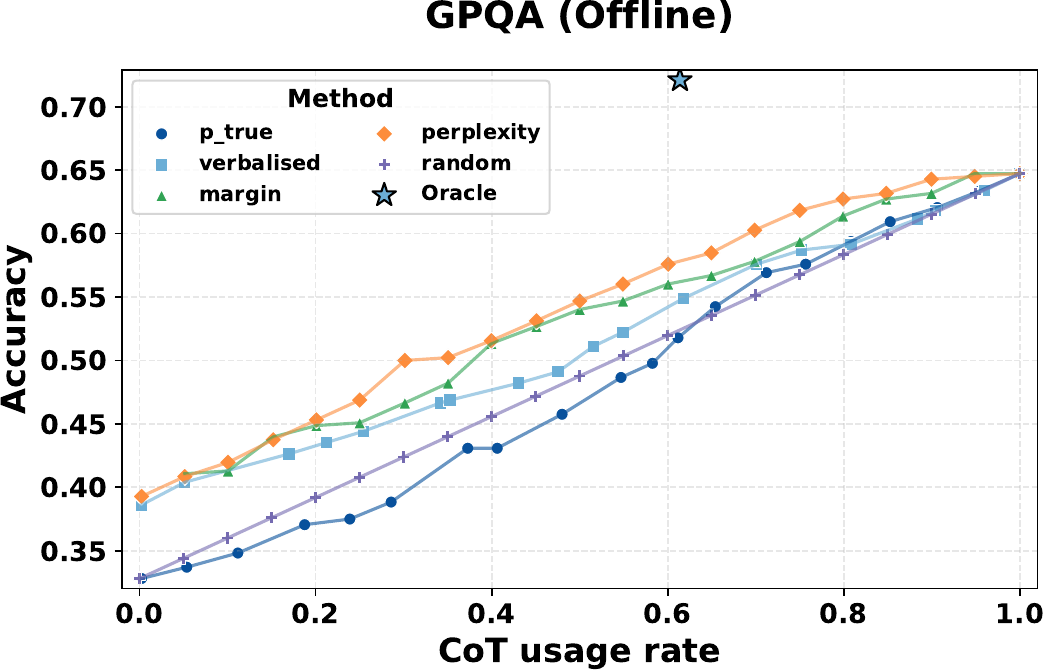}
        \caption{GPT-OSS Medium - CoT Use}
    \end{subfigure}
    \hfill
    \begin{subfigure}{0.32\textwidth}
        \centering
        \includegraphics[width=\linewidth]{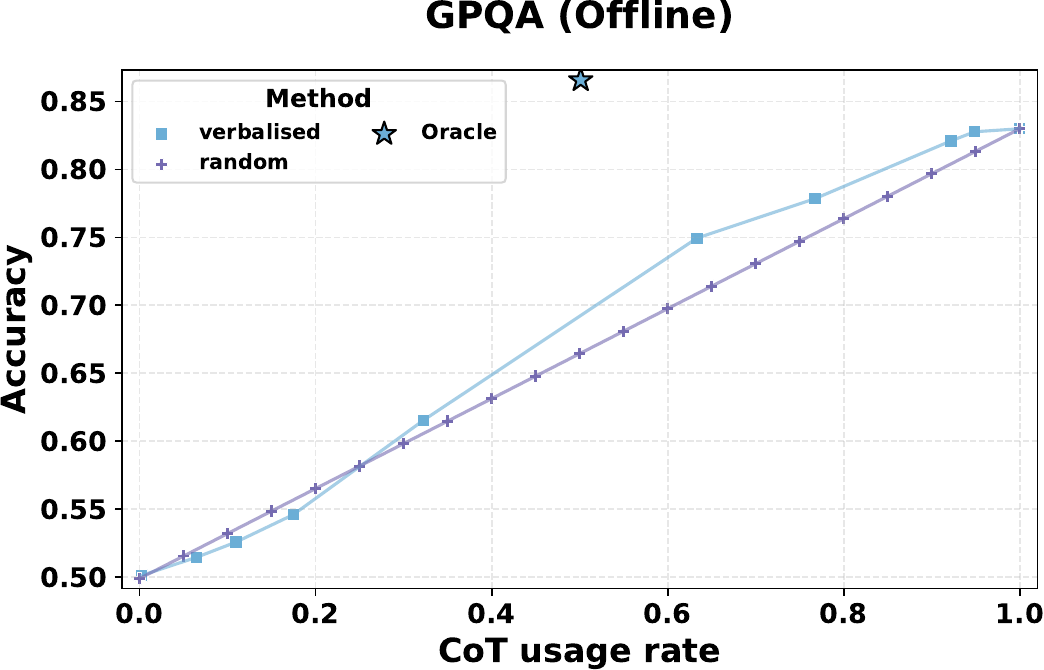}
        \caption{GPT-5 - CoT Use}
    \end{subfigure}
    \hfill
    \begin{subfigure}{0.32\textwidth}
        \mbox{}
    \end{subfigure}

    \vskip 0.5em

    \begin{subfigure}{0.32\textwidth}
        \centering
        \includegraphics[width=\linewidth]{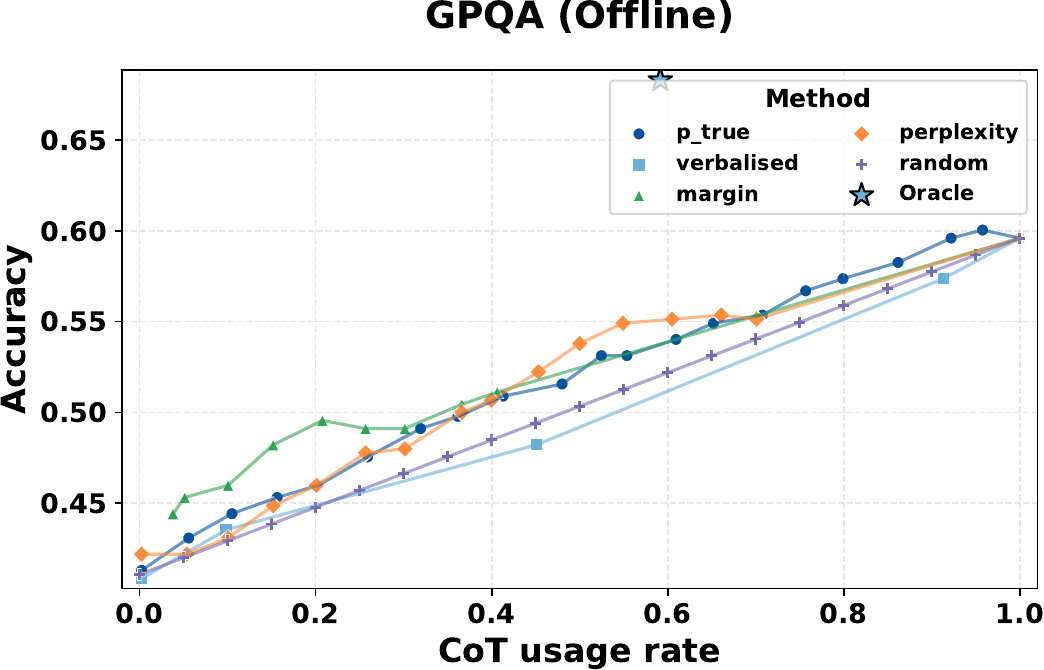}
        \caption{Qwen3-32B - CoT Use}
    \end{subfigure}
    \hfill
    \begin{subfigure}{0.32\textwidth}
        \centering
        \includegraphics[width=\linewidth]{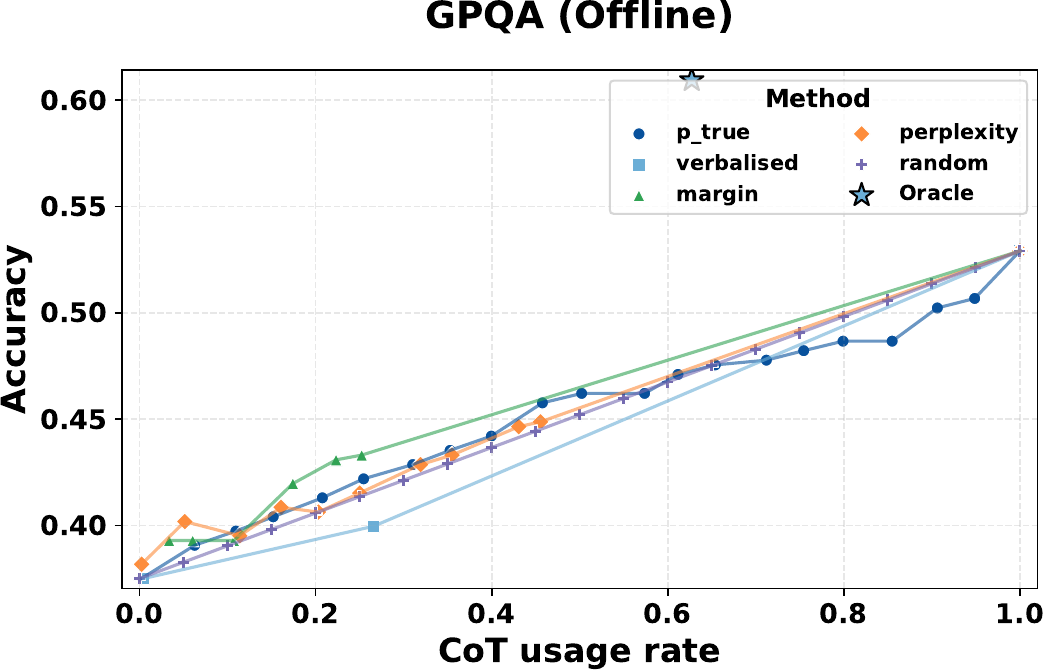}
        \caption{Qwen3-8B - CoT Use}
    \end{subfigure}
    \hfill
    \begin{subfigure}{0.32\textwidth}
        \mbox{}
    \end{subfigure}

    \caption{\textbf{GPQA (Part 1): Accuracy vs. CoT Use.}}
    
    \label{fig:gpqa_cot}
\end{figure}

\begin{figure}[h]
    \centering
    \begin{subfigure}{0.32\textwidth}
        \centering
        \includegraphics[width=\linewidth]{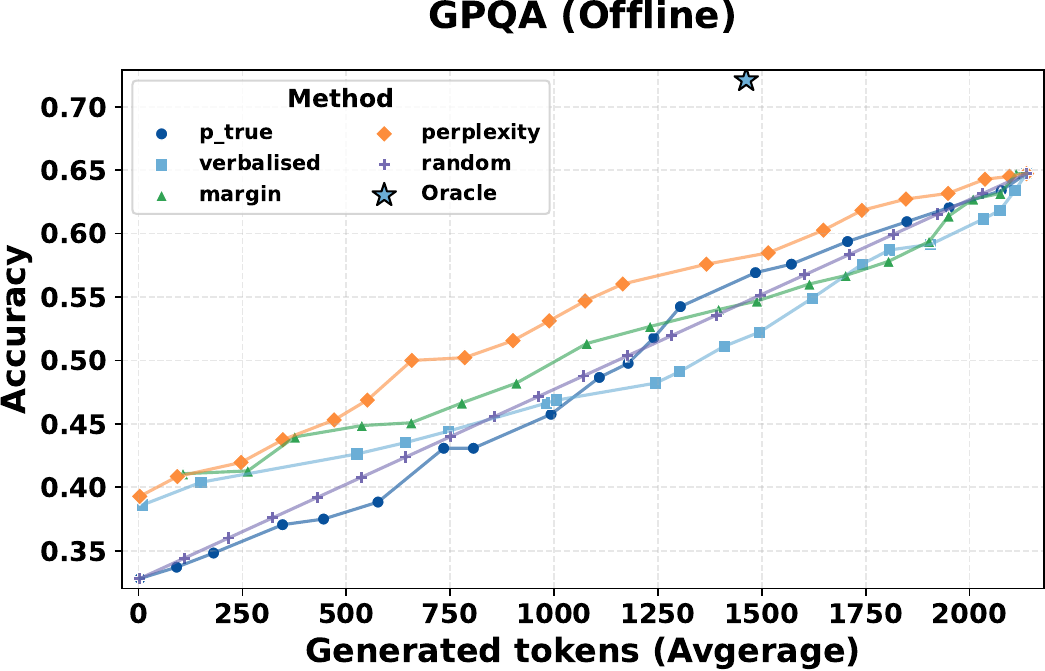}
        \caption{GPT-OSS Medium - Tokens}
    \end{subfigure}
    \hfill
    \begin{subfigure}{0.32\textwidth}
        \centering
        \includegraphics[width=\linewidth]{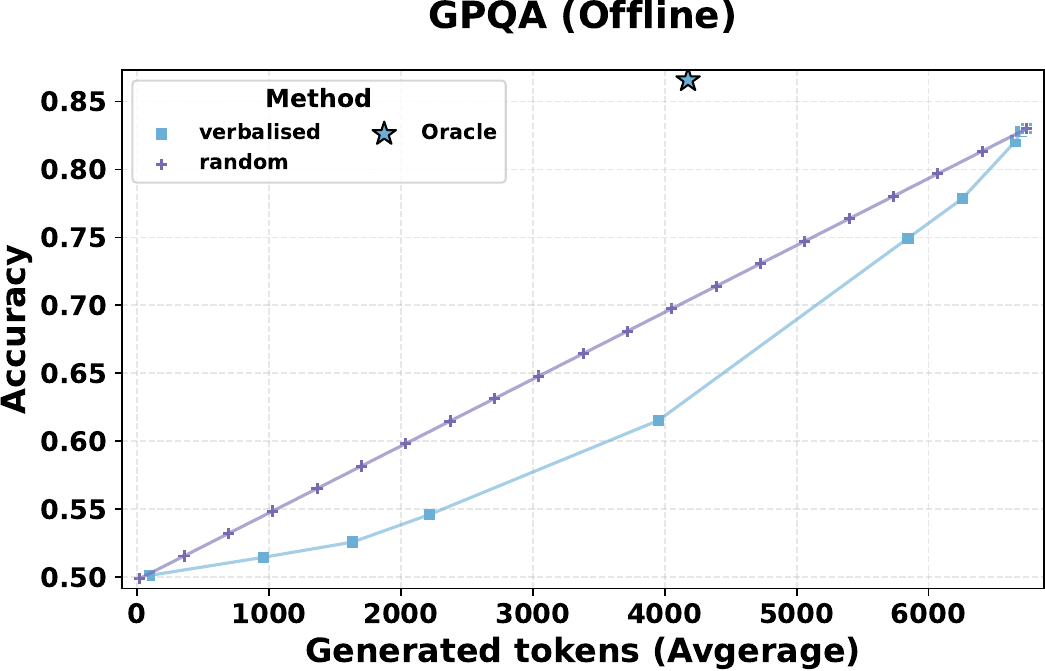}
        \caption{GPT-5 - Tokens}
    \end{subfigure}
    \hfill
    \begin{subfigure}{0.32\textwidth}
        \mbox{}
    \end{subfigure}

    \vskip 0.5em

    \begin{subfigure}{0.32\textwidth}
        \centering
        \includegraphics[width=\linewidth]{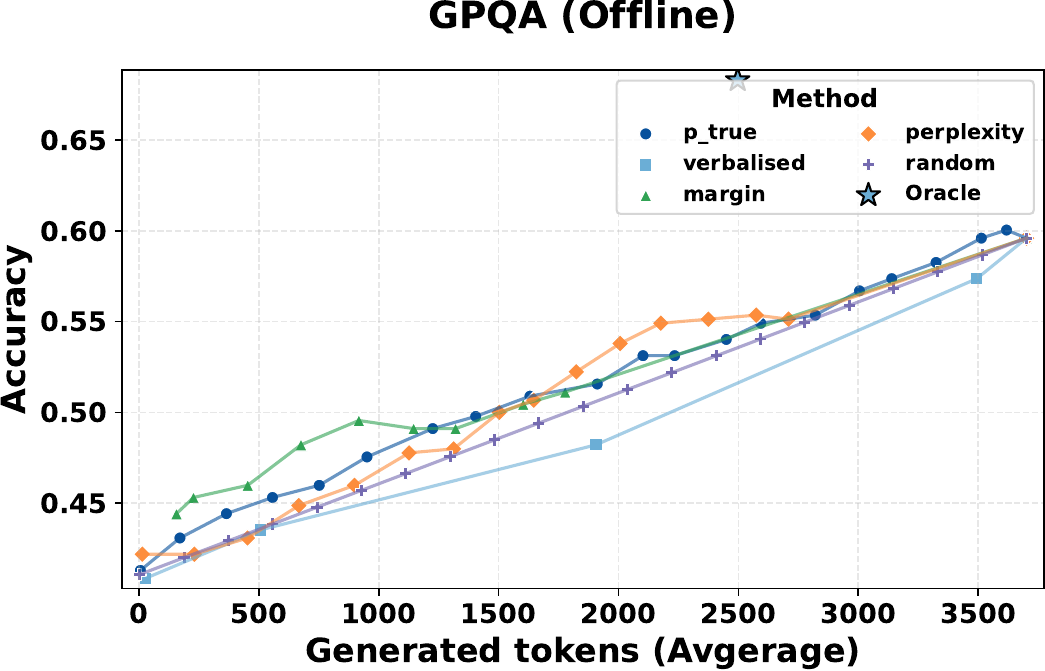}
        \caption{Qwen3-32B - Tokens}
    \end{subfigure}
    \hfill
    \begin{subfigure}{0.32\textwidth}
        \centering
        \includegraphics[width=\linewidth]{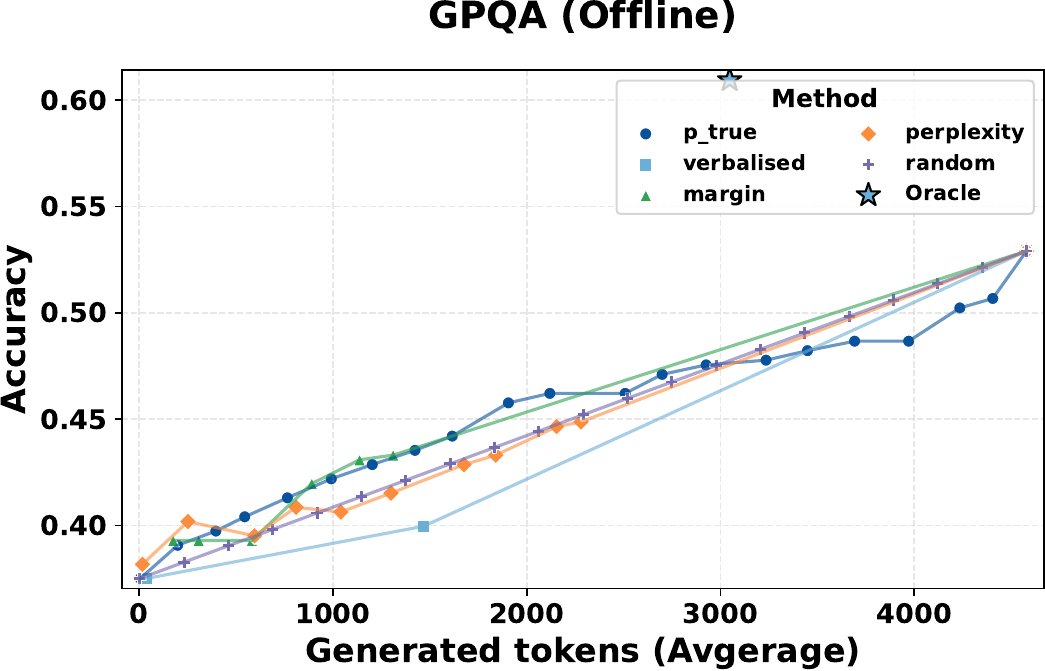}
        \caption{Qwen3-8B - Tokens}
    \end{subfigure}
    \hfill
    \begin{subfigure}{0.32\textwidth}
        \mbox{}
    \end{subfigure}

    \caption{\textbf{GPQA (Part 2): Average Tokens vs. Accuracy.}}
    \label{fig:gpqa_tokens}
\end{figure}


\begin{figure}[h]
    \centering
    \begin{subfigure}{0.32\textwidth}
        \centering
        \includegraphics[width=\linewidth]{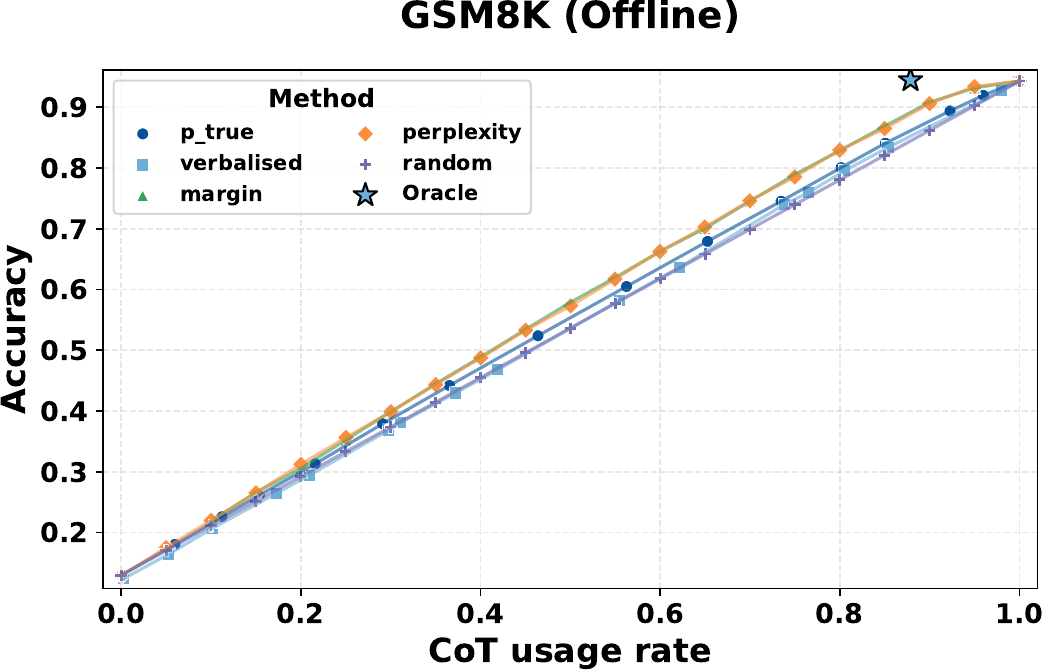}
        \caption{GPT-OSS Medium - CoT Use}
    \end{subfigure}
    \hfill
    \begin{subfigure}{0.32\textwidth}
        \centering
        \includegraphics[width=\linewidth]{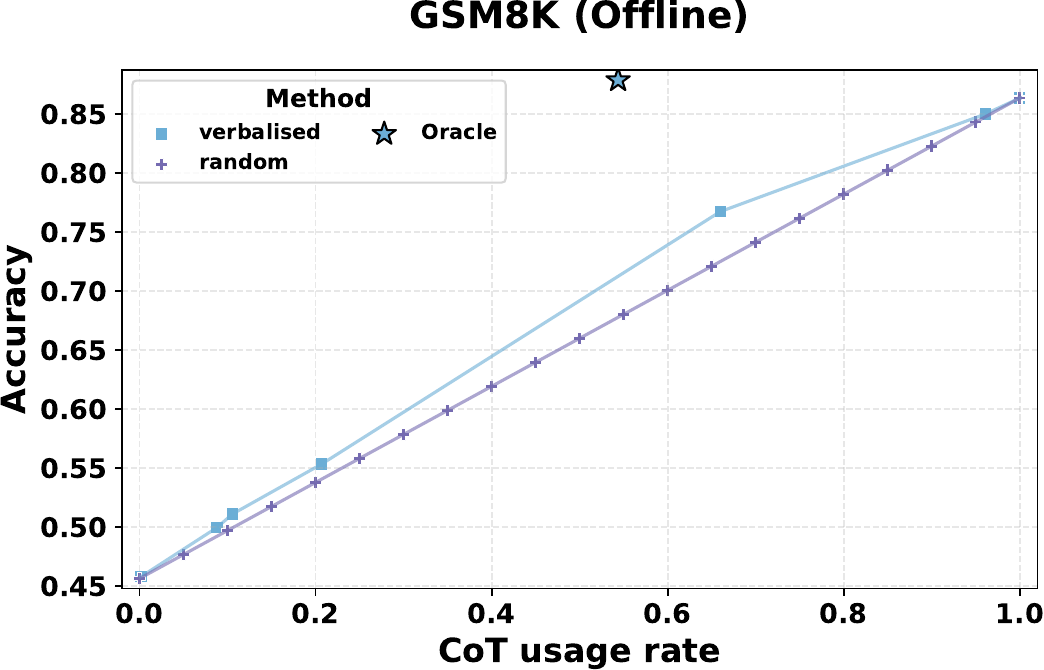}
        \caption{GPT-5 - CoT Use}
    \end{subfigure}
    \hfill
    \begin{subfigure}{0.32\textwidth}
        \mbox{}
    \end{subfigure}

    \vskip 0.5em

    \begin{subfigure}{0.32\textwidth}
        \centering
        \includegraphics[width=\linewidth]{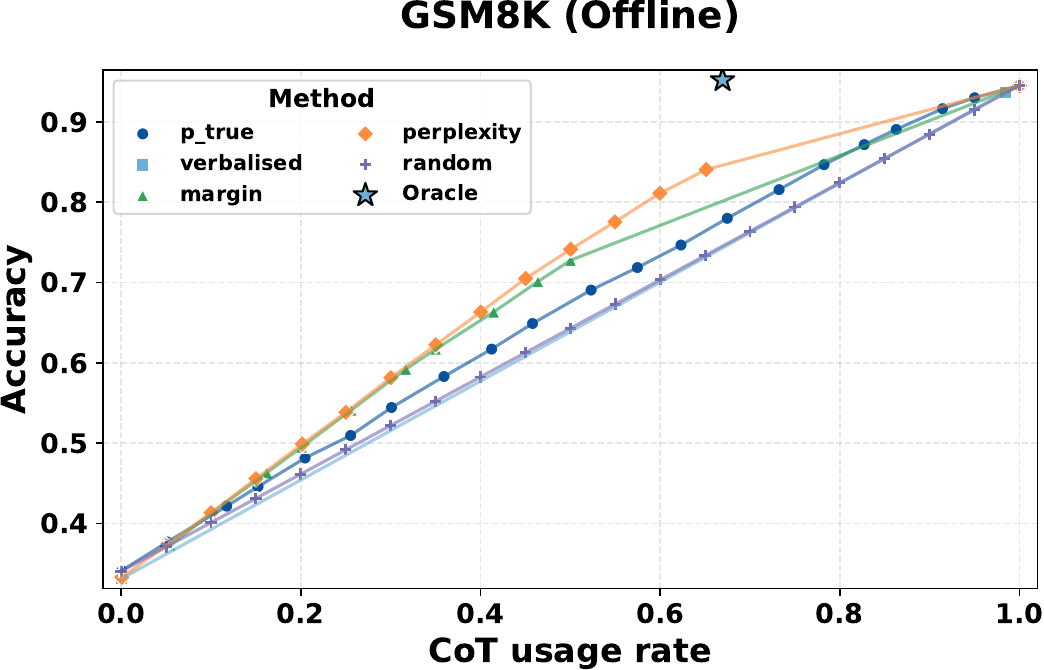}
        \caption{Qwen3-32B - CoT Use}
    \end{subfigure}
    \hfill
    \begin{subfigure}{0.32\textwidth}
        \centering
        \includegraphics[width=\linewidth]{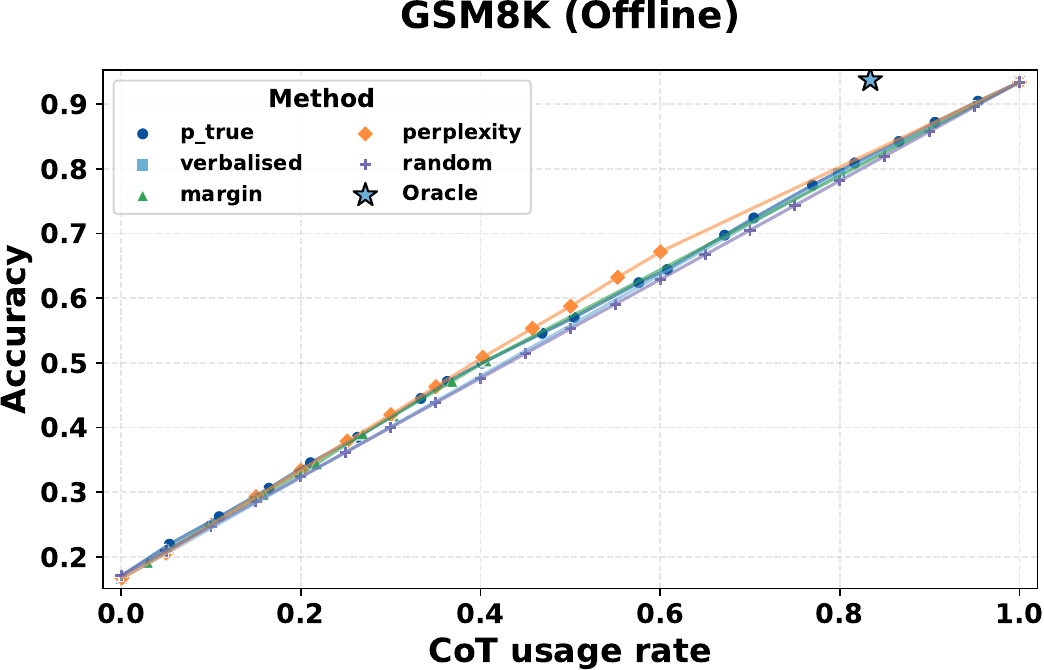}
        \caption{Qwen3-8B - CoT Use}
    \end{subfigure}
    \hfill
    \begin{subfigure}{0.32\textwidth}
        \mbox{}
    \end{subfigure}

    \caption{\textbf{GSM8K (Part 1): Accuracy vs. CoT Use.}}
    
    \label{fig:gsm8k_cot}
\end{figure}

\begin{figure}[h]
    \centering
    \begin{subfigure}{0.32\textwidth}
        \centering
        \includegraphics[width=\linewidth]{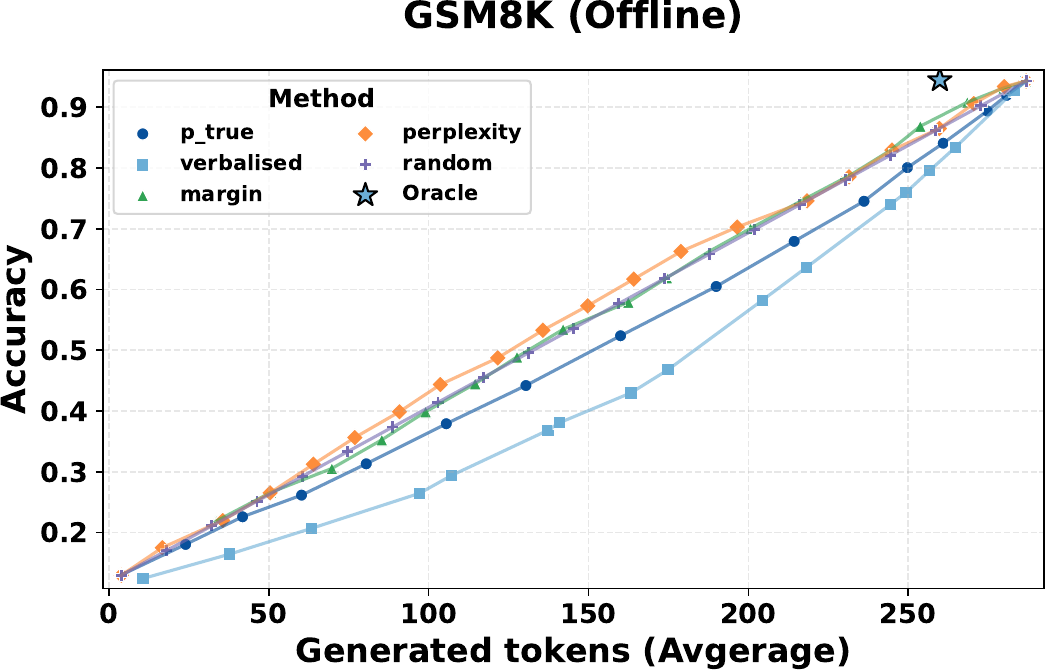}
        \caption{GPT-OSS Medium - Tokens}
    \end{subfigure}
    \hfill
    \begin{subfigure}{0.32\textwidth}
        \centering
        \includegraphics[width=\linewidth]{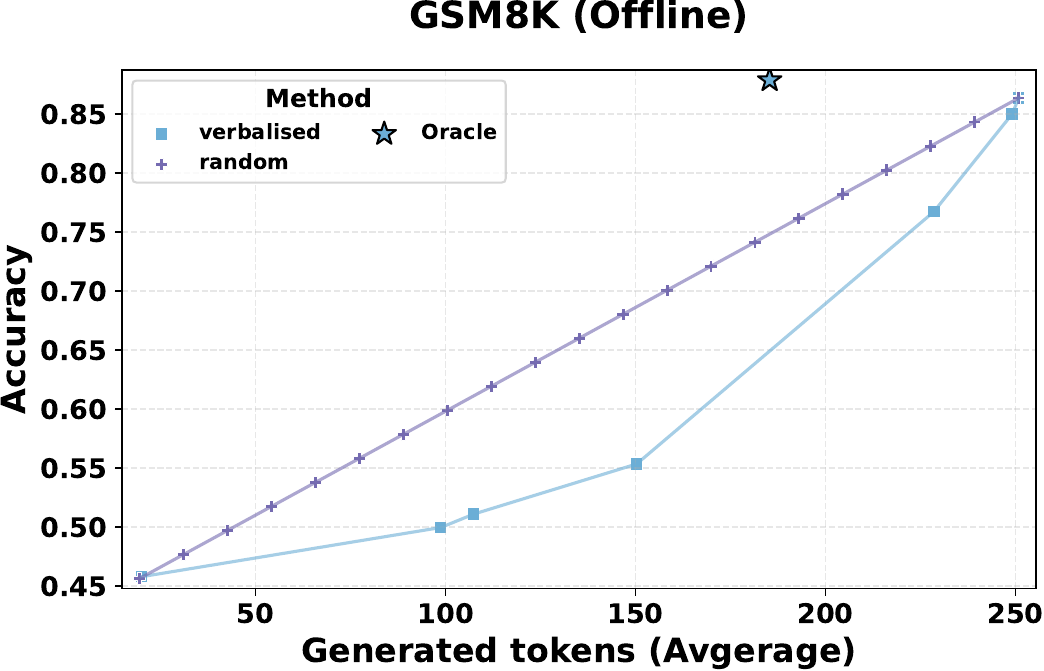}
        \caption{GPT-5 - Tokens}
    \end{subfigure}
    \hfill
    \begin{subfigure}{0.32\textwidth}
        \mbox{}
    \end{subfigure}

    \vskip 0.5em

    \begin{subfigure}{0.32\textwidth}
        \centering
        \includegraphics[width=\linewidth]{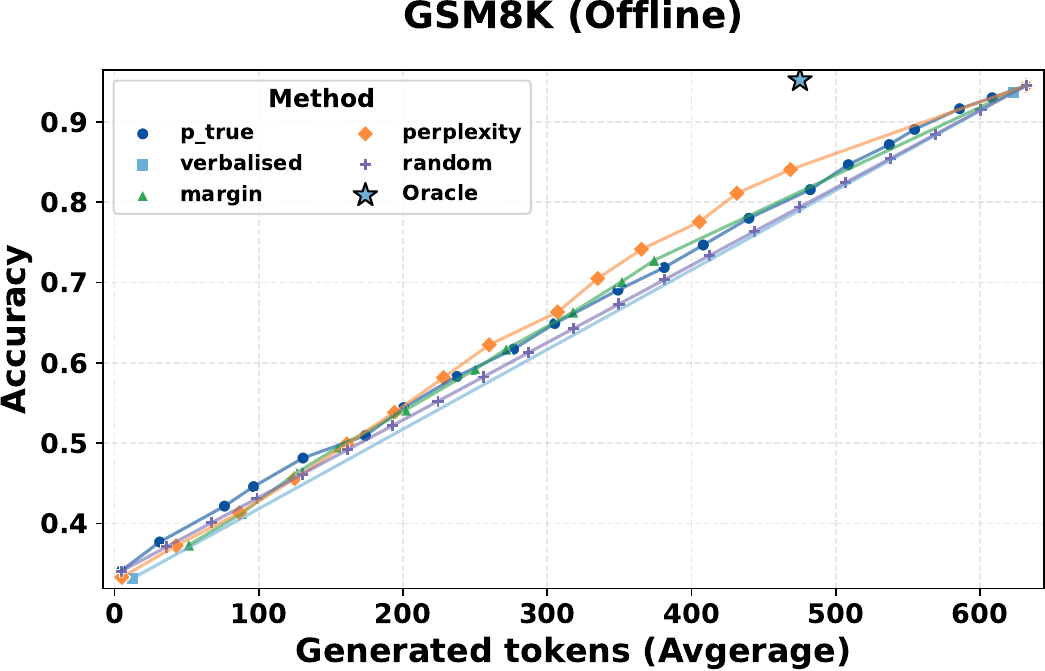}
        \caption{Qwen3-32B - Tokens}
    \end{subfigure}
    \hfill
    \begin{subfigure}{0.32\textwidth}
        \centering
        \includegraphics[width=\linewidth]{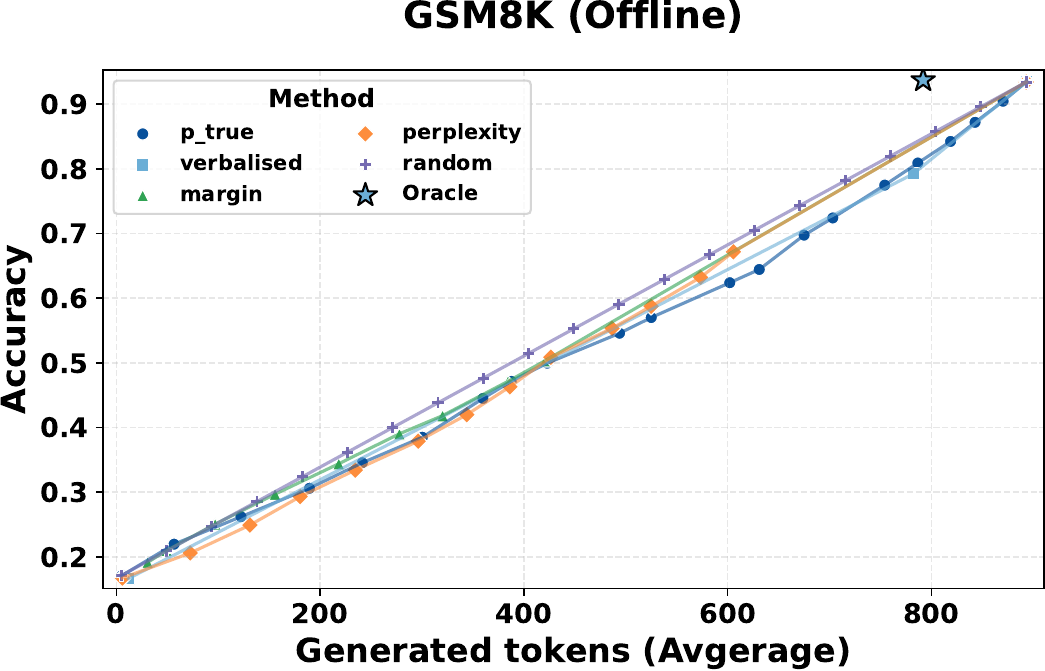}
        \caption{Qwen3-8B - Tokens}
    \end{subfigure}
    \hfill
    \begin{subfigure}{0.32\textwidth}
        \mbox{}
    \end{subfigure}

    \caption{\textbf{GSM8K (Part 2): Average Tokens vs. Accuracy.}}
    \label{fig:gsm8k_tokens}
\end{figure}


\begin{figure}[h]
    \centering
    \begin{subfigure}{0.32\textwidth}
        \centering
        \includegraphics[width=\linewidth]{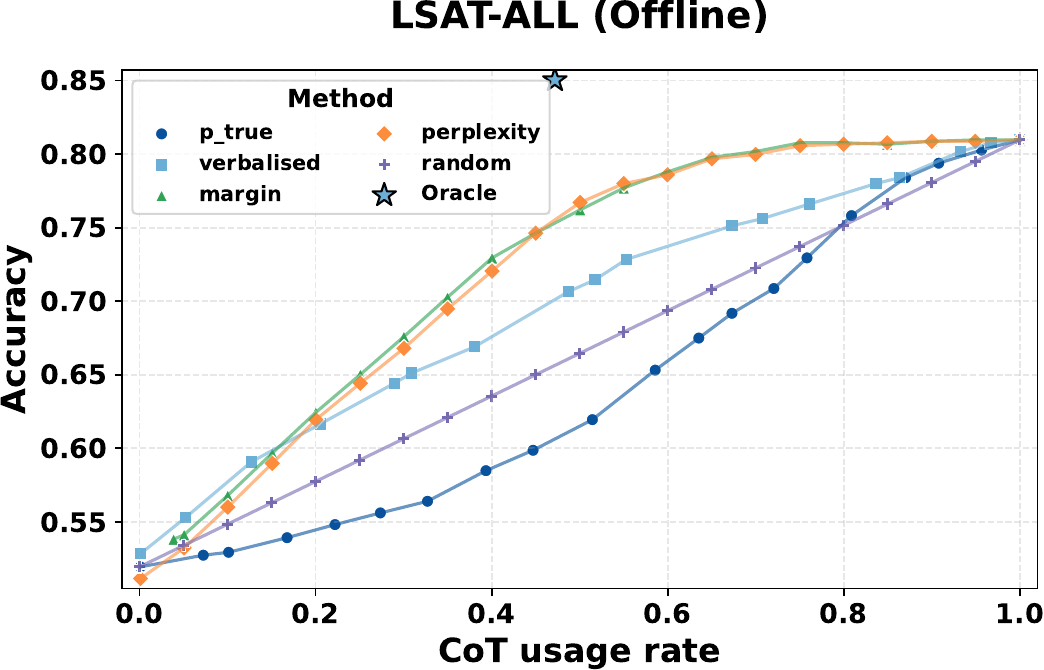}
        \caption{GPT-OSS Medium - CoT Use}
    \end{subfigure}
    \hfill
    \begin{subfigure}{0.32\textwidth}
        \centering
        \includegraphics[width=\linewidth]{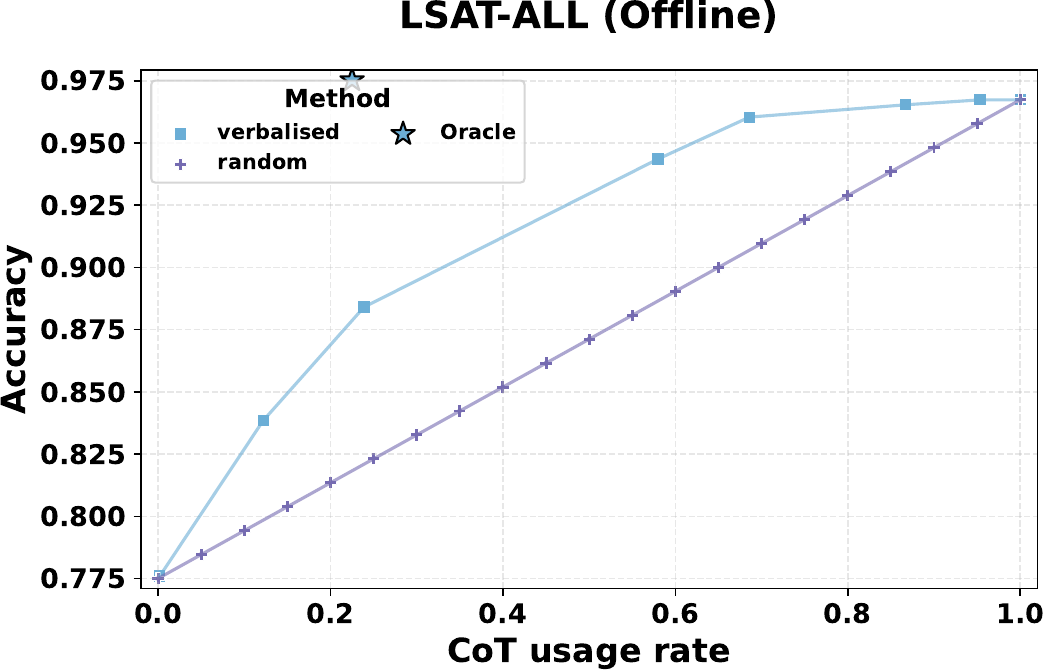}
        \caption{GPT-5 - CoT Use}
    \end{subfigure}
    \hfill
    \begin{subfigure}{0.32\textwidth}
        \mbox{}
    \end{subfigure}

    \vskip 0.5em

    \begin{subfigure}{0.32\textwidth}
        \centering
        \includegraphics[width=\linewidth]{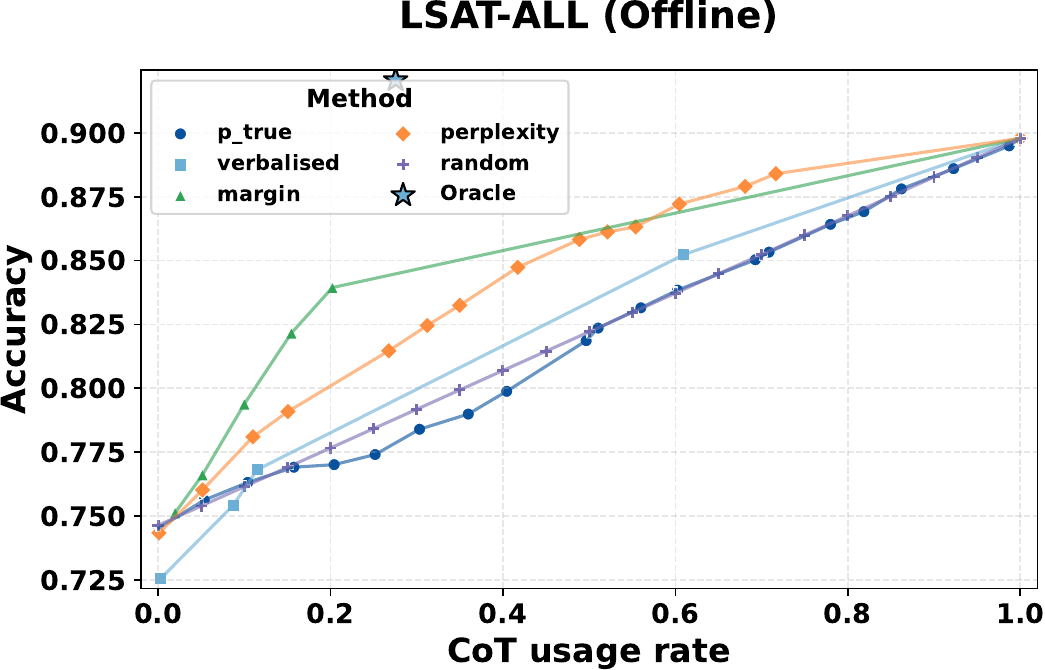}
        \caption{Qwen3-32B - CoT Use}
    \end{subfigure}
    \hfill
    \begin{subfigure}{0.32\textwidth}
        \centering
        \includegraphics[width=\linewidth]{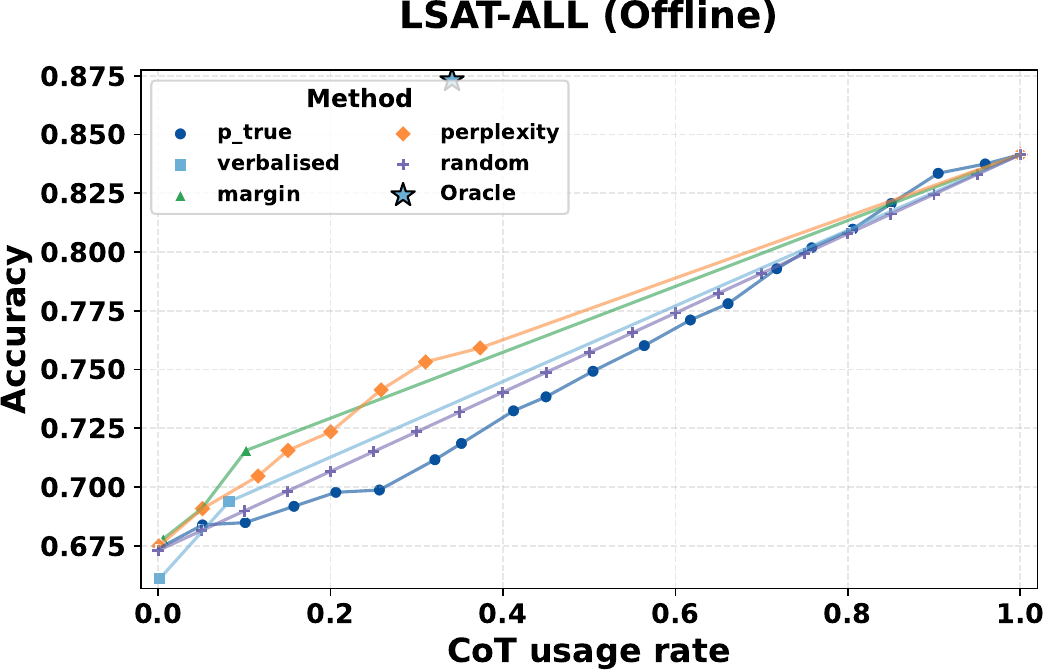}
        \caption{Qwen3-8B - CoT Use}
    \end{subfigure}
    \hfill
    \begin{subfigure}{0.32\textwidth}
        \mbox{}
    \end{subfigure}

    \caption{\textbf{LSAT-All (Part 1): Accuracy vs. CoT Use.}}
    
    \label{fig:lsat_cot}
\end{figure}

\begin{figure}[h]
    \centering
    \begin{subfigure}{0.32\textwidth}
        \centering
        \includegraphics[width=\linewidth]{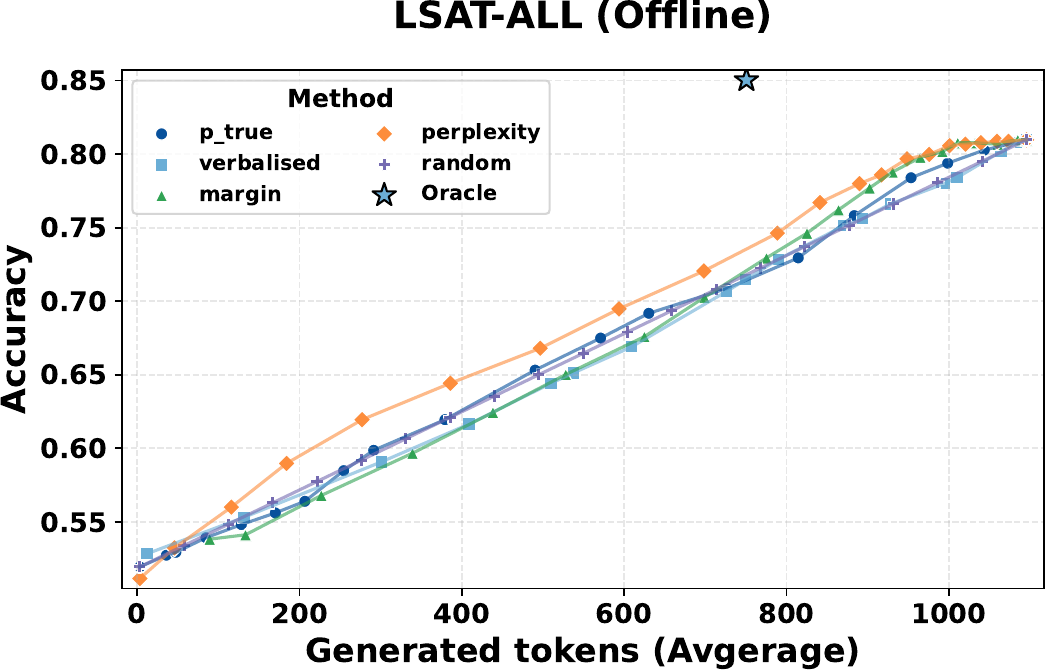}
        \caption{GPT-OSS Medium - Tokens}
    \end{subfigure}
    \hfill
    \begin{subfigure}{0.32\textwidth}
        \centering
        \includegraphics[width=\linewidth]{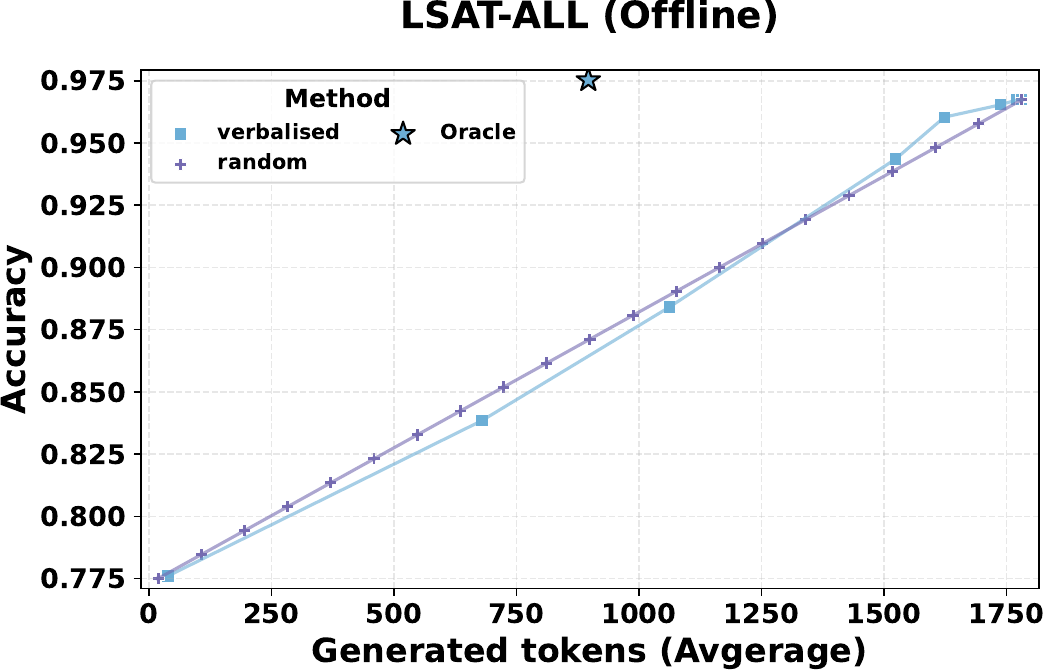}
        \caption{GPT-5 - Tokens}
    \end{subfigure}
    \hfill
    \begin{subfigure}{0.32\textwidth}
        \mbox{}
    \end{subfigure}

    \vskip 0.5em

    \begin{subfigure}{0.32\textwidth}
        \centering
        \includegraphics[width=\linewidth]{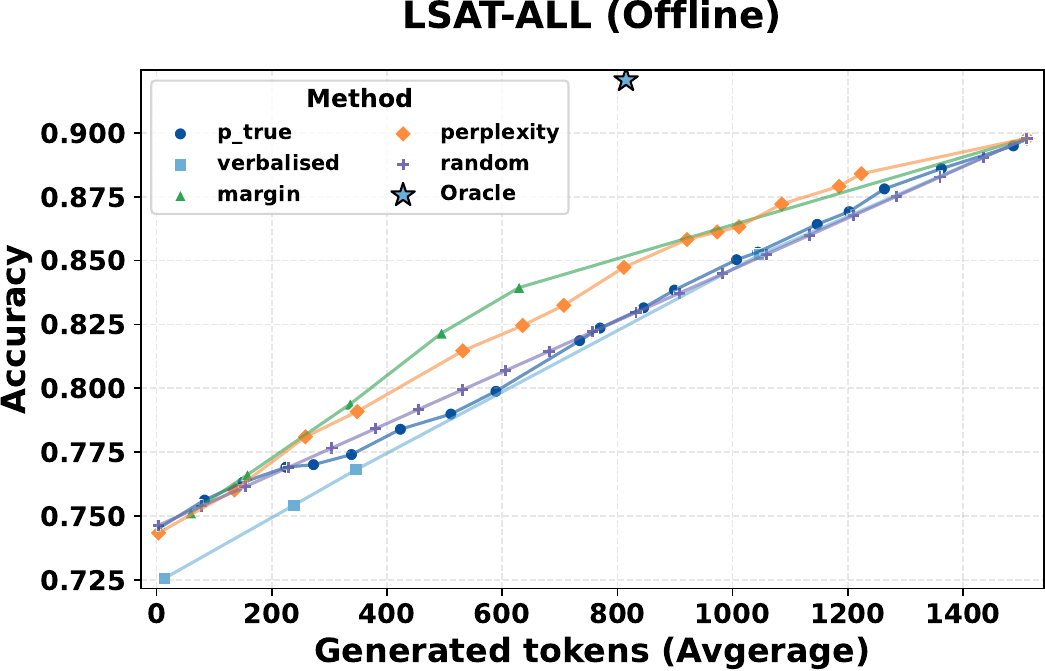}
        \caption{Qwen3-32B - Tokens}
    \end{subfigure}
    \hfill
    \begin{subfigure}{0.32\textwidth}
        \centering
        \includegraphics[width=\linewidth]{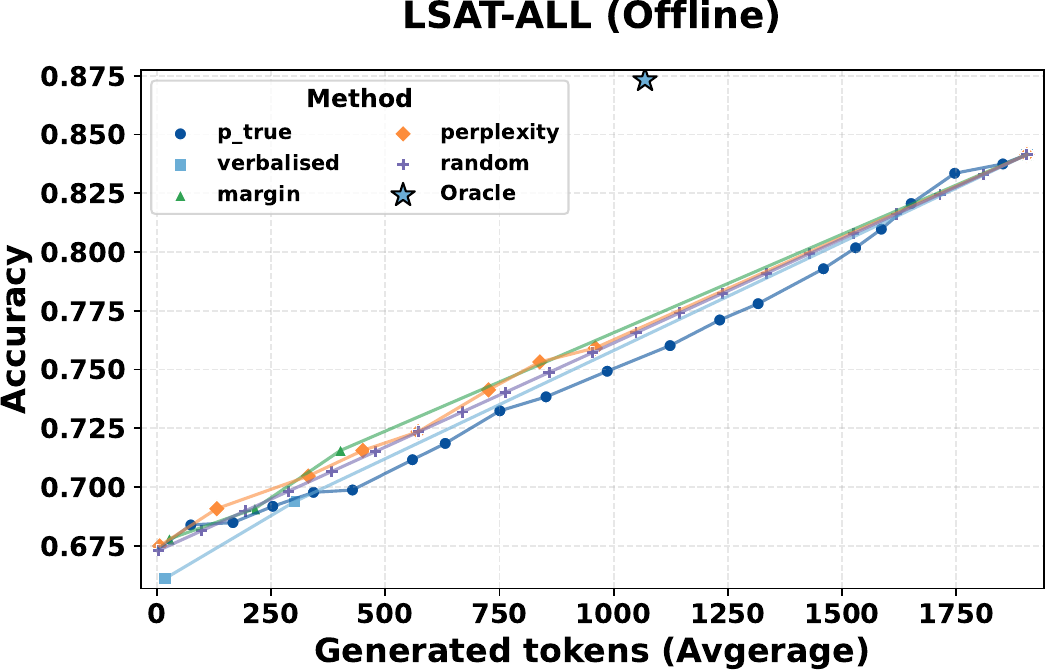}
        \caption{Qwen3-8B - Tokens}
    \end{subfigure}
    \hfill
    \begin{subfigure}{0.32\textwidth}
        \mbox{}
    \end{subfigure}

    \caption{\textbf{LSAT-All (Part 2): Average Tokens vs. Accuracy.}}
    \label{fig:lsat_tokens}
\end{figure}


\begin{figure}[h]
    \centering
    \begin{subfigure}{0.32\textwidth}
        \centering
        \includegraphics[width=\linewidth]{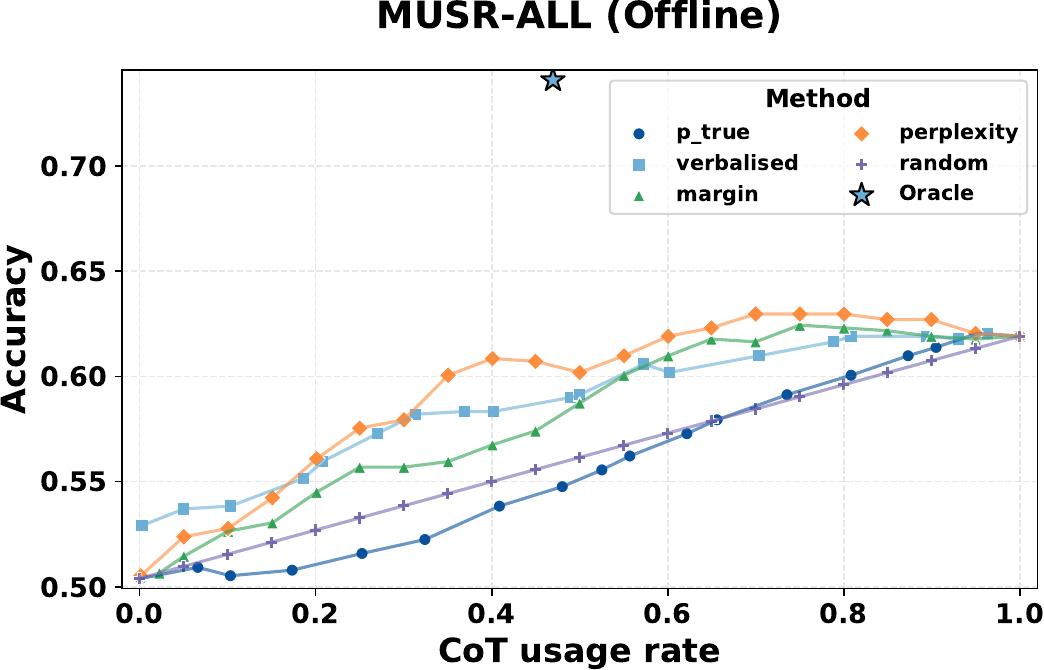}
        \caption{GPT-OSS Medium - CoT Use}
    \end{subfigure}
    \hfill
    \begin{subfigure}{0.32\textwidth}
        \centering
        \includegraphics[width=\linewidth]{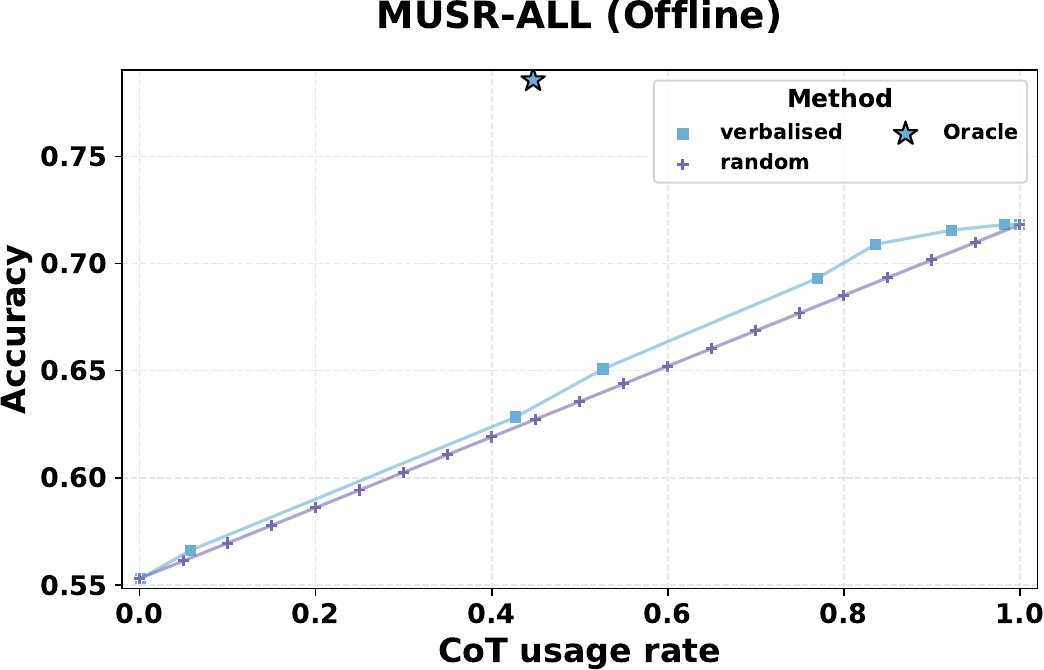}
        \caption{GPT-5 - CoT Use}
    \end{subfigure}
    \hfill
    \begin{subfigure}{0.32\textwidth}
        \mbox{}
    \end{subfigure}

    \vskip 0.5em

    \begin{subfigure}{0.32\textwidth}
        \centering
        \includegraphics[width=\linewidth]{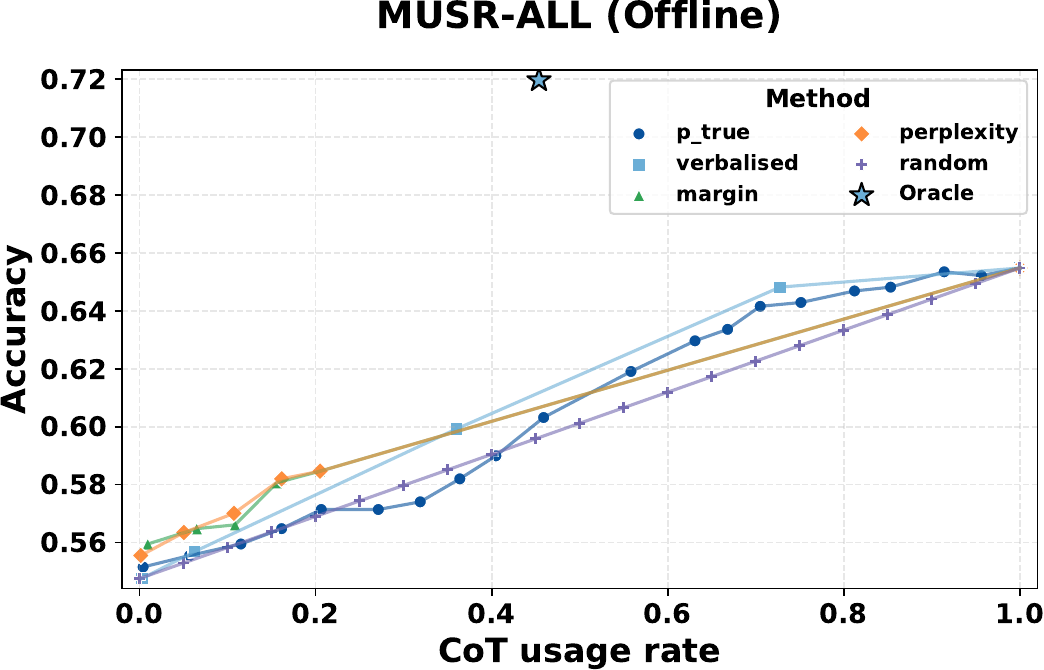}
        \caption{Qwen3-32B - CoT Use}
    \end{subfigure}
    \hfill
    \begin{subfigure}{0.32\textwidth}
        \centering
        \includegraphics[width=\linewidth]{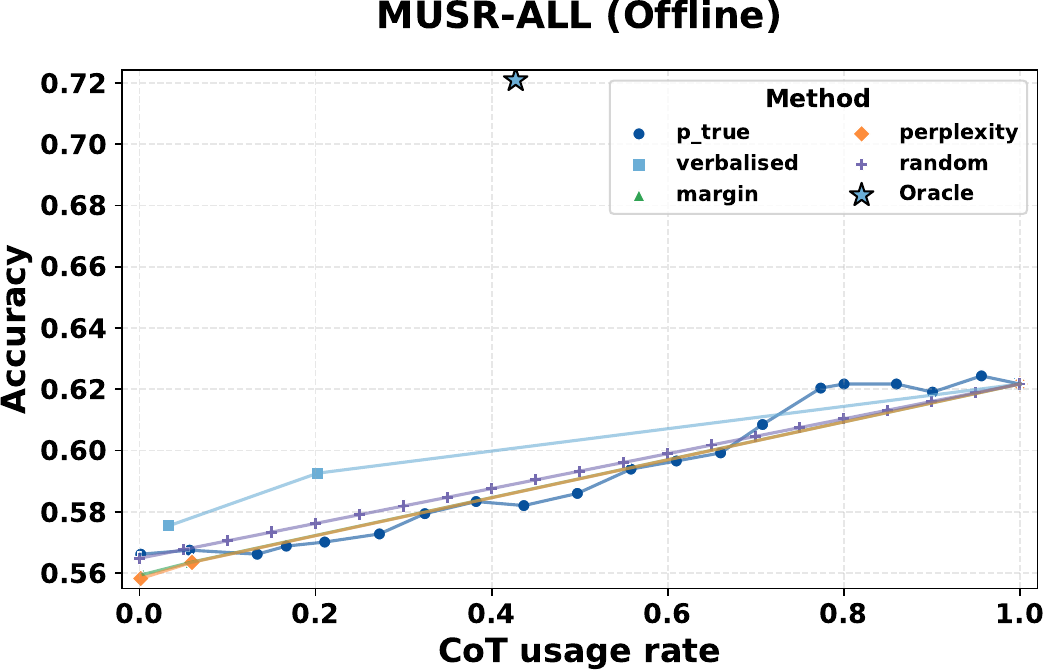}
        \caption{Qwen3-8B - CoT Use}
    \end{subfigure}
    \hfill
    \begin{subfigure}{0.32\textwidth}
        \mbox{}
    \end{subfigure}

    \caption{\textbf{MuSR-All (Part 1): Accuracy vs. CoT Use.}}
    
    \label{fig:musr_cot}
\end{figure}

\begin{figure}[h]
    \centering
    \begin{subfigure}{0.32\textwidth}
        \centering
        \includegraphics[width=\linewidth]{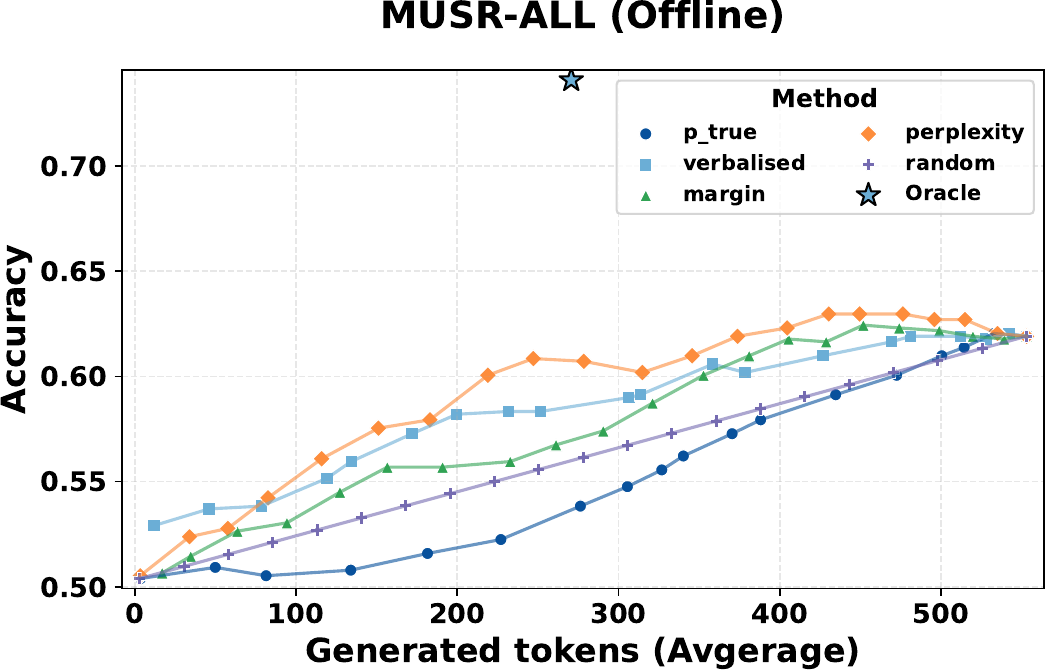}
        \caption{GPT-OSS Medium - Tokens}
    \end{subfigure}
    \hfill
    \begin{subfigure}{0.32\textwidth}
        \centering
        \includegraphics[width=\linewidth]{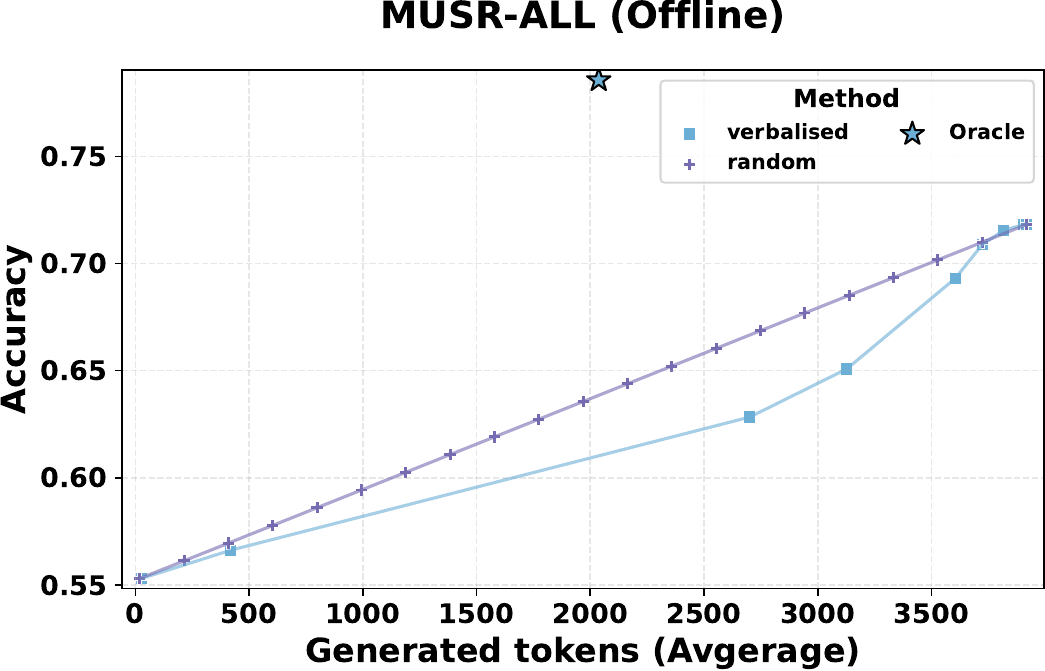}
        \caption{GPT-5 - Tokens}
    \end{subfigure}
    \hfill
    \begin{subfigure}{0.32\textwidth}
        \mbox{}
    \end{subfigure}

    \vskip 0.5em

    \begin{subfigure}{0.32\textwidth}
        \centering
        \includegraphics[width=\linewidth]{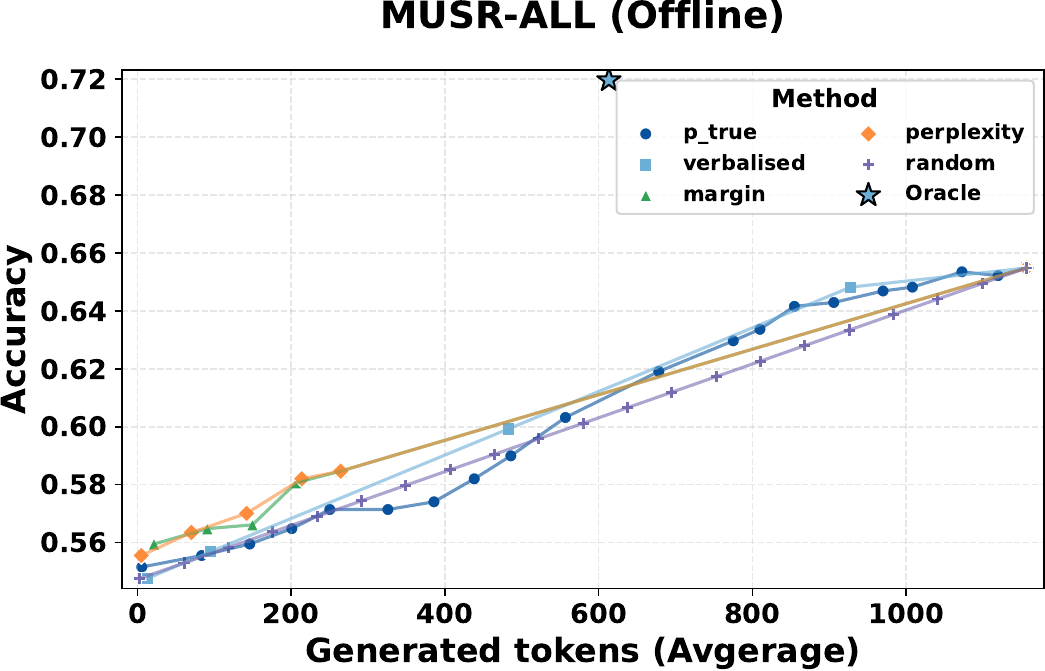}
        \caption{Qwen3-32B - Tokens}
    \end{subfigure}
    \hfill
    \begin{subfigure}{0.32\textwidth}
        \centering
        \includegraphics[width=\linewidth]{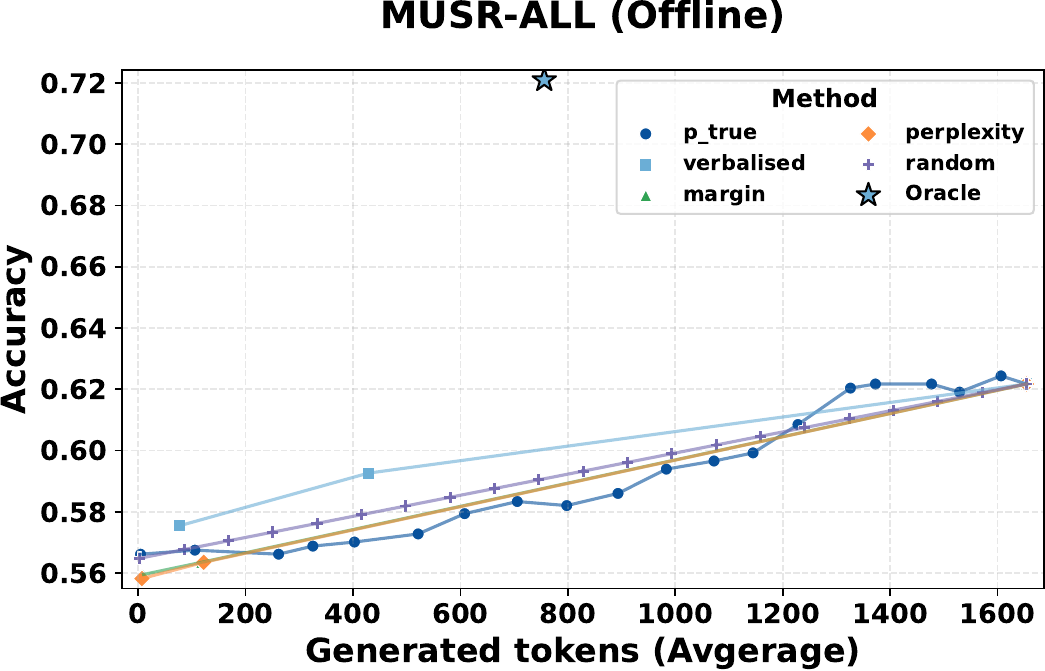}
        \caption{Qwen3-8B - Tokens}
    \end{subfigure}
    \hfill
    \begin{subfigure}{0.32\textwidth}
        \mbox{}
    \end{subfigure}

    \caption{\textbf{MuSR-All (Part 2): Average Tokens vs. Accuracy.}}
    \label{fig:musr_tokens}
\end{figure}


\begin{figure}[h]
    \centering
    \begin{subfigure}{0.32\textwidth}
        \centering
        \includegraphics[width=\linewidth]{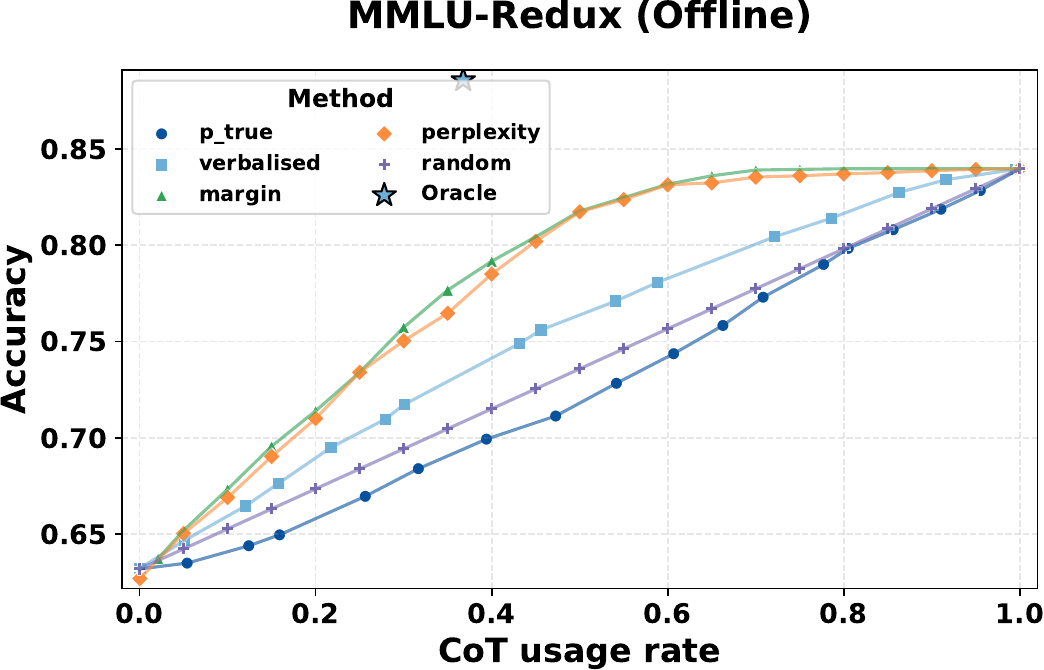}
        \caption{GPT-OSS Medium - CoT Use}
    \end{subfigure}
    \hfill
    \begin{subfigure}{0.32\textwidth}
        \centering
        \includegraphics[width=\linewidth]{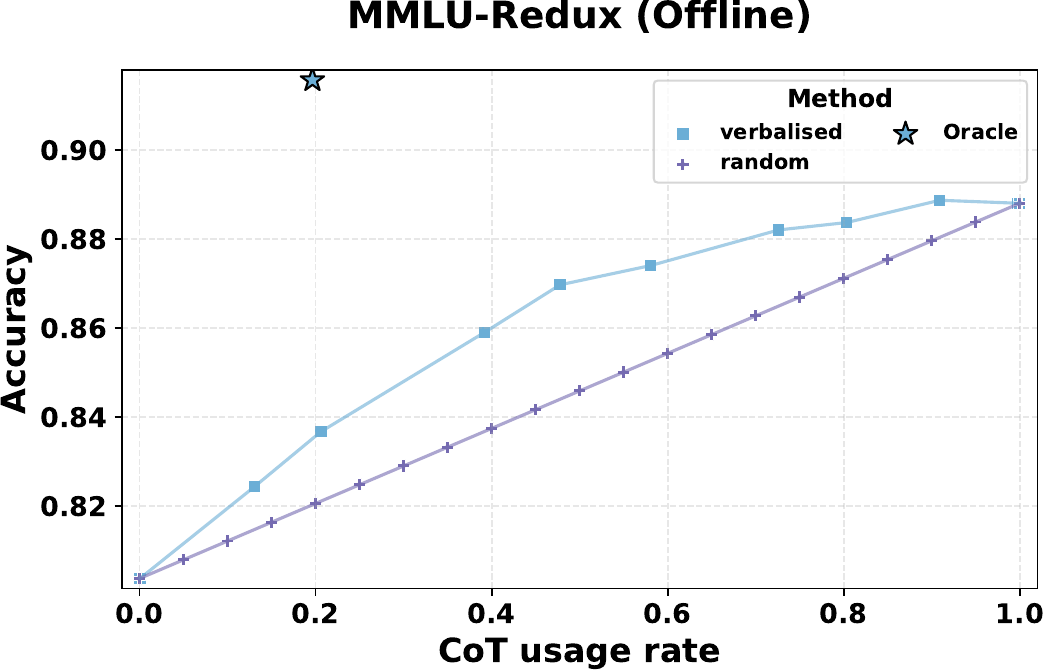}
        \caption{GPT-5 - CoT Use}
    \end{subfigure}
    \hfill
    \begin{subfigure}{0.32\textwidth}
        \mbox{}
    \end{subfigure}

    \vskip 0.5em

    \begin{subfigure}{0.32\textwidth}
        \centering
        \includegraphics[width=\linewidth]{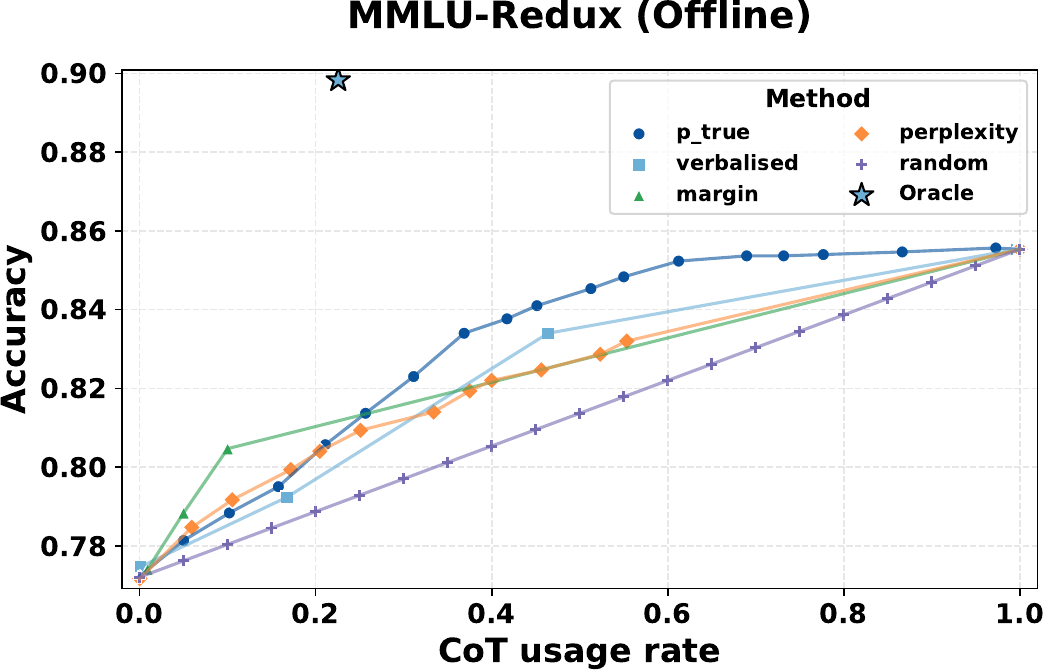}
        \caption{Qwen3-32B - CoT Use}
    \end{subfigure}
    \hfill
    \begin{subfigure}{0.32\textwidth}
        \centering
        \includegraphics[width=\linewidth]{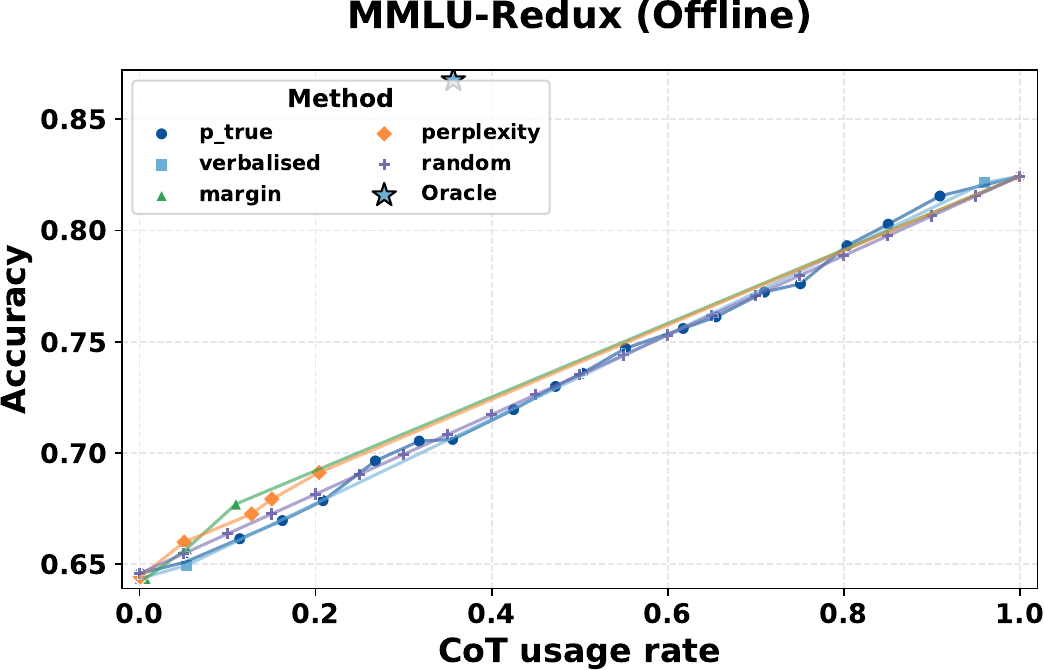}
        \caption{Qwen3-8B - CoT Use}
    \end{subfigure}
    \hfill
    \begin{subfigure}{0.32\textwidth}
        \mbox{}
    \end{subfigure}

    \caption{\textbf{MMLU-Redux (Part 1): Accuracy vs. CoT Use.}}
    
    \label{fig:mmlu_cot}
\end{figure}

\begin{figure}[h]
    \centering
    \begin{subfigure}{0.32\textwidth}
        \centering
        \includegraphics[width=\linewidth]{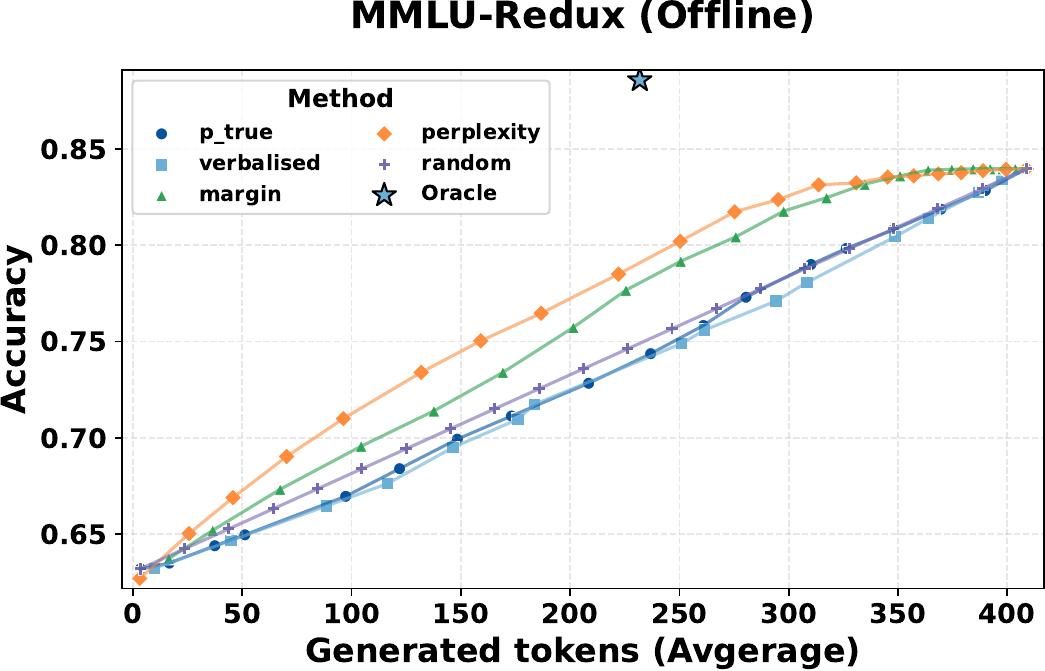}
        \caption{GPT-OSS Medium - Tokens}
    \end{subfigure}
    \hfill
    \begin{subfigure}{0.32\textwidth}
        \centering
        \includegraphics[width=\linewidth]{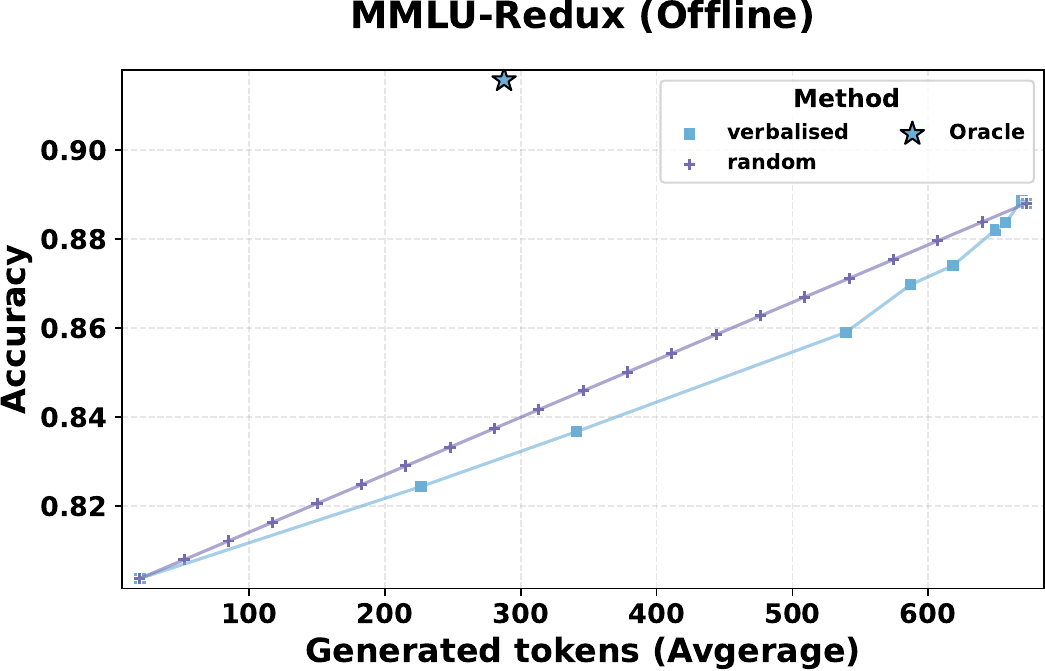}
        \caption{GPT-5 - Tokens}
    \end{subfigure}
    \hfill
    \begin{subfigure}{0.32\textwidth}
        \mbox{}
    \end{subfigure}

    \vskip 0.5em

    \begin{subfigure}{0.32\textwidth}
        \centering
        \includegraphics[width=\linewidth]{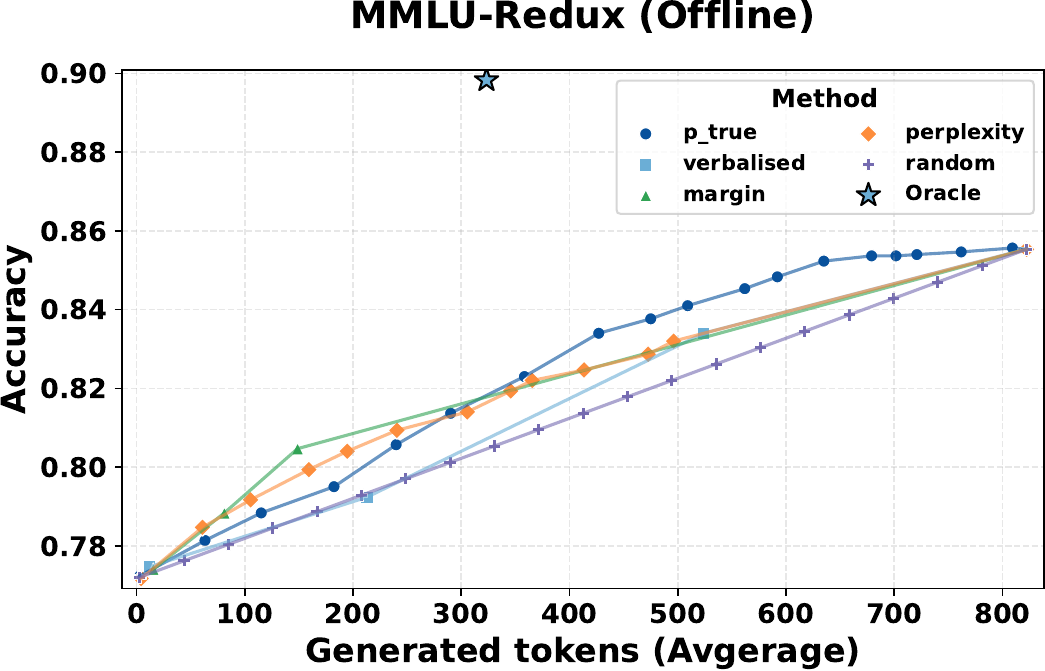}
        \caption{Qwen3-32B - Tokens}
    \end{subfigure}
    \hfill
    \begin{subfigure}{0.32\textwidth}
        \centering
        \includegraphics[width=\linewidth]{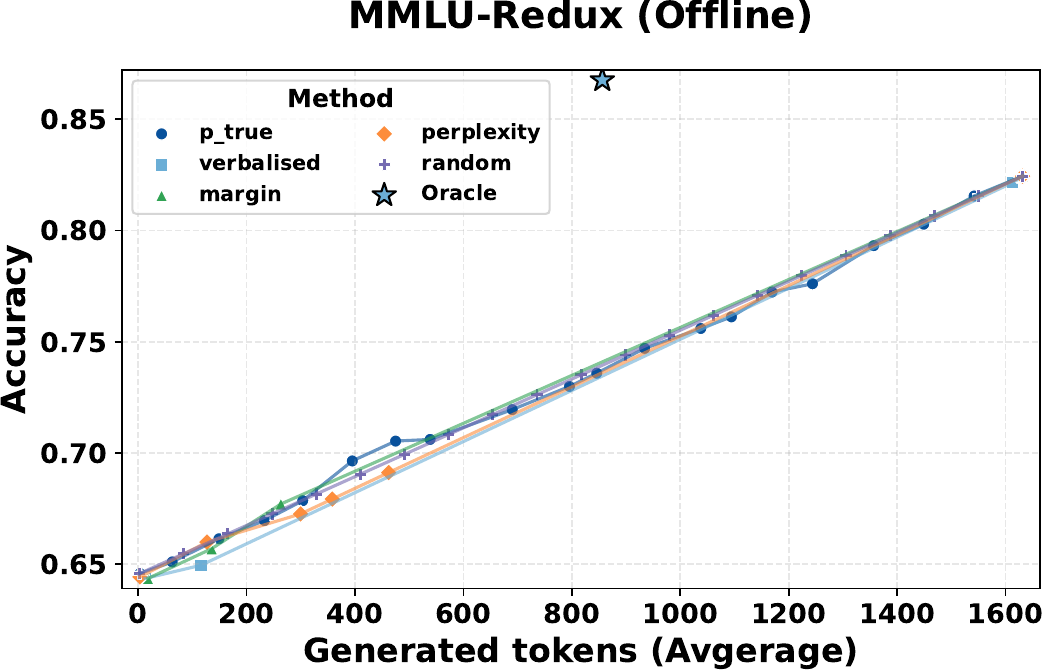}
        \caption{Qwen3-8B - Tokens}
    \end{subfigure}
    \hfill
    \begin{subfigure}{0.32\textwidth}
        \mbox{}
    \end{subfigure}

    \caption{\textbf{MMLU-Redux (Part 2): Average Tokens vs. Accuracy.}}
    \label{fig:mmlu_tokens}
\end{figure}


\begin{figure}[h]
    \centering
    \begin{subfigure}{0.32\textwidth}
        \centering
        \includegraphics[width=\linewidth]{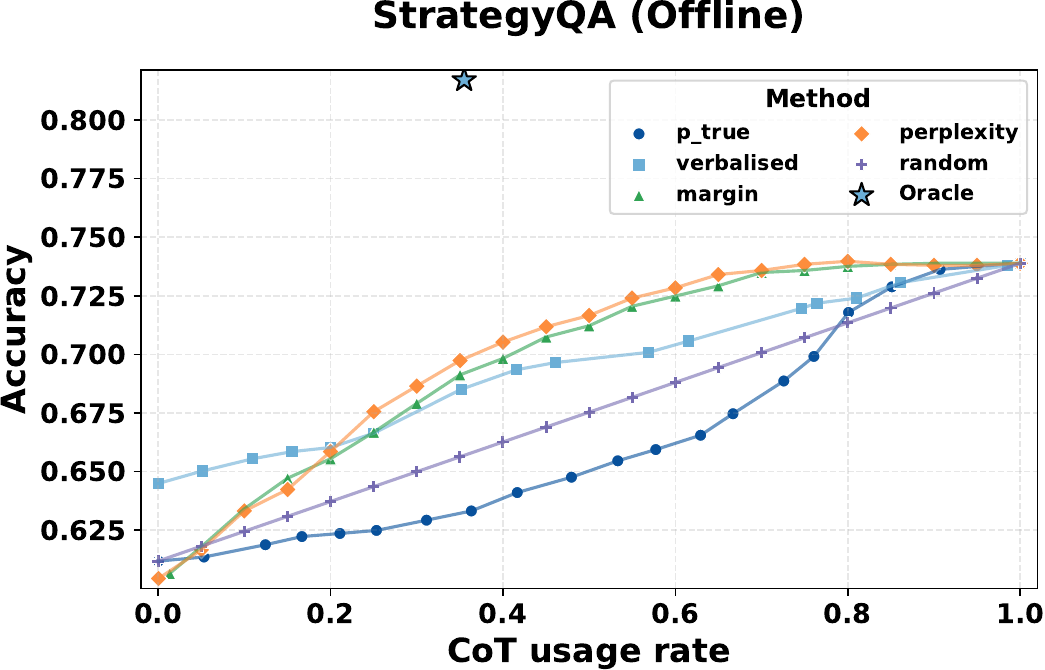}
        \caption{GPT-OSS Medium - CoT Use}
    \end{subfigure}
    \hfill
    \begin{subfigure}{0.32\textwidth}
        \centering
        \includegraphics[width=\linewidth]{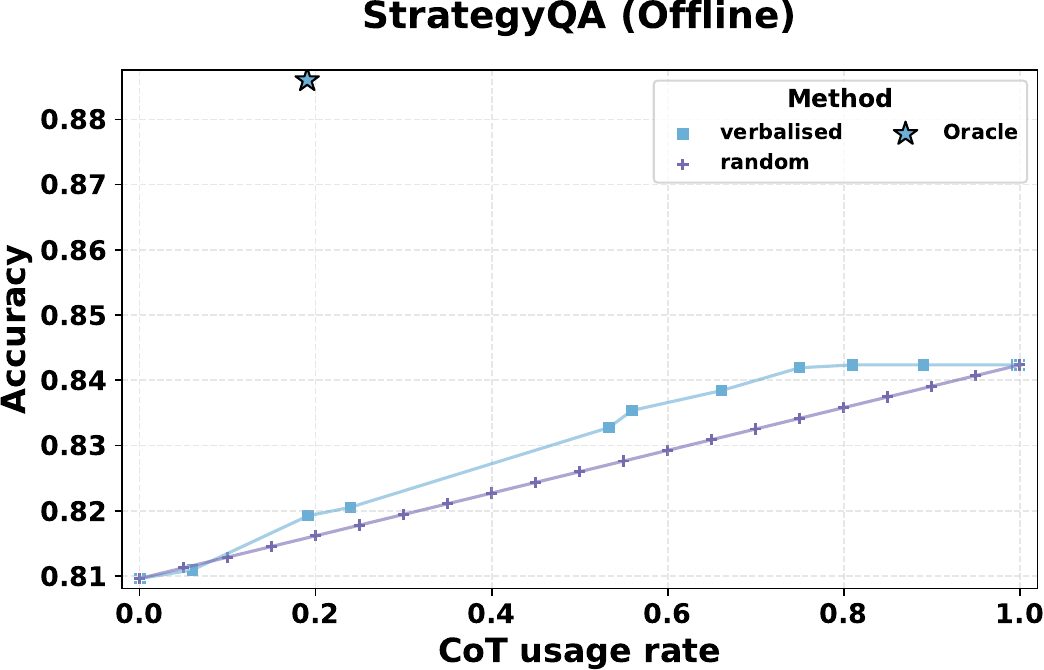}
        \caption{GPT-5 - CoT Use}
    \end{subfigure}
    \hfill
    \begin{subfigure}{0.32\textwidth}
        \mbox{}
    \end{subfigure}

    \vskip 0.5em

    \begin{subfigure}{0.32\textwidth}
        \centering
        \includegraphics[width=\linewidth]{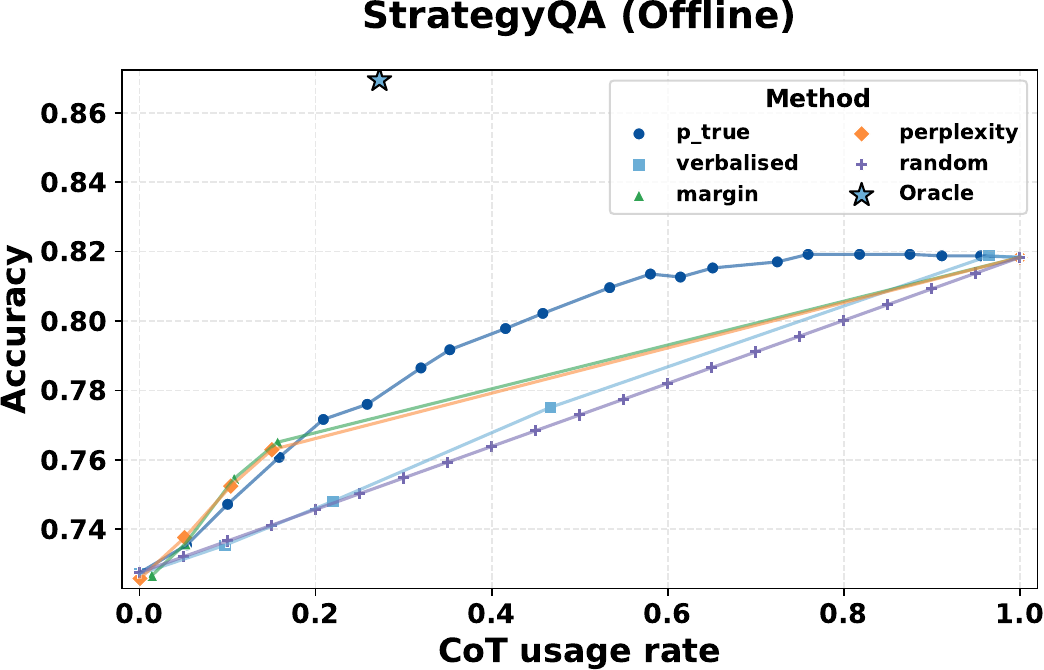}
        \caption{Qwen3-32B - CoT Use}
    \end{subfigure}
    \hfill
    \begin{subfigure}{0.32\textwidth}
        \centering
        \includegraphics[width=\linewidth]{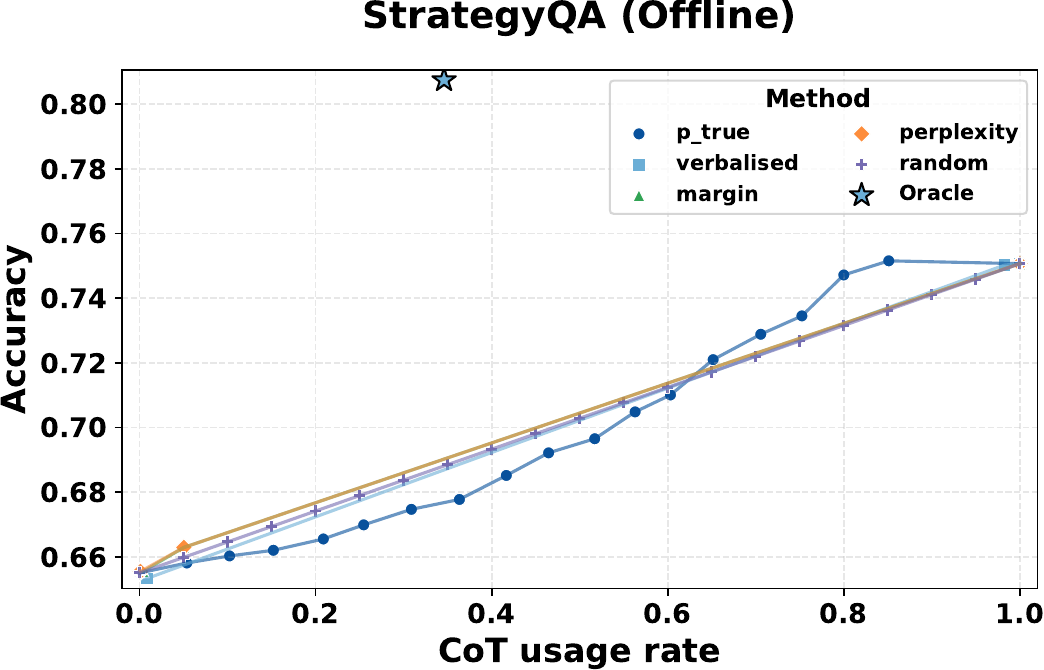}
        \caption{Qwen3-8B - CoT Use}
    \end{subfigure}
    \hfill
    \begin{subfigure}{0.32\textwidth}
        \mbox{}
    \end{subfigure}

    \caption{\textbf{StrategyQA (Part 1): Accuracy vs. CoT Use.}}
    
    \label{fig:strategyqa_cot}
\end{figure}

\begin{figure}[h]
    \centering
    \begin{subfigure}{0.32\textwidth}
        \centering
        \includegraphics[width=\linewidth]{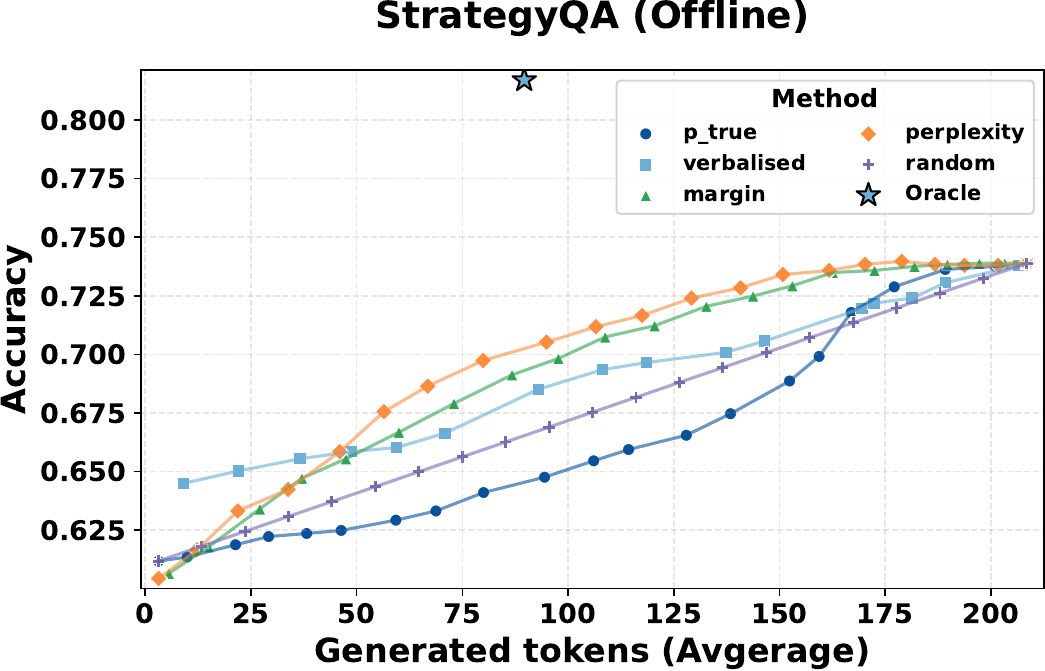}
        \caption{GPT-OSS Medium - Tokens}
    \end{subfigure}
    \hfill
    \begin{subfigure}{0.32\textwidth}
        \centering
        \includegraphics[width=\linewidth]{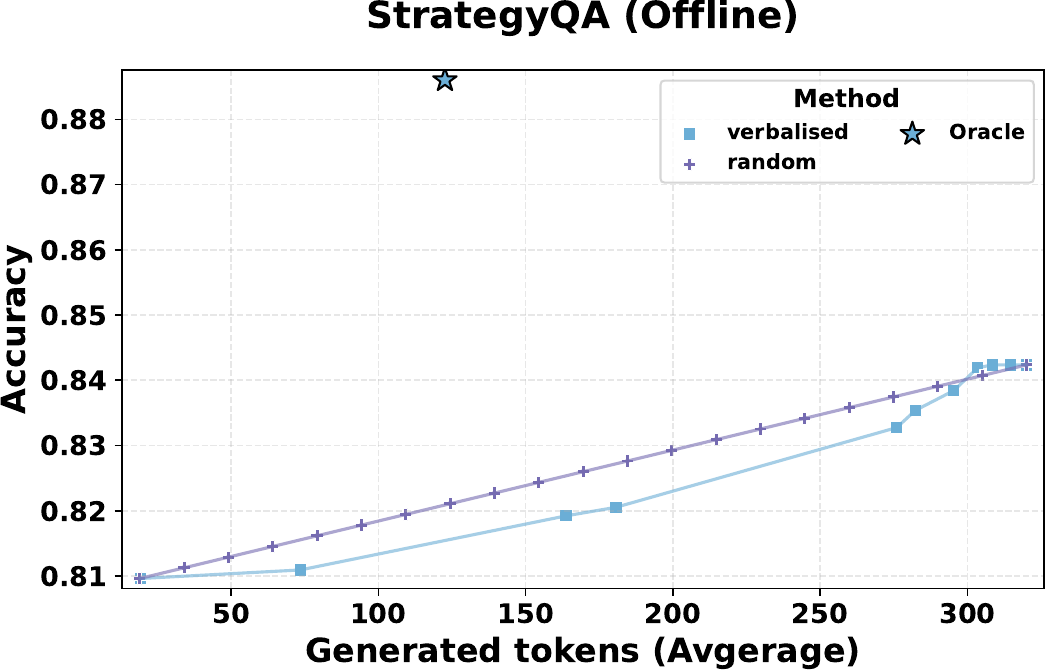}
        \caption{GPT-5 - Tokens}
    \end{subfigure}
    \hfill
    \begin{subfigure}{0.32\textwidth}
        \mbox{}
    \end{subfigure}

    \vskip 0.5em

    \begin{subfigure}{0.32\textwidth}
        \centering
        \includegraphics[width=\linewidth]{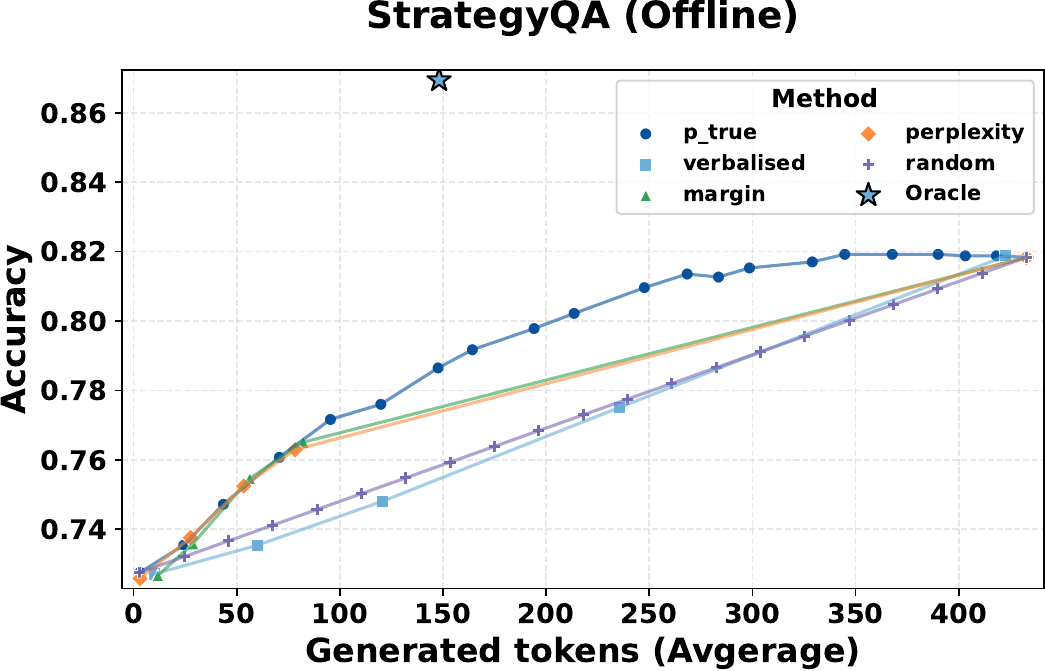}
        \caption{Qwen3-32B - Tokens}
    \end{subfigure}
    \hfill
    \begin{subfigure}{0.32\textwidth}
        \centering
        \includegraphics[width=\linewidth]{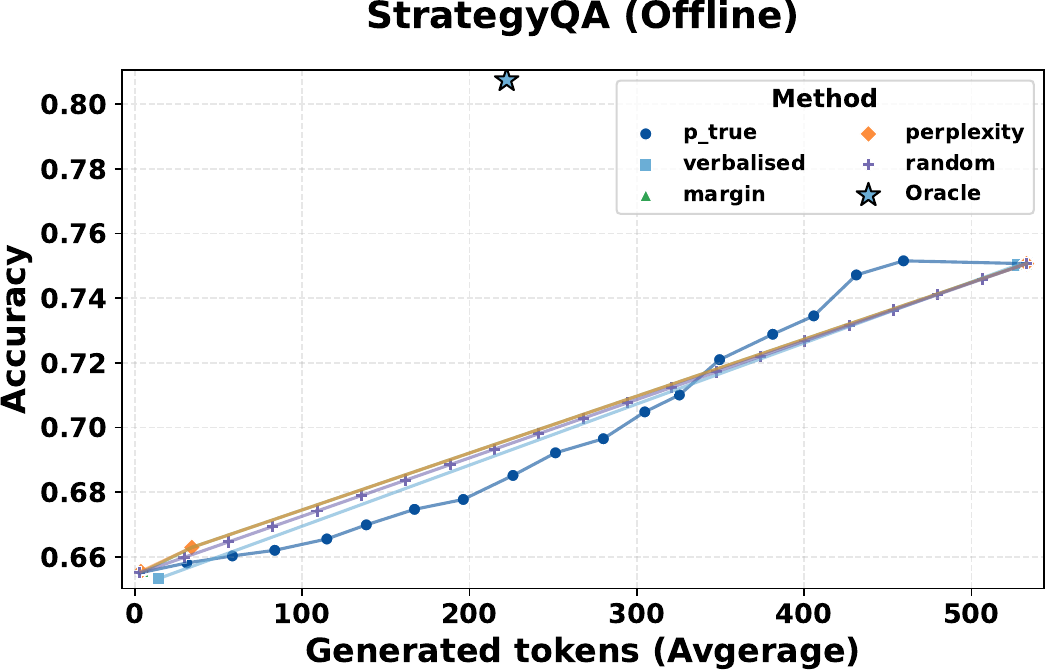}
        \caption{Qwen3-8B - Tokens}
    \end{subfigure}
    \hfill
    \begin{subfigure}{0.32\textwidth}
        \mbox{}
    \end{subfigure}

    \caption{\textbf{StrategyQA (Part 2): Average Tokens vs. Accuracy.}}
    \label{fig:strategyqa_tokens}
\end{figure}

\clearpage
\section{Reliability Diagrams}
\label{sec:reliability-diagrams}

\begin{figure}[h!] 
    \centering
    
    \begin{subfigure}[t]{1.0\linewidth}
        \centering
        \includegraphics[width=0.95\linewidth, height=4.5cm, keepaspectratio]{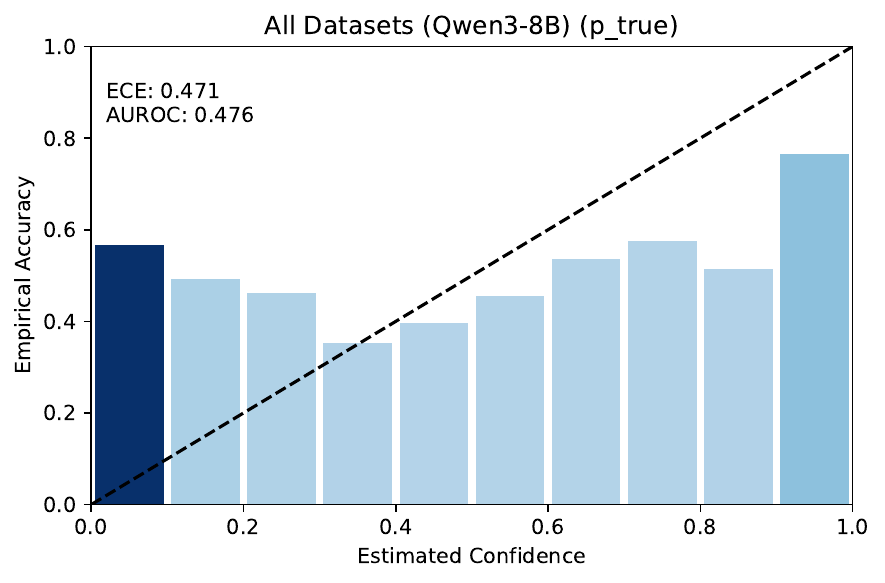}
        \subcaption{p\_true}
    \end{subfigure}
    
    \vspace{0.5em} 

    \begin{subfigure}[t]{1.0\linewidth}
        \centering
        \includegraphics[width=0.95\linewidth, height=4.5cm, keepaspectratio]{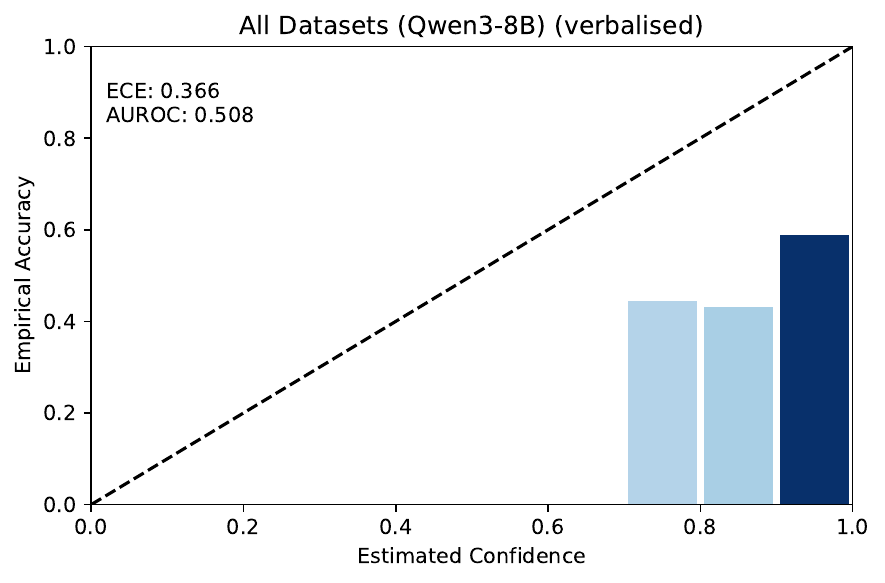}
        \subcaption{Verbalised}
    \end{subfigure}
    
    \vspace{0.5em}

    \begin{subfigure}[t]{1.0\linewidth}
        \centering
        \includegraphics[width=0.95\linewidth, height=4.5cm, keepaspectratio]{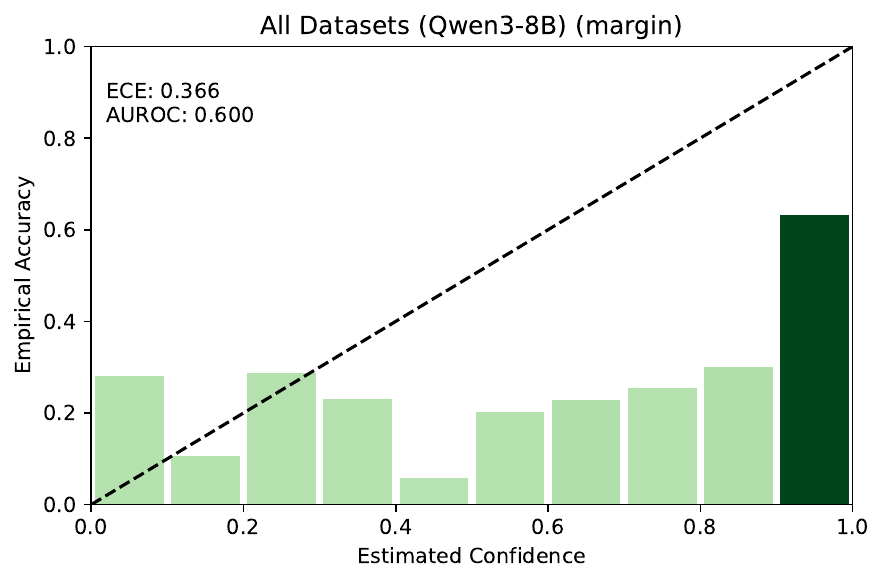}
        \subcaption{Margin}
    \end{subfigure}
    
    \vspace{0.5em}

    \begin{subfigure}[t]{1.0\linewidth}
        \centering
        \includegraphics[width=0.95\linewidth, height=4.5cm, keepaspectratio]{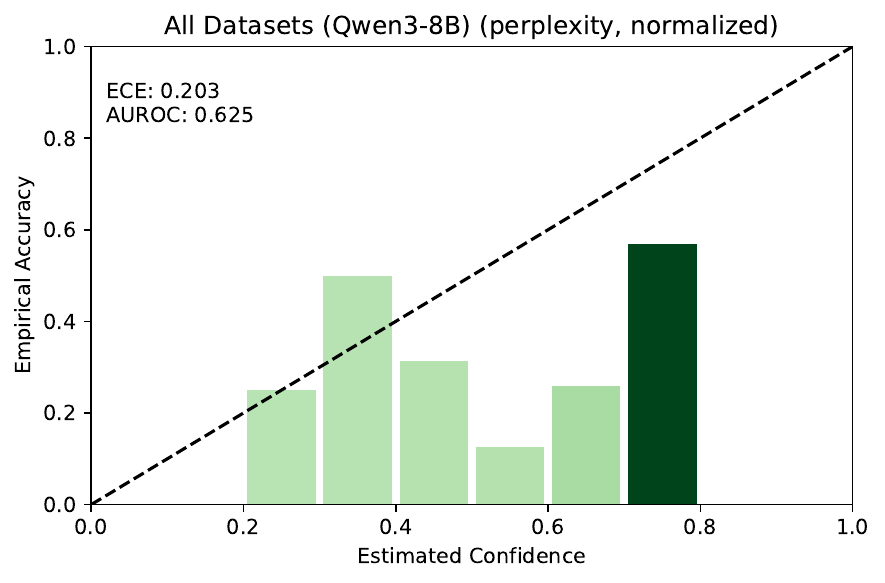}
        \subcaption{Perplexity (norm)}
    \end{subfigure}

    \caption{Reliability: \textbf{Qwen3-8B}. Bars darken with bin count; dashed line is perfect calibration.}
    \label{fig:reliability-qwen3-8b}
\end{figure}

\begin{figure}[h!]
    \centering
    
    \begin{subfigure}[t]{1.0\linewidth}
        \centering
        \includegraphics[width=0.95\linewidth, height=4.2cm, keepaspectratio]{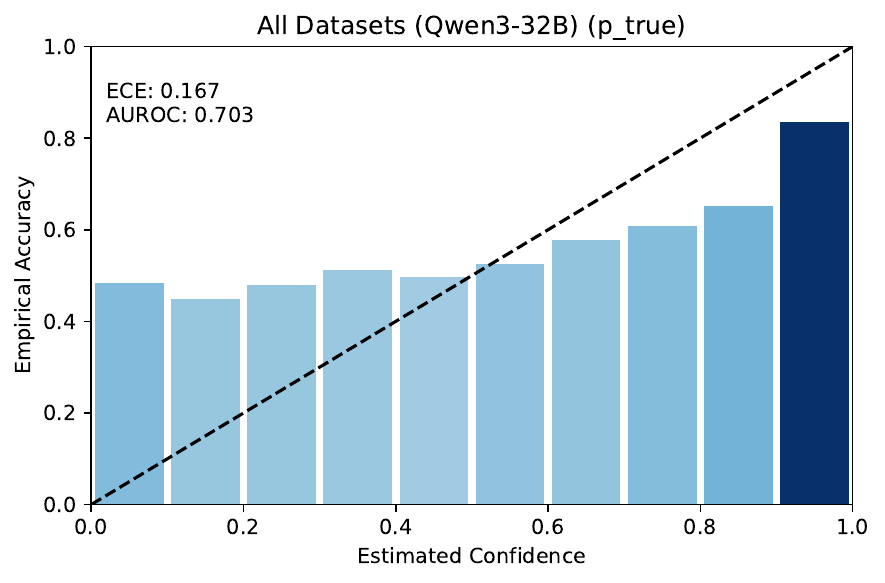}
        \subcaption{p\_true}
        \label{fig:rel-q32b-ptrue}
    \end{subfigure}
    
    \vspace{0.5em}

    \begin{subfigure}[t]{1.0\linewidth}
        \centering
        \includegraphics[width=0.95\linewidth, height=4.2cm, keepaspectratio]{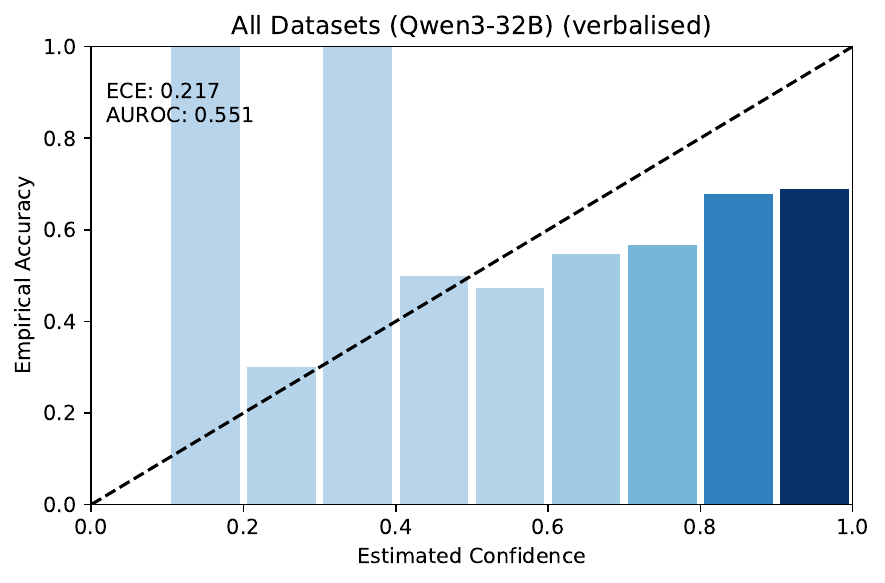}
        \subcaption{Verbalised}
        \label{fig:rel-q32b-verbalised}
    \end{subfigure}

    \vspace{0.5em}

    \begin{subfigure}[t]{1.0\linewidth}
        \centering
        \includegraphics[width=0.95\linewidth, height=4.2cm, keepaspectratio]{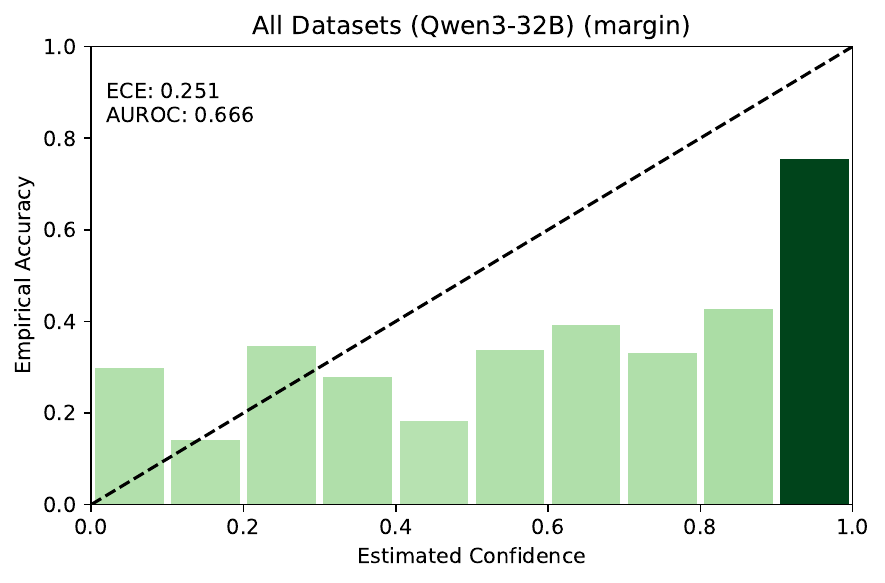}
        \subcaption{Margin}
        \label{fig:rel-q32b-margin}
    \end{subfigure}

    \vspace{0.5em}

    \begin{subfigure}[t]{1.0\linewidth}
        \centering
        \includegraphics[width=0.95\linewidth, height=4.2cm, keepaspectratio]{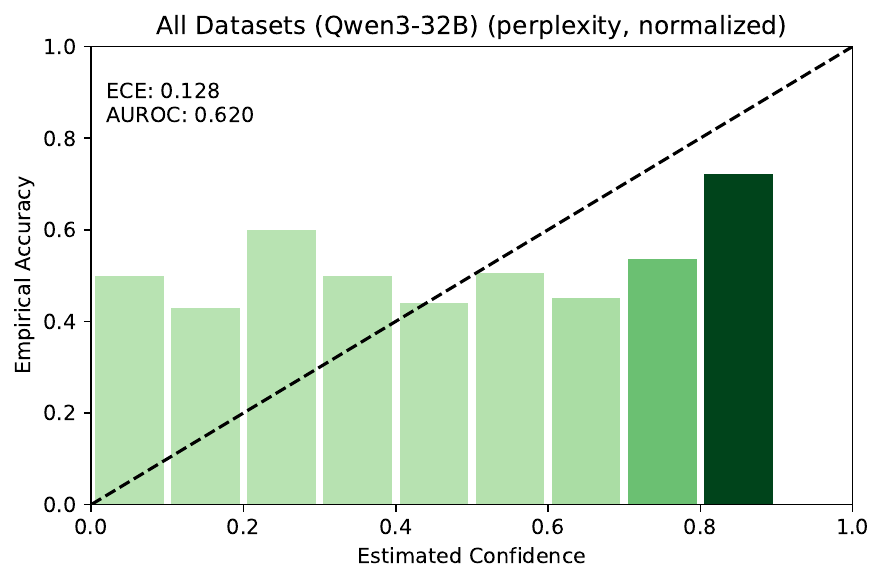}
        \subcaption{Perplexity (norm)}
        \label{fig:rel-q32b-perplexity}
    \end{subfigure}

    \caption{Reliability diagrams for \textbf{Qwen3-32B}. Bars darken with bin count; dashed line is perfect calibration.}
    \label{fig:reliability-qwen3-32b}
\end{figure}

\clearpage 

\begin{figure}[h!]
    \centering
    
    \begin{subfigure}[t]{1.0\linewidth}
        \centering
        \includegraphics[width=0.95\linewidth, height=4.2cm, keepaspectratio]{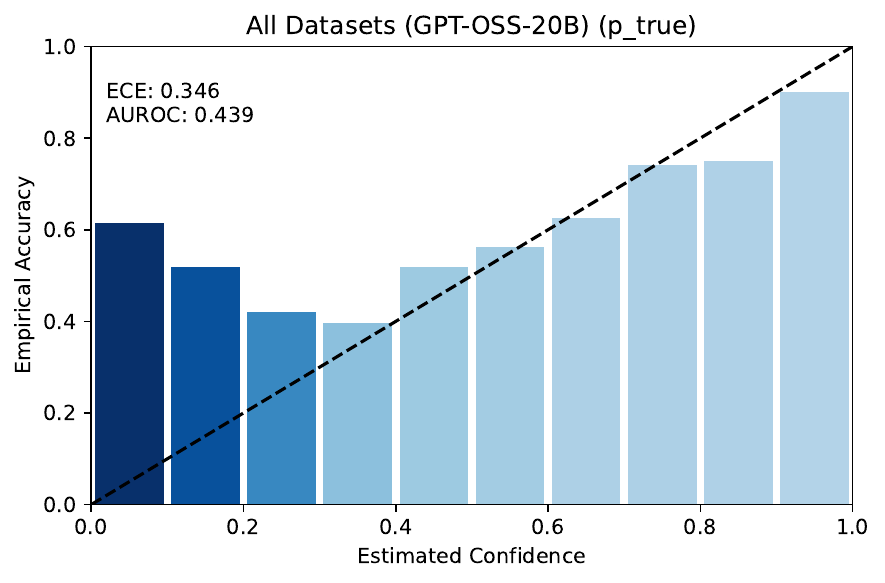}
        \subcaption{p\_true}
        \label{fig:rel-gptoss-ptrue}
    \end{subfigure}
    
    \vspace{0.5em}

    \begin{subfigure}[t]{1.0\linewidth}
        \centering
        \includegraphics[width=0.95\linewidth, height=4.2cm, keepaspectratio]{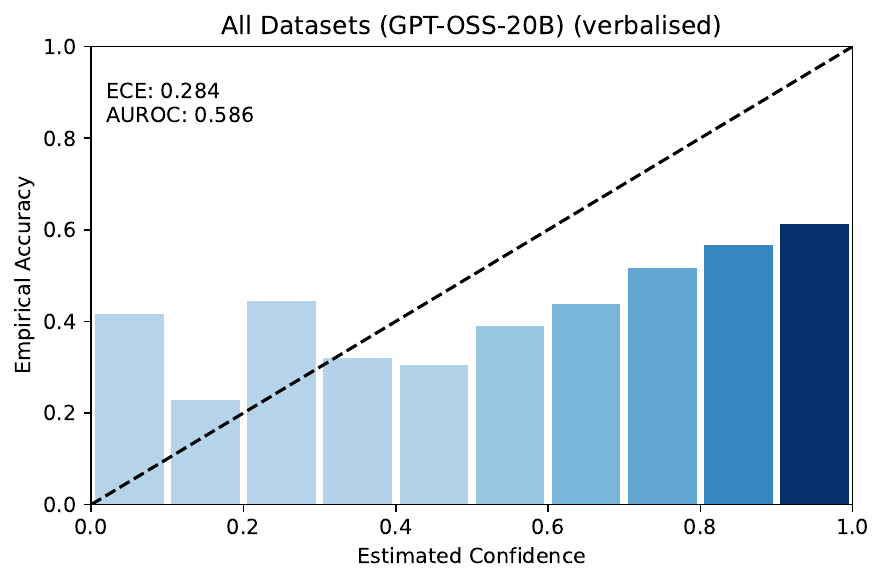}
        \subcaption{Verbalised}
        \label{fig:rel-gptoss-verbalised}
    \end{subfigure}

    \vspace{0.5em}

    \begin{subfigure}[t]{1.0\linewidth}
        \centering
        \includegraphics[width=0.95\linewidth, height=4.2cm, keepaspectratio]{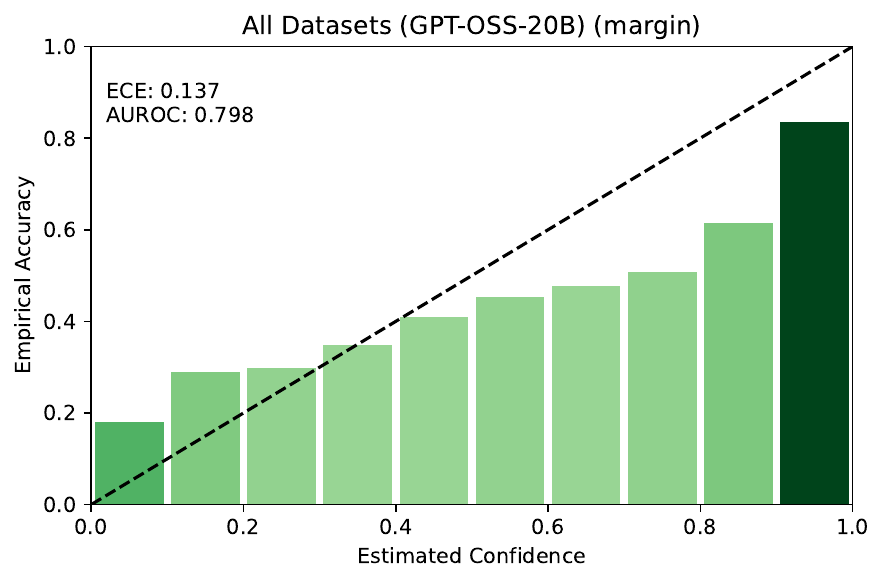}
        \subcaption{Margin}
        \label{fig:rel-gptoss-margin}
    \end{subfigure}

    \vspace{0.5em}

    \begin{subfigure}[t]{1.0\linewidth}
        \centering
        \includegraphics[width=0.95\linewidth, height=4.2cm, keepaspectratio]{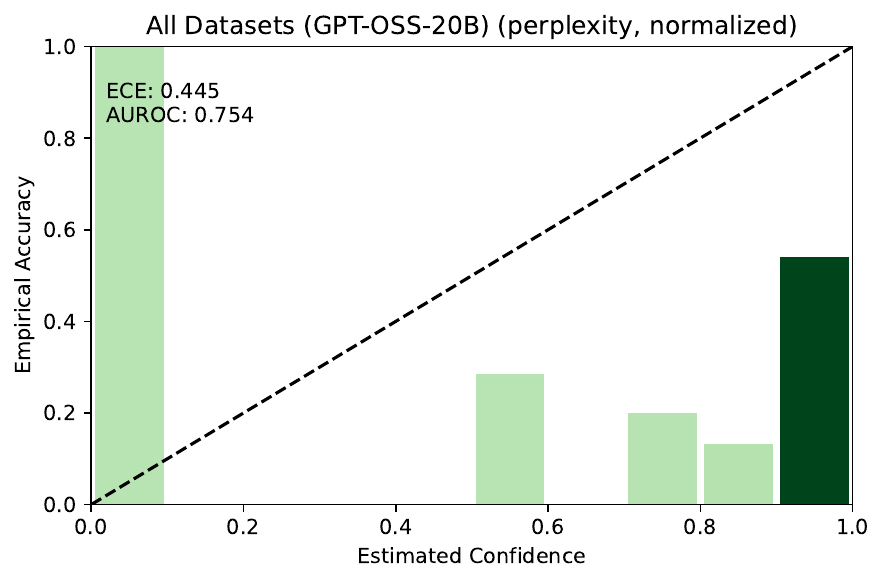}
        \subcaption{Perplexity (norm)}
        \label{fig:rel-gptoss-perplexity}
    \end{subfigure}

    \caption{Reliability diagrams for \textbf{GPT-OSS-20B}. Bars darken with bin count; dashed line is perfect calibration.}
    \label{fig:reliability-gptoss}
\end{figure}
\newpage
\section{GPT-OSS full results}
\label{app:gpt-oss-full-results}
\begin{figure}[h]
    \centering
    \includegraphics[width=\linewidth]{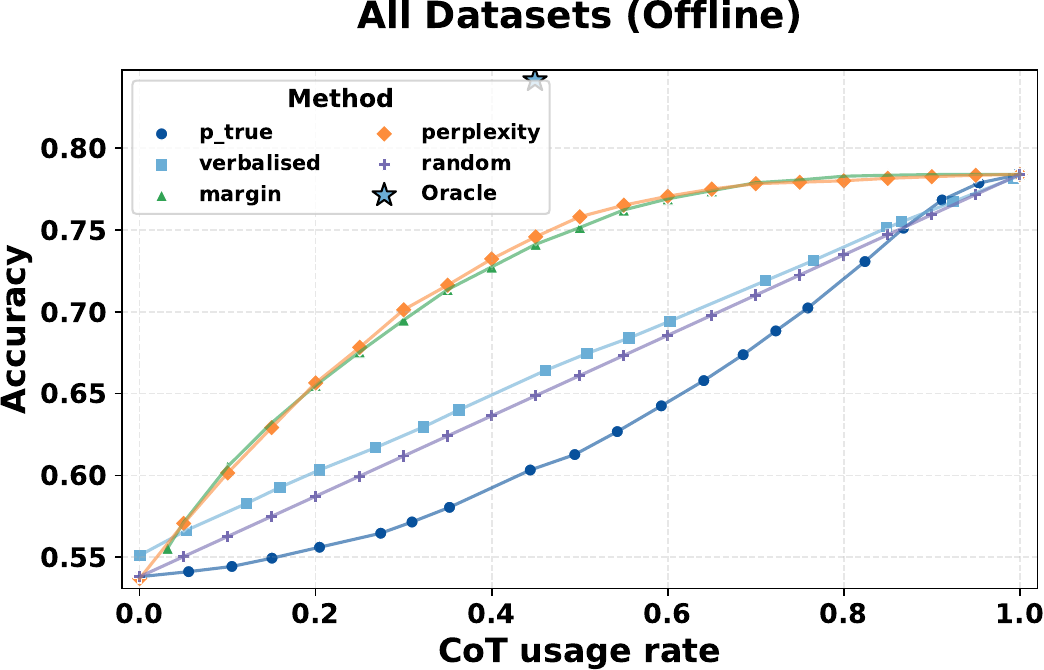}
    \caption{GPT-OSS-20B Accuracy vs. CoT usage results (Low reasoning effort)}
    \label{fig:gpt-oss-low-cot-use}
\end{figure}

\label{sec:gpt-oss-full results}
\begin{figure}[h]
    \centering
    \includegraphics[width=\linewidth]{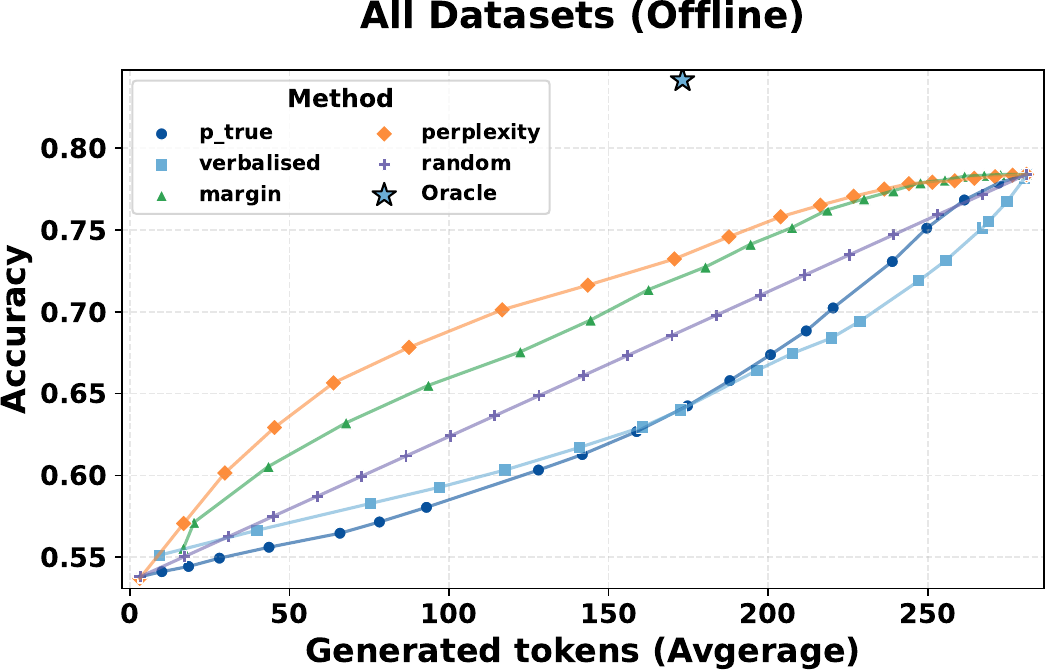}
    \caption{GPT-OSS-20B Accuracy vs. Average Tokens results (Low reasoning effort)}
    \label{fig:gpt-oss-low-tokens}
\end{figure}

\begin{figure}[h]
    \centering
    \includegraphics[width=\linewidth]{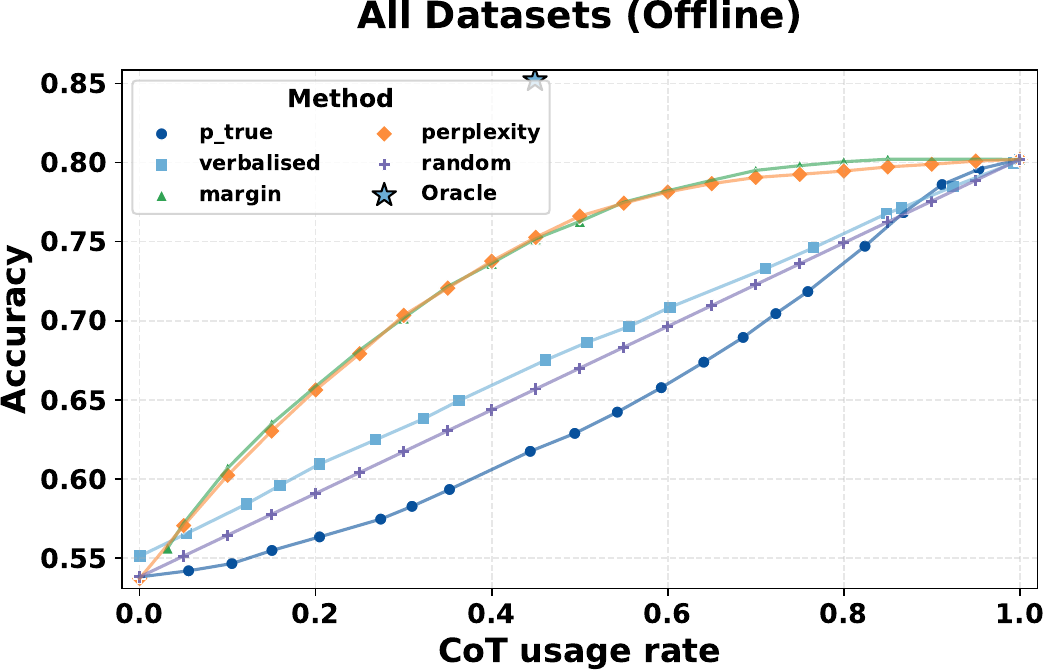}
    \caption{GPT-OSS-20B Accuracy vs. CoT usage results (High reasoning effort)}
    \label{fig:gpt-oss-high-cot-use}
\end{figure}

\begin{figure}[h]
    \centering
    \includegraphics[width=\linewidth]{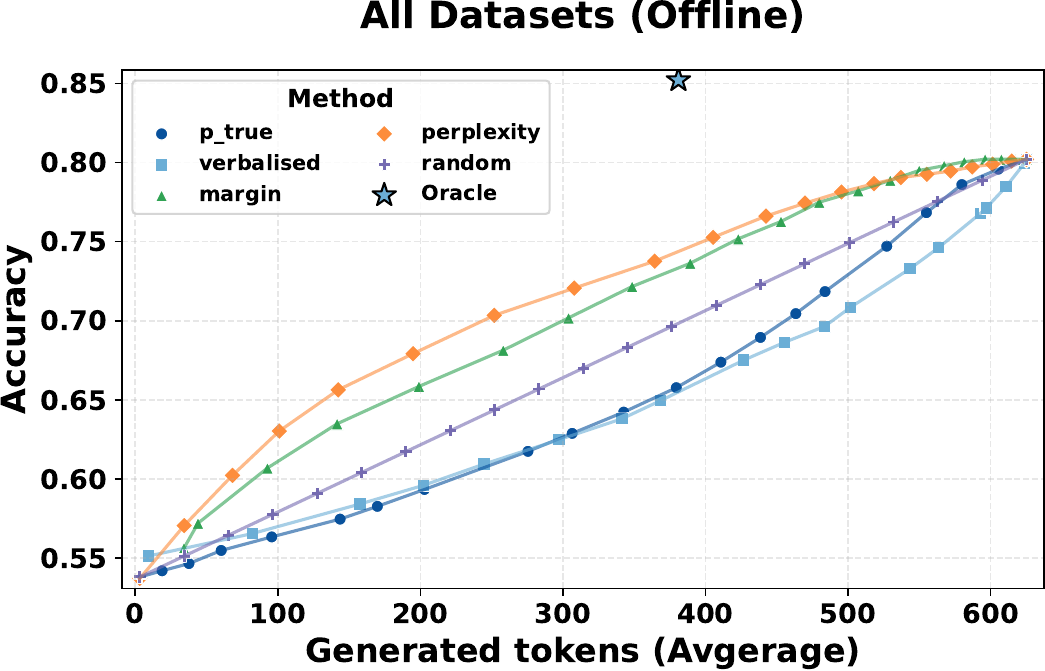}
    \caption{GPT-OSS-20B Accuracy vs. Average Tokens results (High reasoning effort)}
    \label{fig:gpt-oss-high-tokens}
\end{figure}

\end{document}